\documentclass[10pt]{article}
\usepackage{graphicx}
\usepackage{amsmath}
\usepackage{amssymb}
\usepackage{color}
\usepackage{url}
\usepackage{fullpage}

\newcommand{\ba}{\left[\begin{array}}
\newcommand{\ea}{\end{array} \right]}

\def\real{\mathbb{R}}

\def\cut#1{{}}
\def\cutTwo#1{{}}

\begin{document}
\title{Detachable Object Detection: \\ %\vspace{-.2cm} 
{\large Segmentation and Depth Ordering From Short-Baseline Video}}

\author{Alper~Ayvaci \and Stefano~Soatto}

\date{September 5, 2011}
\maketitle

\begin{abstract}
We describe an approach for segmenting an image into regions that correspond to surfaces in the scene that are partially surrounded by the medium. It integrates both appearance and motion statistics into a cost functional, that is seeded with occluded regions and minimized efficiently by solving a linear programming problem. Where a short observation time is insufficient to determine whether the object is detachable, the results of the minimization can be used to seed a more costly optimization based on a longer sequence of video data. The result is an entirely unsupervised scheme to detect and segment an arbitrary and unknown number of objects. We test our scheme to highlight the potential, as well as limitations, of our approach.
\end{abstract}
% \begin{keywords}
% Object Detection, Video Segmentation, Occlusion, Layers, Graph Cuts, Ordering Constraints, Model Selection.
% \end{keywords}}

\section{Introduction}
\label{sect-intro}

\begin{figure}[htb]
\begin{center}
\includegraphics[height=3.1cm]{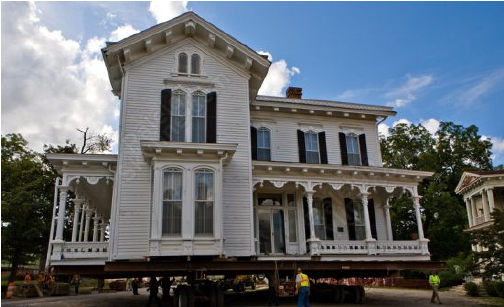}
\includegraphics[height=3.1cm]{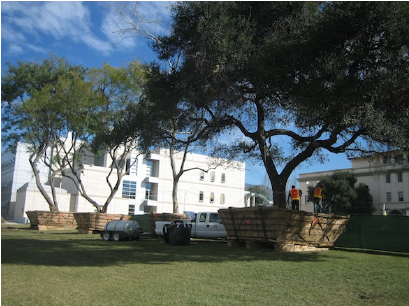}
\includegraphics[height=3.1cm]{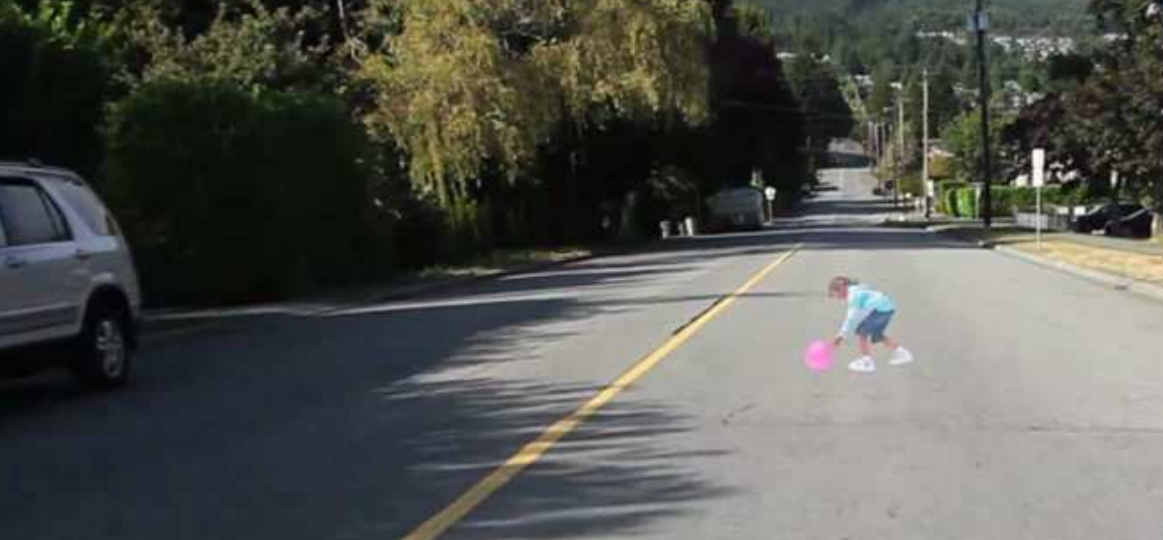}
\end{center}
%\vspace{-.3cm} 
\caption{\sl Although our definition of ``detachable'' can be inconsistent with the dictionary notion of ``something that can be detached'' the two may be closer than they appear. Houses and trees satisfy both definitions, as these images illustrate. The girl figure painted on the road, on the other hand, fails both definitions: It can be erased, but not detached without breaking its surfaces. While it is in contact with the medium (air), it is not partially sur{\em rounded} by it.}
\label{fig-webster}
\end{figure}

 A ``detached object'' was defined by Gibson  \cite{gibson84} as {\em ``a layout of surfaces completely surrounded by the medium.''} He argued that a ``topologically closed surface can be moved without breaking its surface.'' This property is functionally important as it gives objects {\em ``typical affordances like graspability''}. 
Unfortunately, unless they are floating in midair, most objects are attached to something. Absent the ability to actively intervene by attempting to grasp an object, the most we can determine from passive imaging data is whether it is {\em partially} surrounded by the medium.  Therefore, we call {\em ``detachable object''} a {\em (compact and simply-connected) subset of the domain of an image that back-projects onto a layout of surfaces that is partially surrounded by the medium}. These include protruding objects resting on the ground plane or hanging (Fig. \ref{fig-occlusions}), but not flat pictures (Fig. \ref{fig-webster}-right). Detachable objects are defined in the {\em image}, rather than the {\em scene}, since we want to detect them without having to explicitly reconstruct a spatial layout of surfaces. However, they correspond to regions of space that {\em may} be detached, given sufficient force (Fig. \ref{fig-webster}-left).
A direct consequence of the definition is that detachable objects yield {\em occlusion phenomena} in response to either object or viewer {\em motion.} A single image is not sufficient to determine whether an object is detachable. At least {\em two} are necessary (three, in our approach), and more are beneficial.

In this paper, therefore, we detect detachable objects in two stages: First, we detect {\em occlusion regions}. They provide local (and sparse) evidence of detachable objects, as well as {\em local} depth ordering constraints. In a second stage, these constraints are integrated into a partition of the entire image into different depth layers (Sect. \ref{sect-formulation}).

The first stage requires motion. While in some cases two adjacent video frames may be sufficient to generate occlusion regions at the interface with the medium, there is typically no motion discontinuity at the support region where the object is attached (e.g., the right foot in Fig. \ref{fig-occlusions}-left). Photometric statistics (color, texture, intensity) may provide evidence of the boundary of the object, but in general extended observation is needed to determine whether the contact region changes, and therefore the object is detachable. For instance, the right foot in Fig. \ref{fig-occlusions}-right, is eventually lifted, the bench in Fig. \ref{fig-qualitative2} (top row) is seen against a varying background during the short sequence, and a moving car changes its point of contact with the ground, so they are all detachable, even though at no point are they actually detached.

The partition of the overall detection task into two stages sacrifices end-to-end optimality. However, it comes with a considerable benefit: The first stage is known to admit a convex relaxation \cite{ayvaciRS11IJCV}; we show that the second reduces to a {\em linear programming} problem. The {\bf key idea} of our approach is to use occlusion regions as ``seeds'' in a supervised segmentation scheme, so the overall system is entirely unsupervised, and does not require any manual labeling or annotation. We also perform model selection, that is the determination of the number of detachable objects, {\em all by solving a linear program.} What we pay in end-to-end optimality we gain in computational efficiency. If optimality and long-term temporal integration is a concern,  we can always use our results to seed a global variational optimization scheme \cite{jacksonYS06}.

\subsection{Related work}
\label{sect-related}

This paper relates to a vast literature on video-based segmentation, for instance \cite{jacksonYS06,schoenemannC08,cremersS05,huangLM09,baiSS09,ungerMPB09}. Such approaches work well with few independently moving objects and require a reasonable initialization that is increasingly difficult as the scene becomes more cluttered. Our method does not require knowledge of the number of objects, nor any user input, initial bounding boxes, or scribbles \cite{wangXYSC04,baiSS09}. Instead, occluded regions, detected with \cite{ayvaciRS11IJCV}, provide the ``supervision mechanism'' to seed to our method. For this reason, we refer to ``short-baseline video'' in our title, although the method is agnostic to whether one has short- or wide-baseline motion, video or unordered snapshots, ego-motion or object motion, so long as they yield occlusion evidence. In principle we could forgo a two-step approach and directly segment a video by processing it in a batch as in \cite{huangLM09,broxM10,brendelT10}; however, we find that the algorithmic benefits outweigh the loss of end-to-end optimality \cite{vazquezAPM10}.

Occlusions have of course been used before as a cue for layering: \cite{brostowE99}, however, assumes a static camera; \cite{ogaleFA05} classifies three kinds of occlusions to prime motion segmentation; \cite{feldmanW08} perform local analysis using spatio-temporal filters to estimate depth ordering. Other  methods for layered motion segmentation \cite{iraniP93,jepsonFB02,pawanTZ08,smithDC04} also take occlusions into account, but use more restrictive parametric motion models, and do not scale well as the number of object increases beyond few. Similarly, \cite{steinSH08,apostoloffF06} use occlusion boundaries inferred using appearance, motion and depth cues \cite{steinH09,sarginBMR09,heY10} or T-junctions \cite{apostoloffF05} to segment image sequences; however, they require the number of segments to be known \emph{a priori}. Our work is rather different in spirit from those attempting to obtain a depth map from a single image \cite{morelS08,amerRT10}. 

We capitalize on efficient {\em supervised segmentation} algorithms that, starting from labeled seeds (``scribbles'' for background/foreground \cite{grady06,boykovVZ02}), produce a segmentation by solving a linear program. However, we {\em use the occluded regions as ``local scribbles'' in a supervised segmentation scheme, Fig. \ref{fig-occlusions}.} This evidence is globally integrated into a graph partitioning algorithm that can be solved by linear programming. It assigns each pixel to a depth ordering label that provides a putative segmentation of detachable objects, even in the presence of {\em multiple objects} that otherwise cause standard {segmentation schemes to lose convexity \cite{chanE05}}. This idea is formalized in the next section where we introduce our optimization scheme.

In our implementation, for computational convenience we solve our linear program on a superpixel graph, rather than the image lattice. However, this step is not conceptually necessary, and can be foregone.

\section{From local occlusion ordering to global consistency}
\label{sect-formulation}

\begin{figure}[bth!]
  \centerline 
      {
        \hbox
            {%\\
               \begin{tabular}{ccc|ccc}
               {\includegraphics[width=0.16\textwidth]{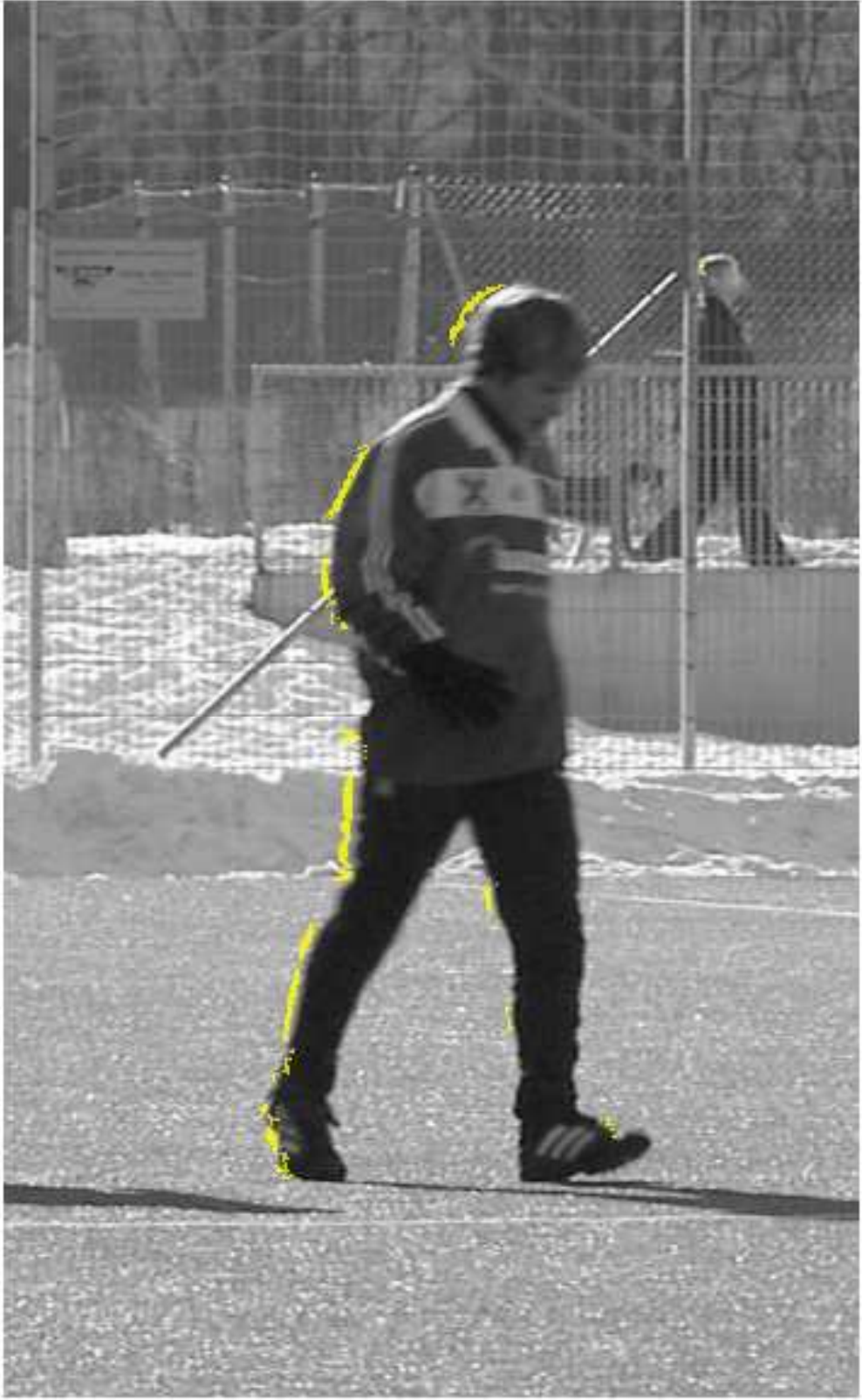}}   	                
               {\includegraphics[width=0.16\textwidth]{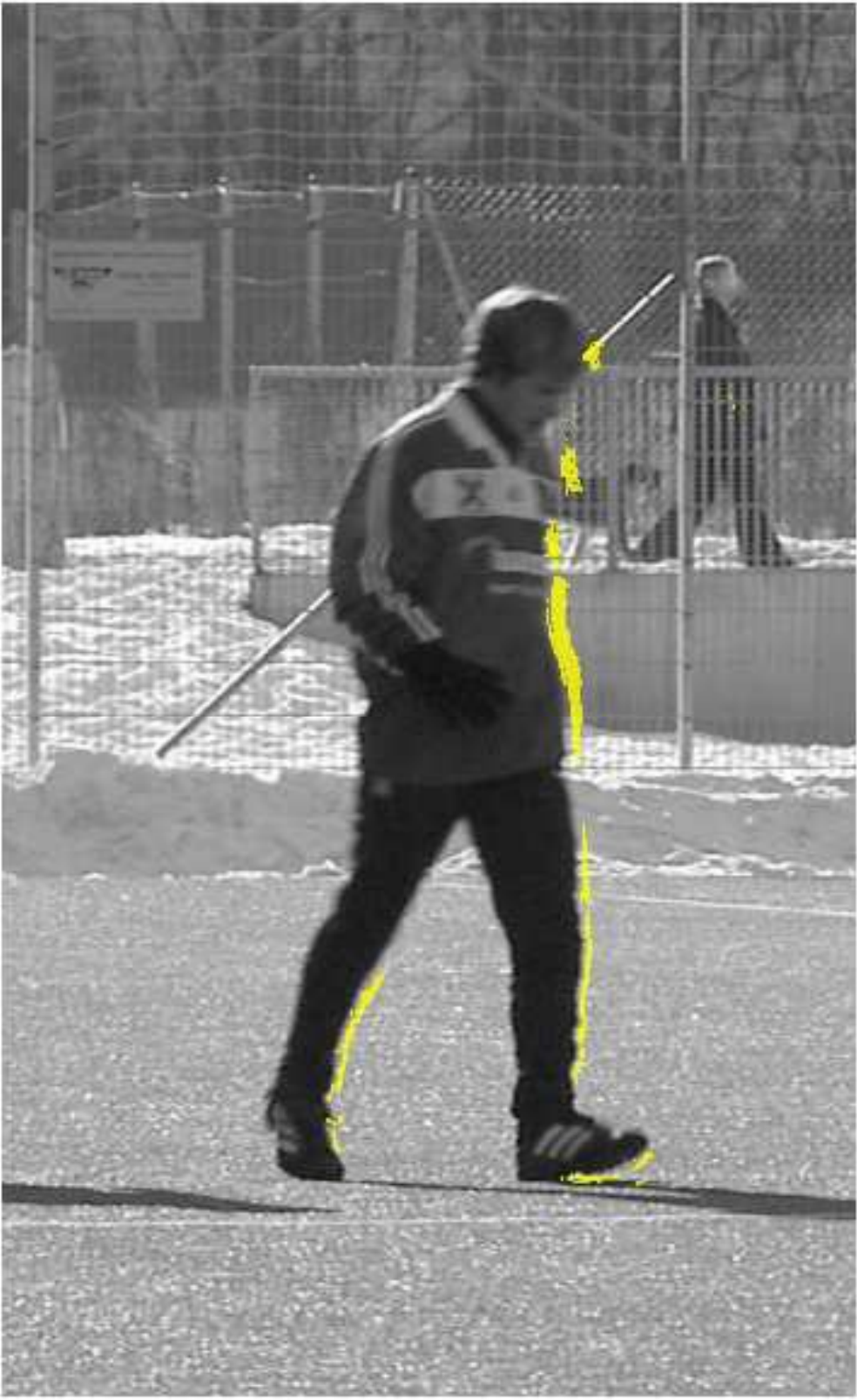}}   	                
               {\includegraphics[width=0.16\textwidth]{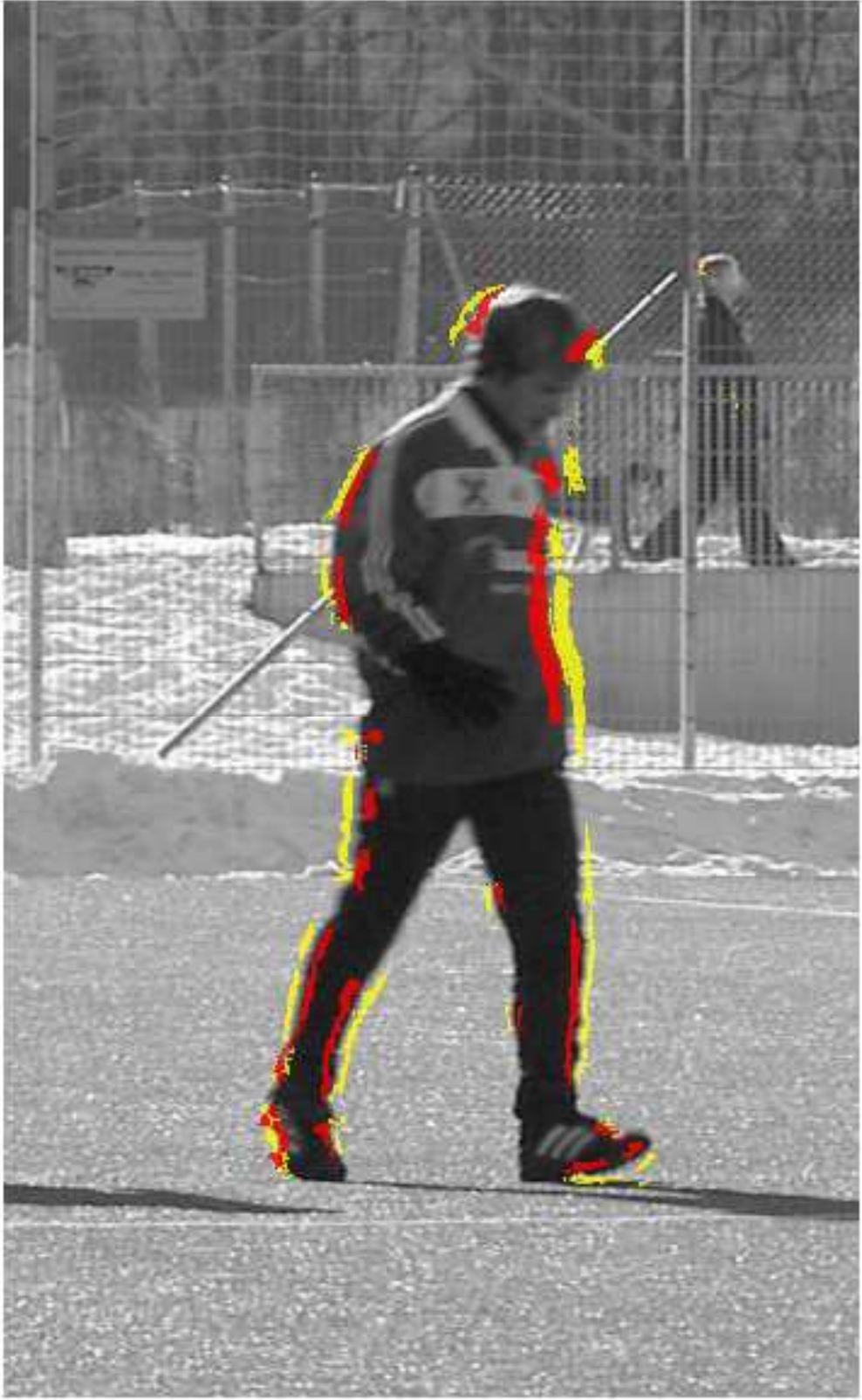}}
             {\includegraphics[width=0.16\textwidth]{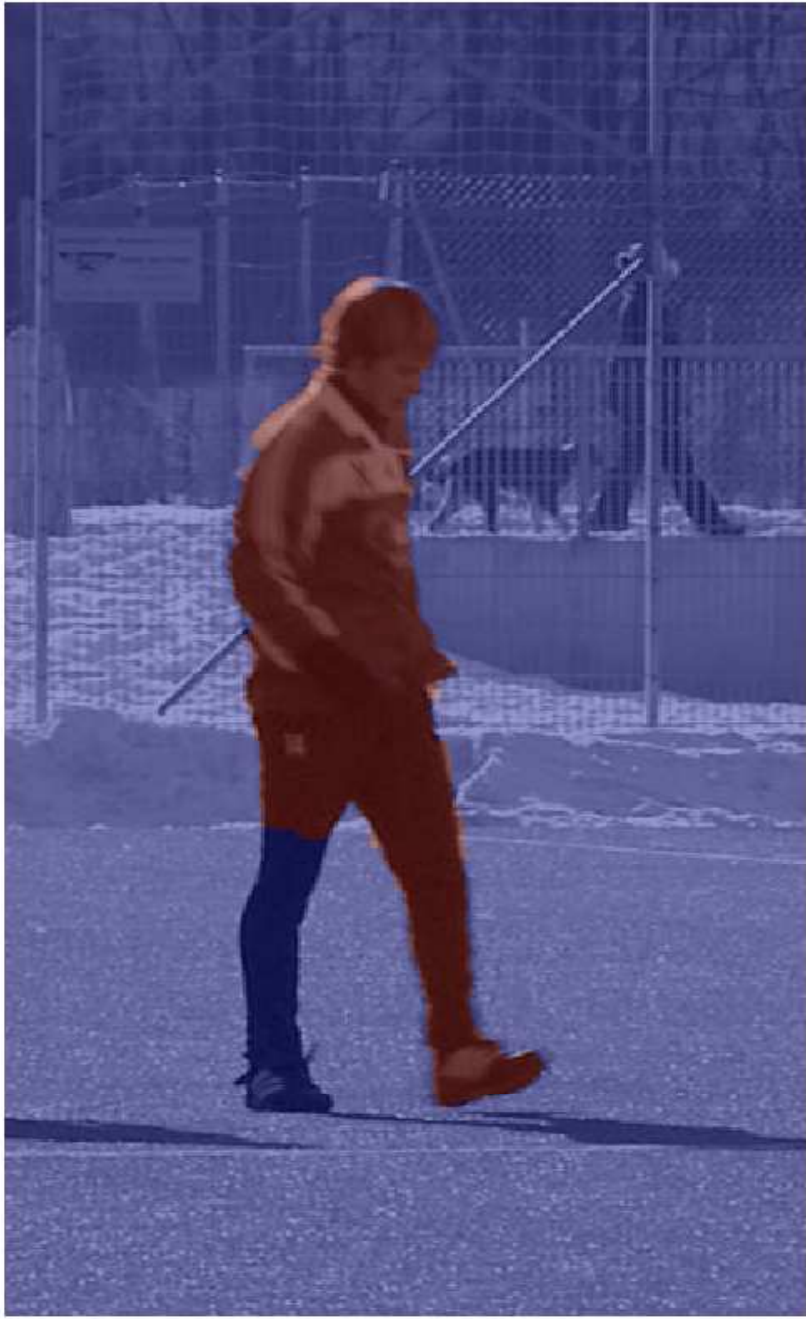}}   	                
             {\includegraphics[width=0.16\textwidth]{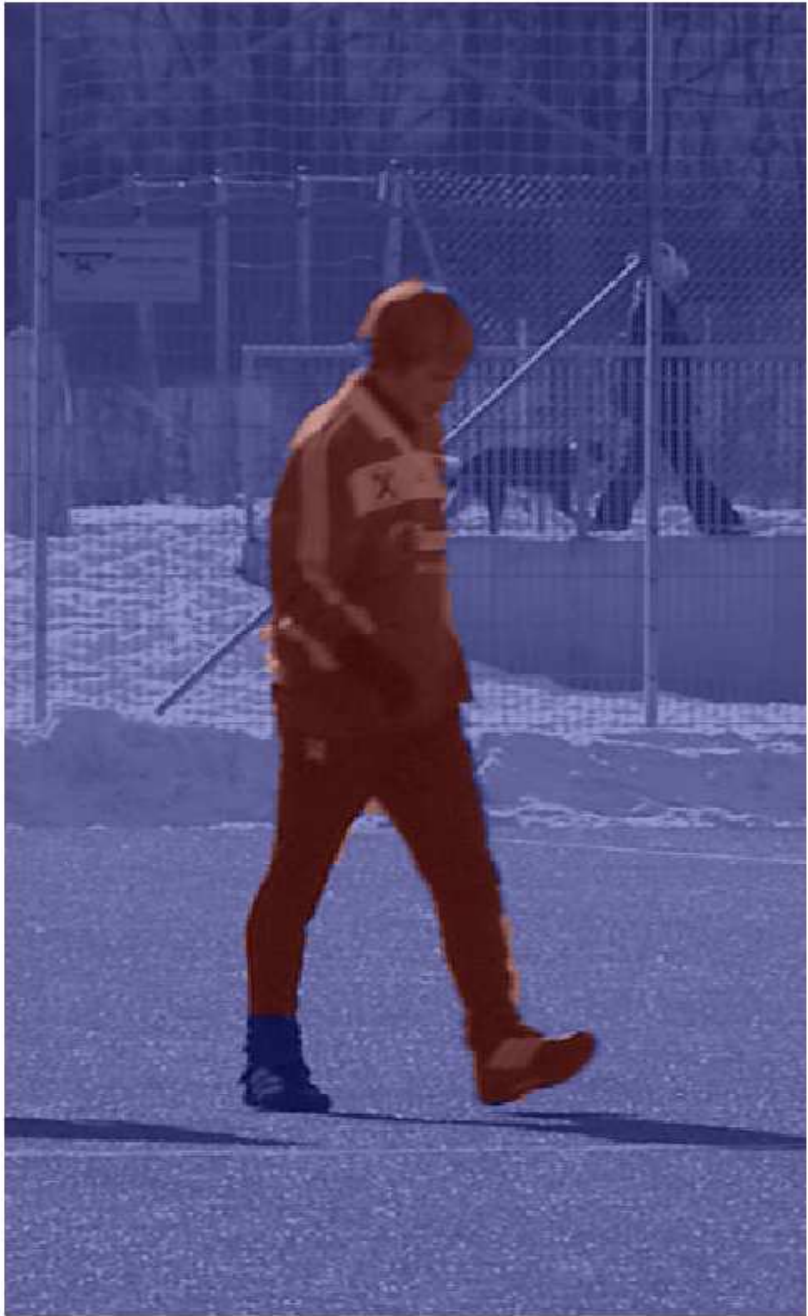}}   	                
             {\includegraphics[width=0.16\textwidth]{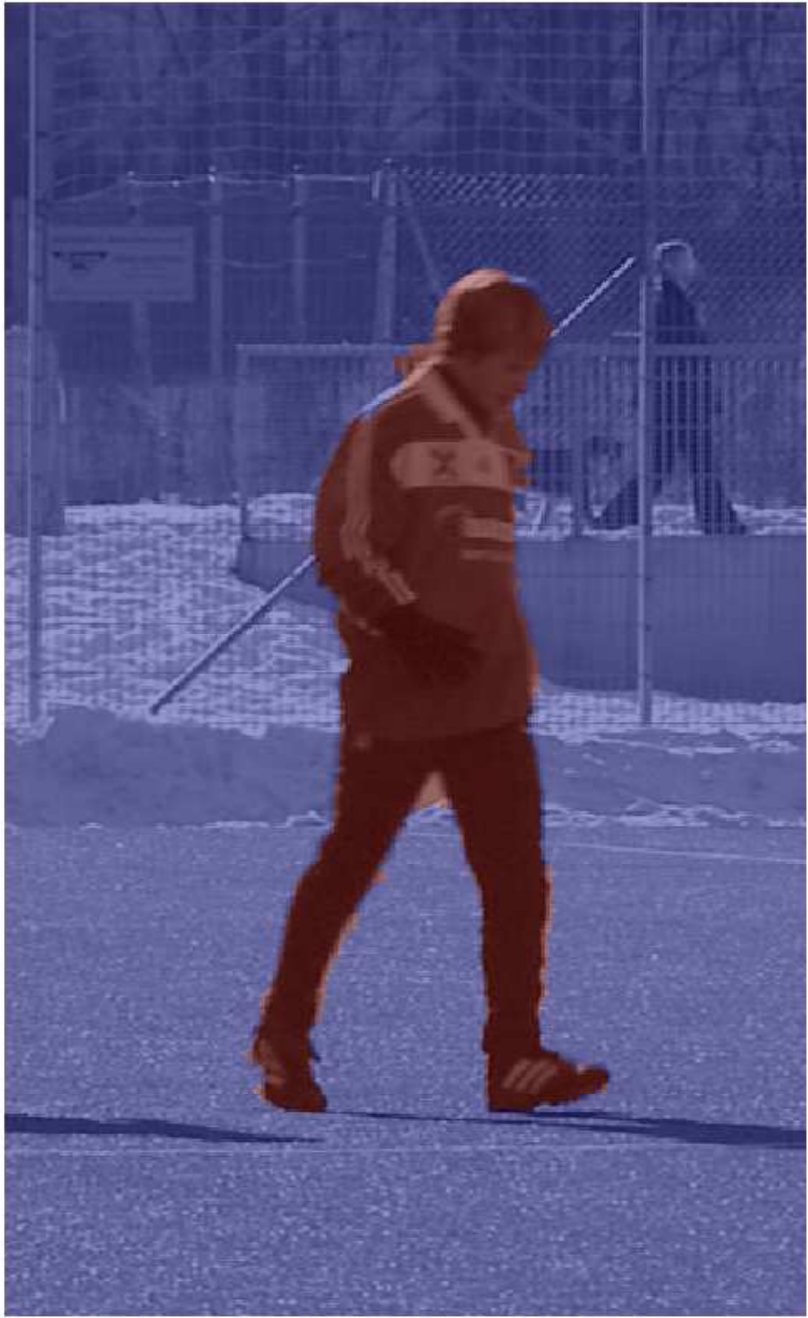}}
            \end{tabular}
            }
      }

		\caption{\sl Left to right: $\Omega_{-}(t)$ (yellow); $\Omega_{+}(t)$ (yellow); $\Omega$ (yellow) and $\Omega^c$ (red) on the 168$^{th}$ frame of the \emph{Soccer} sequence \cite{boltz2008motion}. Segmentation based on short-baseline motion does not allow determining whether the right foot and leg are detachable; however, extended temporal observation enables eventually to associate the entire leg with the body, and therefore detecting the person as a whole detachable object.}		
		 \label{fig-occlusions}
\end{figure}

Let $I:D \subset \real^2\times \real^+ \rightarrow \real^+; \ (x,t) \mapsto I_t(x)$ be a grayscale time-varying image sequence defined on a domain $D$\cut{ which is quantized into an ${M \times N}$ lattice}. Under the assumption of Lambertian reflection, constant illumination and co-visibility, $I_t(x)$ is related to its (forward and backward) neighbors $I_{t+dt}(x)$, $I_{t-dt}(x)$ by the usual brightness-constancy equation 
\begin{equation}
I_{t}(x) = I_{t\pm dt}(x+ v_{\pm t}(x)) + n_\pm(x), ~~ x\in D \backslash \Omega_{\pm}(t)
\end{equation}
where $v_{+t}$ and $v_{-t}$ are the forward and backward motion fields and the additive residual lumps together all unmodeled phenomena. {\em In the co-visible regions}, such a residual is typically small (in some norm) and spatially and temporally uncorrelated. However, in the presence of parallax motion, there generally are regions in the current image that are not visible in the forward (backward) neighbor, $\Omega_+(t)$ ($\Omega_-(t)$). In these regions, one cannot find a motion field $v_{+t}$ ($v_{-t}$) that maps the image onto its neighbors. Therefore, the residual is typically large and not uncorrelated (unless the occluded region has identical statistics of the occluder, in which case we cannot tell that there has been an occlusion in the first place). One can use this observation to simultaneously determine the motion fields, $v_{\pm t}(x)$ and the occluded regions $\Omega_\pm(t)$ by solving a convex optimization problem \cite{ayvaciRS11IJCV}. From now on, therefore, we assume to be given, at each time $t$, the forward (occlusion) and backward (un-occlusion) time-varying regions $\Omega_+(t), \ \Omega_-(t)$, possibly with errors, and drop the subscript $\pm$ for simplicity. The local complement of $\Omega$, i.e. a subset of $D\backslash \Omega$ in a neighborhood of $\Omega$, is indicated by $\Omega^c$ and can be obtained by using morphological dilation operators, or simply by duplicating the occluded region on the opposite side of the occlusion boundary  (Fig. \ref{fig-occlusions}). 

It is important to note that these regions are in general {\em multiply-connected}, so $\Omega = \cup_{k=1}^K \Omega_k$, and each connected component $\Omega_k$ may correspond to a different  occluded region. However, occlusion detection is a {\em binary classification} problem because each region of an image is either {\em co-visible} (visible in a temporally adjacent image) or not, regardless of how many detachable objects populate the scene. In order to detect the {\em (multiple)} detachable objects, we must aggregate local depth-ordering information into a global depth-ordering model.  To this end, we define  a {\em label field} $c:D \times \real^+ \rightarrow {\mathbb Z}^+; x \mapsto c(x,t)$ that maps each pixel $x$ at time $t$ to an integer indicating the depth order, $c(x,t)$. For each connected component $k$ of an occluded region $\Omega$, we have that if $x \in \Omega_k$ and $y \in \Omega^c_k$, then $c(x,t) > c(y,t)$ (larger values of $c$ correspond to objects that are closer to the viewer). If $x$ and $y$ belong to the same object, then $c(x,t) = c(y,t)$. To enforce label consistency within each object, we therefore want to minimize $|c(x,t)-c(y,t)|$, but we want to integrate this constraint against a data-dependent measure that allows it to be violated across object boundaries. Such a measure, $d\mu(x,y)$, depends on both motion and texture statistics, for instance, for the simplest case of grayscale statistics, we have
$d\mu(x,y) = W(x,y) dxdy$ where

\begin{small}
\begin{equation}
	W(x,y) = \begin{cases} 
	 	        ~ \alpha e^{-(I_t(x)-I_t(y))^2} +  \beta e^{-\|v_t(x)-v_t(y)\|_2^2} \\
	            ~ 0 ~~~ {\rm otherwise;} ~~~~~~~~~~~ \| x-y \|_2 < \epsilon,
             \end{cases}
\end{equation} 
\end{small}

\noindent where $\epsilon$ identifies the neighborhood, $\alpha$ and $\beta$ are  the coefficients that weight the intensity and motion components of the measure. 
We then have 
\begin{equation}
\begin{split}
\hat c = & \arg\min_{c: D\rightarrow {\mathbb Z}} \int_D | c(x,t)  - c(y,t) | d\mu(x,y) \\
& {\rm s.\ t.} \ c(x,t) < c(y,t)  \ \forall \ x\in \Omega_k(t), y\in \Omega_k^c(t), \ k = 1,.., K,
\end{split}
\label{eq-pde}
\end{equation}
and $ \|x-y\|_2 < \epsilon$. This problem would be solved trivially by a constant, e.g., $c(x) = 0$, if it were not for the boundary conditions imposed by occlusions. 

To translate this into a linear program, we quantize $D$ into an $M \times N$ grid-graph $G=(V,E)$ with the vertex (node) set $V$ (pixels or super-pixels), and the edge set $ E \subseteq V \times V$ denoting adjacency of two nodes $i,j \in V$ via $i \sim j$.  We then identify $i, j$ with $x_i, x_j$, their corresponding depth ordering  $c_i = c(x_i,t), \ c_j = c(x_j,t)$, and the measure $d\mu(x_i,x_j)$ is a symmetric positive-definite matrix { $w_{ij} = W(x_i, x_j)$} that measures the {\em affinity} between two nodes $i,j$. The problem (\ref{eq-pde}) then becomes the search for the discrete-valued function $c:V\rightarrow {\mathbb Z}^+$
\begin{equation}
\begin{split}
\{\hat c_i\}_{i=1}^{MN} = & \arg\min_{c}   ~~ \sum_{i \sim j}  w_{ij} |c_i - c_j|  \\
\text{s. t. } & \ c_i < c_j,~~ i \sim j,~  i \in {\Omega}_k(t), ~  j \in \Omega_k^c(t), \\
\end{split}
\label{equation:relaxed-depth-ordering}
\end{equation}
with $1 \leq c_i \leq L$. In the case of $L=2$, this problem can be interpreted as binary graph cut \cite{sinopG07}. Unfortunately,  for $L > 2$ this is an NP-hard problem so, as customary, we relax it by dropping the integer constraint and allowing $c: V \rightarrow {\mathbb R}^+$. 

\section{Automatic model selection}
\label{sect-model-selection}

A natural criterion for model selection is to trade off model complexity with data fidelity, as customary in {\em minimum-description length} (MDL)  \cite{rissanen78}. In our case, an obvious complexity cost is the number of objects, that is the largest value taken by the {\em label field},  $\| c \|_\infty \doteq \max\{ |c_i| \}_{i = 1}^{MN}$. This leads to the straightforward modification of the problem (\ref{equation:relaxed-depth-ordering}) into  
\begin{equation}
\begin{split}
\{\hat c_i\}_{i=1}^{MN} = & \arg\min_{c} ~~ \sum_{i \sim j}  w_{ij} |c_i - c_j| + \gamma \|c\|_{\infty}  \\
\text{s. t. } & \ c_i < c_j,~~ i \sim j,~  x_i \in {\Omega}_k(t), ~  x_j \in \Omega_k^c(t), \\
\end{split}
\label{equation:relaxed-depth-ordering-mdl}
\end{equation}
with $1 \leq c_i$ where $\gamma$ is the cost for adding a new layer. 
While this problem preserves the convexity properties of the original model, it is not amenable to being solved using linear programming (LP). Therefore, we introduce auxiliary variables  $\{u_{ij} | i \sim j\}$ and $\sigma$, so that (\ref{equation:relaxed-depth-ordering-mdl}) can be written as 
\begin{equation}
\begin{split}
\min_{u_{ij}, c_i, \sigma} & \sum_{i \sim j} w_{ij} u_{ij} + \gamma \sigma   \\
\text{s. t. } 
& ~~  1 \preceq c \preceq \sigma, \\ 
& ~~  c_j - c_i \geq 1, ~~ i \sim j,~  x_i \in {\Omega}_k(t), ~  x_j \in \Omega_k^c(t) \\
& ~ -u_{ij} \leq c_i - c_j \leq u_{ij}.
\end{split}
\label{equation:relaxed-depth-ordering-lp}
\end{equation}
\cut{The addition of the auxiliary variables may at first appear to further complicate the model (\ref{equation:relaxed-depth-ordering-mdl}) and increase its complexity (measured by the number of free parameters). However, }This  makes the problem amenable to deployment of a vast arsenal of efficient numerical methods.
Note that we have relaxed the integer constraint by allowing the difference between label values to be {\em at least} one. As customary, we will quantize the label map after solving the optimization problem by rounding its values to the nearest integer. 

\section{Seeding extended temporal observations}

As we have anticipated, one could wrap occlusion detection, motion estimation, and depth ordering into one large optimization problem. This can be done by combining (\ref{eq-pde}) with the cost functional in \cite{ayvaciRS11IJCV} and summing over time through an entire video sequence. The result is equation (2) of \cite{jacksonYS06}, and the ensuing optimization is unwieldy. Therefore, for simplicity and modularity, we prefer to keep the two stages separate, and describe the simplest form of temporal integration, that is to use the results of (\ref{equation:relaxed-depth-ordering-mdl}) at each instant as initialization to the optimization at the subsequent time, using the field $v_{-(t)}$. We redefine the measure to incorporate the previous layer estimate by replacing $W(x,y)$ with 
\begin{equation}
W(x,y) + \frac{1}{\tau} H(c(x + v_{-t}(x), t-1), c(y + v_{-t}(y), t-1)), 
\end{equation} 
where $H: \real \times \real \rightarrow \{ 0,1 \}$ is defined by
\begin{equation}
H(a,b) = \begin{cases} 1, ~~ a=b, a > 1, b > 1, \\
                       0, ~~ {\rm otherwise,}
\end{cases} 
\end{equation}
and $\tau$ is a forgetting factor. We show in the experimental section that this simple model is sufficient to aggregate parts detachable objects in the great majority of cases.

\section{Revisiting occlusion errors}
\label{sect-algorithm}

Since we have decomposed the original problem into two separate stages, it is important for the second to handle the inevitable errors made by the first. To model errors in occlusion detection, we introduce slack variables $\{\xi_k\}_{k=1} ^K$ to relax the hard constraints, to 
\begin{equation}
\begin{split}
\min_{u_{ij}, c_i} & \sum_{i \sim j} w_{ij} u_{ij}  + \lambda \sum_{k=1}^K \xi_k\\
\text{s. t. } 
& ~~ 1 \preceq c \preceq L, \\ 
& ~~ c_j - c_i \geq 1 - \xi_k, ~~ i \sim j,~  i \in {\Omega}_k(t), ~ j \in \Omega_k^c(t) \\
& ~~ 0 \leq \xi_k \leq 1 \ \forall \ k, \\
& ~ -u_{ij} \leq c_i - c_j \leq u_{ij}.
\end{split}
\label{equation:relaxed-depth-ordering-soft}
\end{equation}
where $\lambda$ is {the penalty for violating the ordering constraints.} The problem of detachable object detection is finally in a form that is suitable for 
a standard numerical solver, for instance \cite{cvx}. The (forward-backward) occluded regions $\Omega_{\pm}(t)$ are given from \cite{ayvaciRS11IJCV}. Note that layers obtained may consist of multiple objects, so to enforce simple connectivity of a detachable objects we can isolate each connected component using standard morphological operators within each depth level on $c$.

\section{Experiments}
\label{sect-experiments}

Rather than solving (\ref{equation:relaxed-depth-ordering-soft}) on the pixel grid, we pre-compute a partition of the domain into $N$ non-overlapping superpixels $\{s_i\}_{i=1}^N$ such that $\bigcup_{i=1}^N s_i = D, ~ s_i \cap s_j = \emptyset, ~ \forall i \neq j$, as done by \cite{steinH09}, using a watershed-based approach driven by a statistical multi-cue edge detector \cite{martinFM04}. However, since superpixels may cross occlusion boundaries, we subdivide them to ensure that each superpixel is a subset of one of the three regions: $\Omega$, $\Omega^c$ and $D \backslash (\Omega \cup \Omega^c)$. This superpixelization is not necessary, but it reduces the size of the linear program while enabling integration of simple low-level cues. Since the minimization (\ref{equation:relaxed-depth-ordering-soft}) is already written for a general graph, it does not matter whether it is performed on the pixel grid or the superpixel graph. The only change is the weight matrix $w_{ij}$, that in the case of superpixels is given by 
\begin{equation}
	w_{ij} = |\partial s_i \cap \partial s_j| [ \alpha e^{-(\bar{I}(s_i) - \bar{I}(s_j) )^2}   + \beta e^{-\|\bar{v}(s_i) - \bar{v}(s_j) \|_2^2}  + \kappa(1-\overline{Pb}(s_i, s_j)) ],
\end{equation}
where $
\bar{I}(s) = \dfrac{1}{|s|}\int_{s} I_t(x)dx$, $\bar{v}(s) = \dfrac{1}{|s|}\int_{s} v_t(x)dx$ and 
\begin{equation}
\overline{Pb}(s, s^{\prime})= \dfrac{1}{\partial s \cap \partial s^{\prime}}\int_{\partial s \cap \partial s^{\prime}} Pb(x)dx, 
\end{equation}
 and $Pb: D \rightarrow [0,1]$ is the probability of a location being a (material, occlusion, or illumination) boundary. Note that the edge features are incorporated into the computation of the weights since the domain $D$ is partitioned based on $Pb$. In our experiments, we have assigned the parameters $\alpha$, $\beta$ and $\kappa$ to $0.25$, $0.5$ and $0.25$  respectively.

\cut{In our experiments, to reach to a segmentation for layer assignment, we simply find the connected components for each depth level.}
 
{We have used the CMU Occlusion/Object Boundary Dataset\footnote{http://www.cs.cmu.edu/$\sim$stein/occlusion\_data/} \cite{steinH09} to evaluate our approach. It includes $16$ test sequences with a variety of indoor and outdoor scenes, mostly seen under small camera motion.  It provides ground truth segmentation for a single reference frame in each sequence. We also report our results on the {\emph Soccer} sequence \cite{boltz2008motion}, to illustrate the typical effect of extended temporal observations, and on a sequence portraying a drawing of a girl playing ball near an intersection in West Vancouver (Fig. \ref{fig-painted-child}). 
}
\subsection{Qualitative performance}

\begin{figure*}[bth!]
  \centerline 
      {
        \hbox
            {%\\
               \begin{tabular}{cccc}
               {\includegraphics[width=0.24\textwidth]{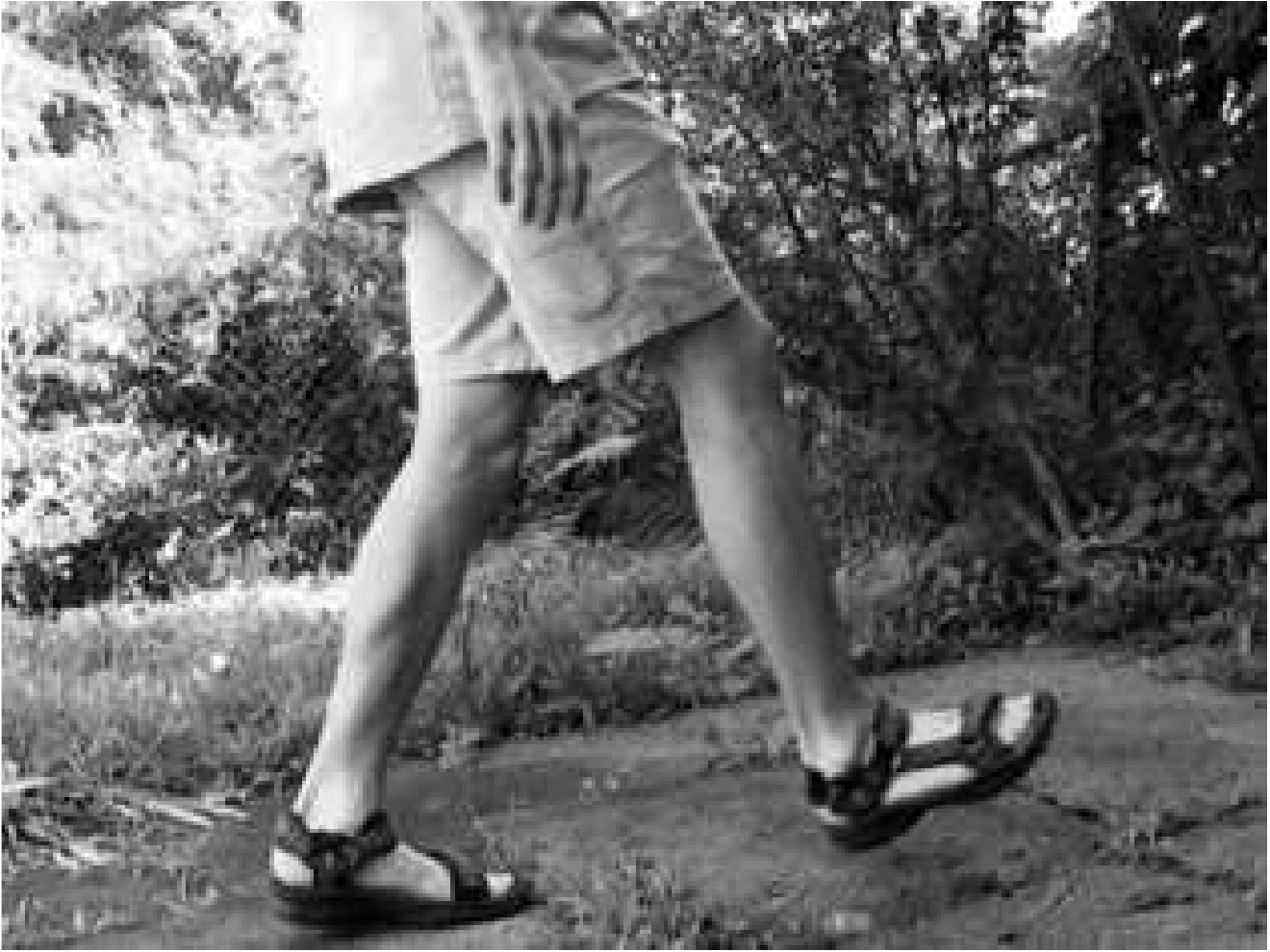}}   	                
               {\includegraphics[width=0.24\textwidth]{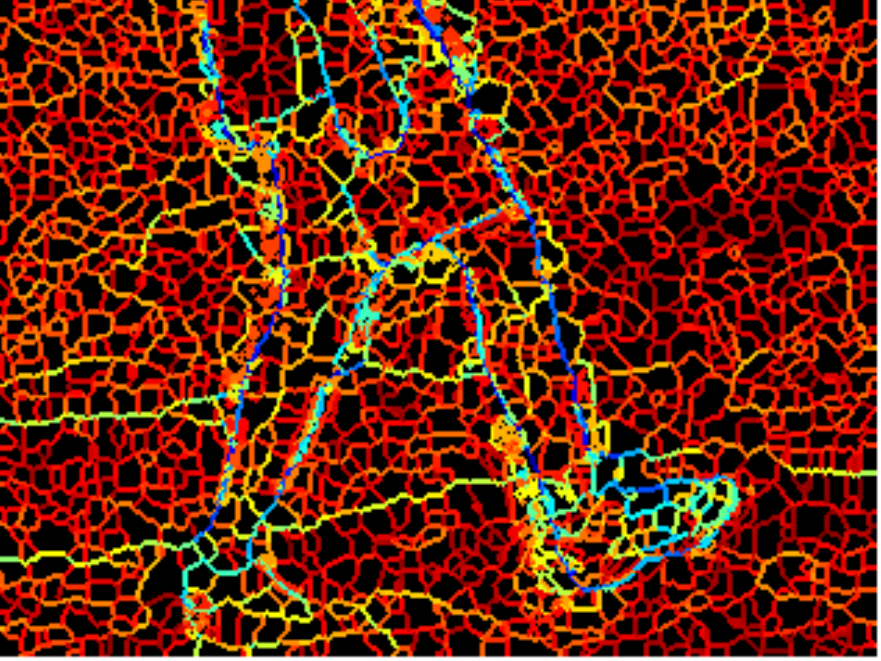}}   	                
               {\includegraphics[width=0.24\textwidth]{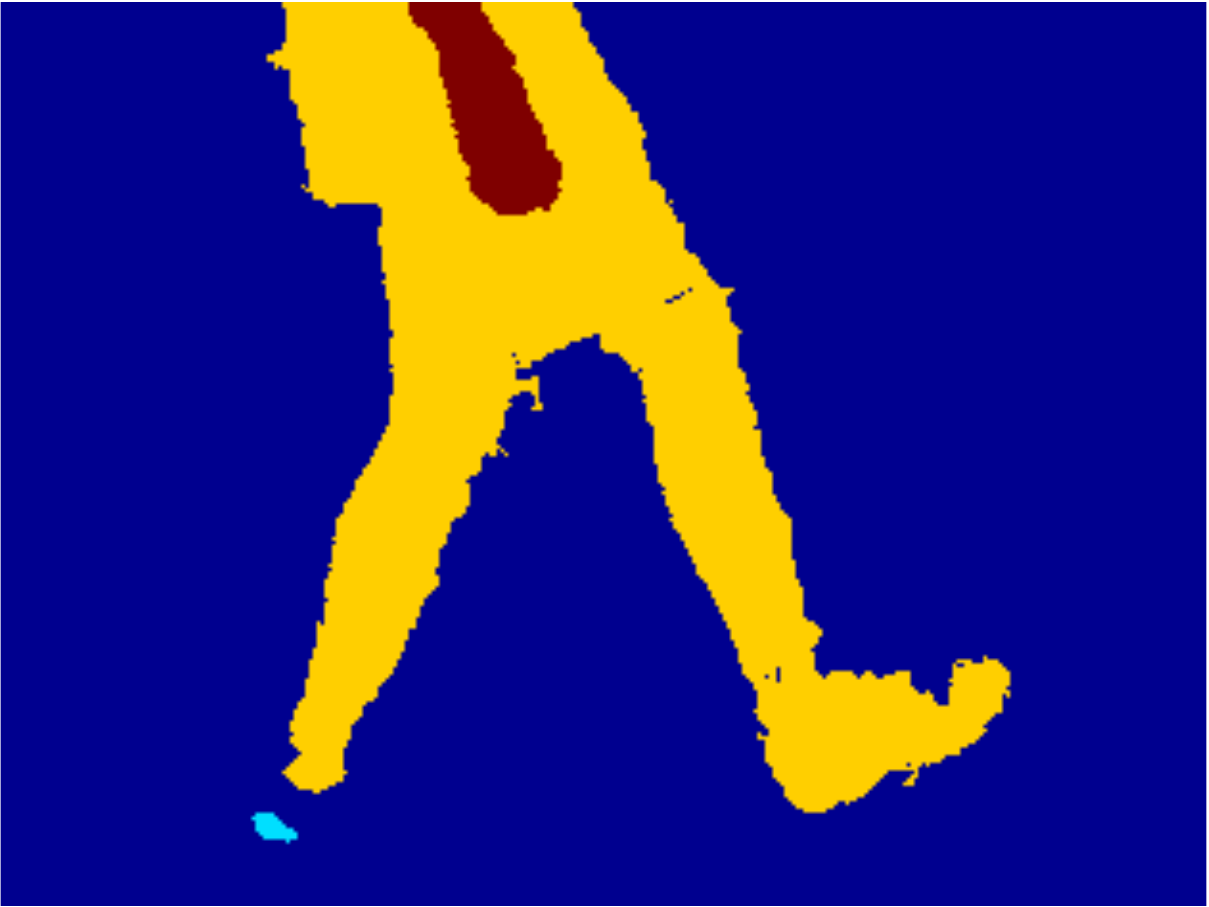}}   	                
			   {\includegraphics[width=0.24\textwidth]{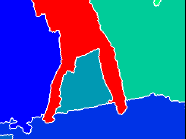}} \\
			   {\includegraphics[width=0.24\textwidth]{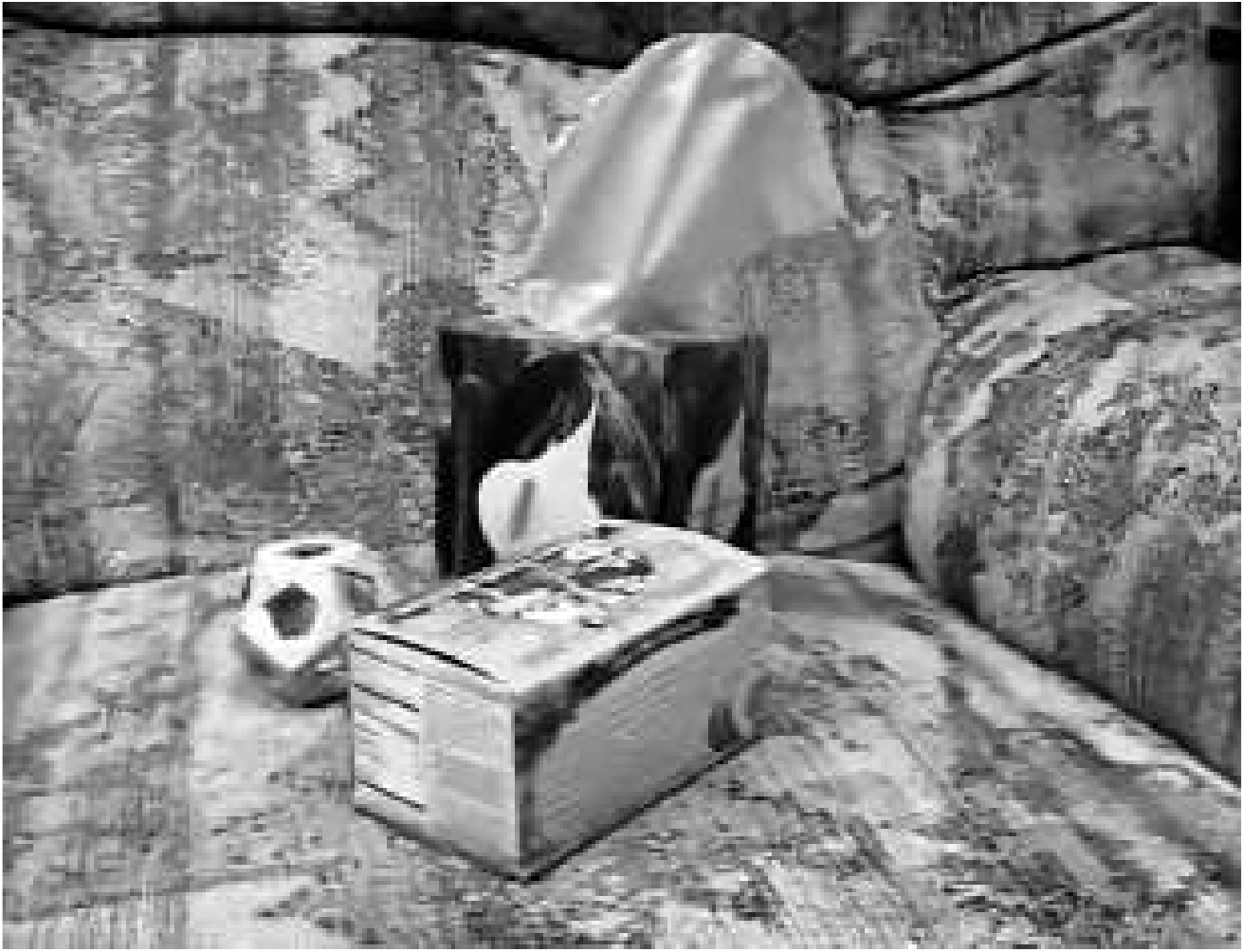}}      
               {\includegraphics[width=0.24\textwidth]{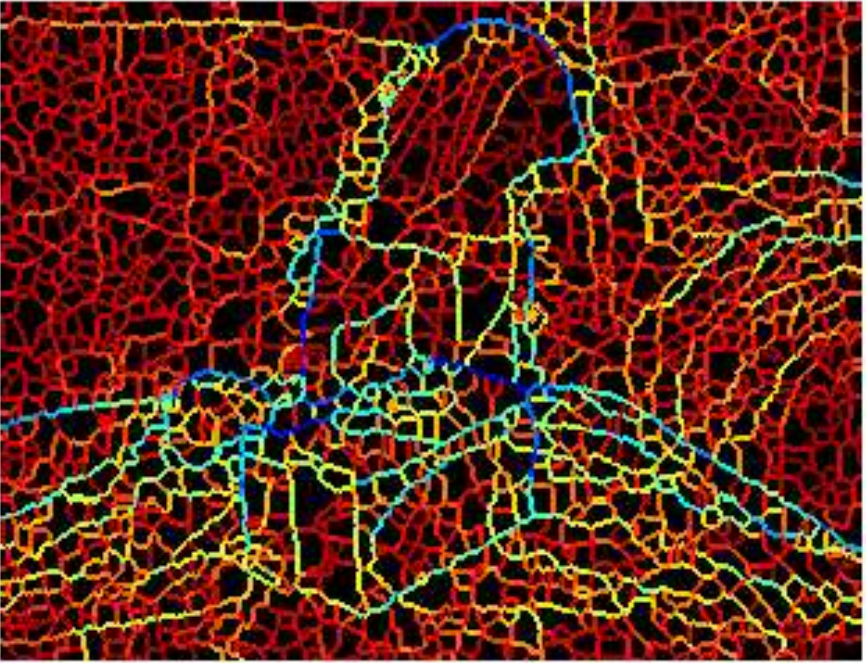}}   	                
               {\includegraphics[width=0.24\textwidth]{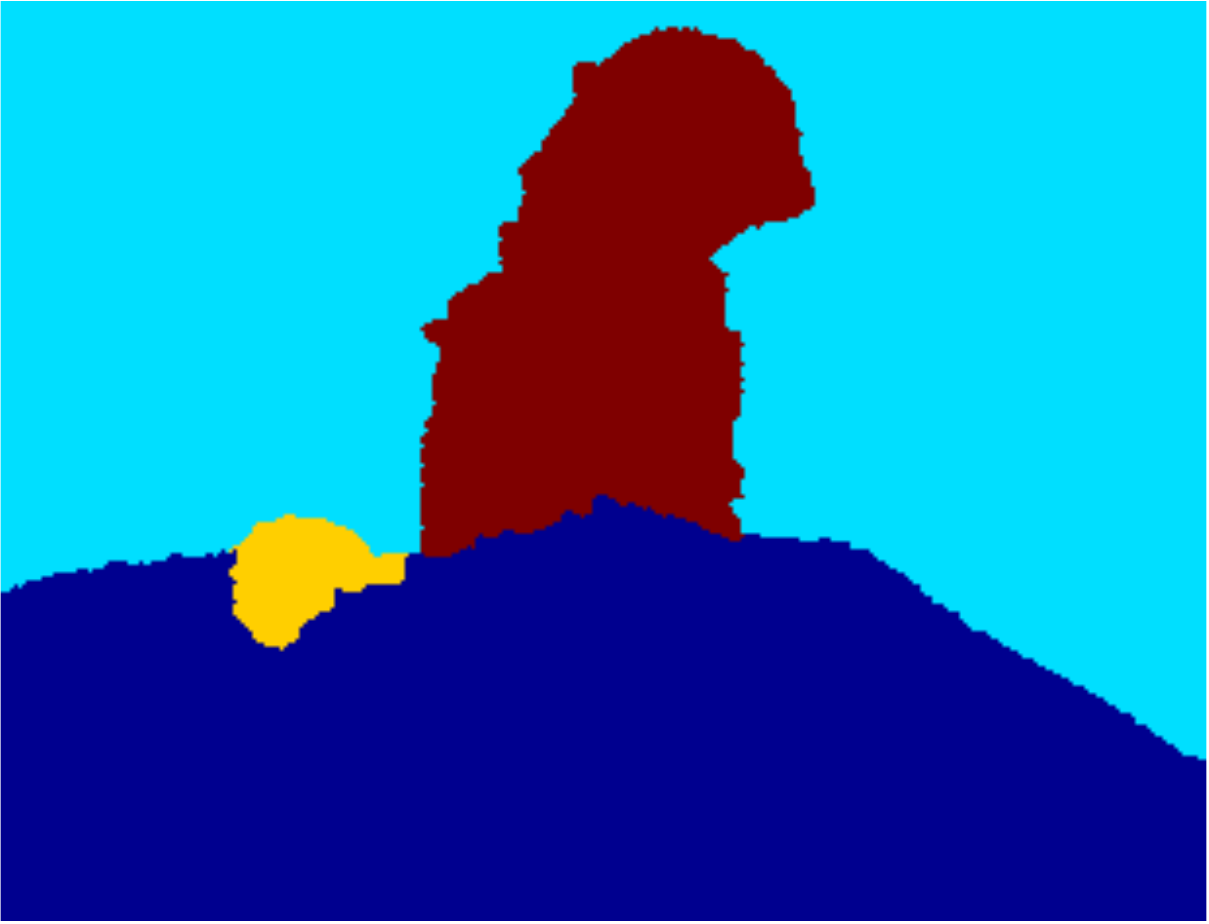}}
			   {\includegraphics[width=0.24\textwidth]{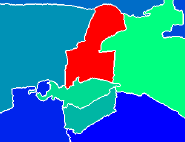}} \\
			   %\begin{subfloatrow}[4]
			%	\subfloat[1)]
			   {\includegraphics[width=0.24\textwidth]{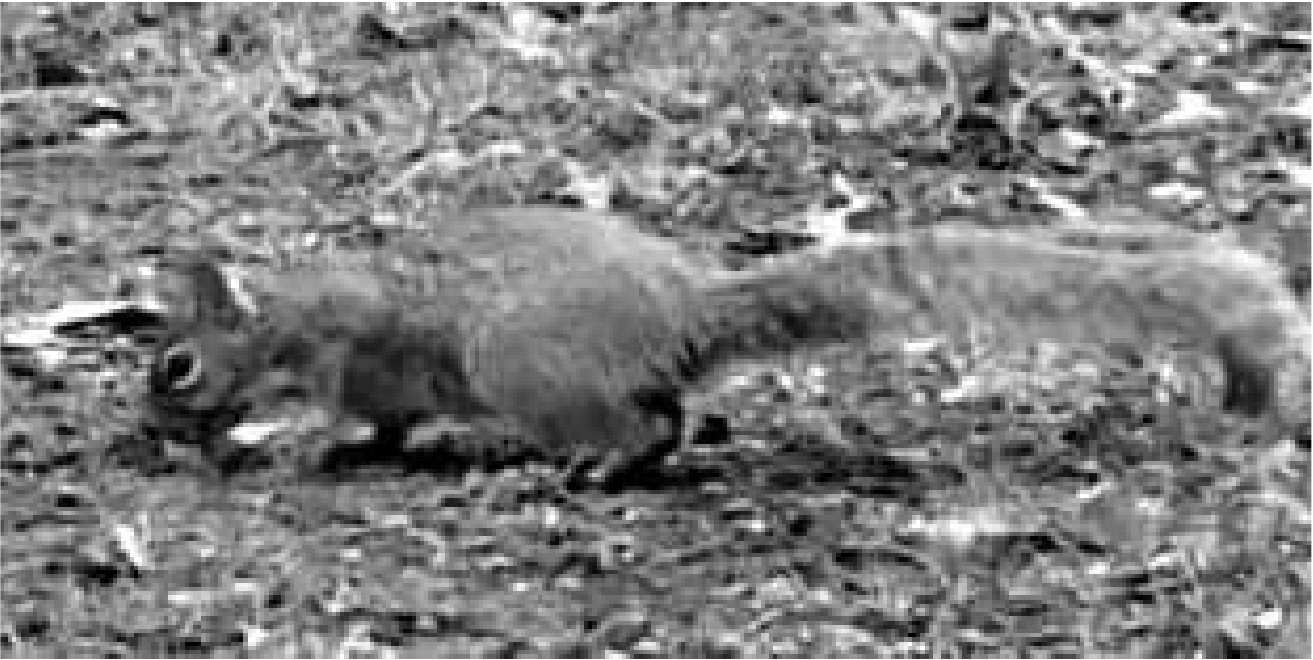}} 
             % \subfloat[2)]
  			   {\includegraphics[width=0.24\textwidth]{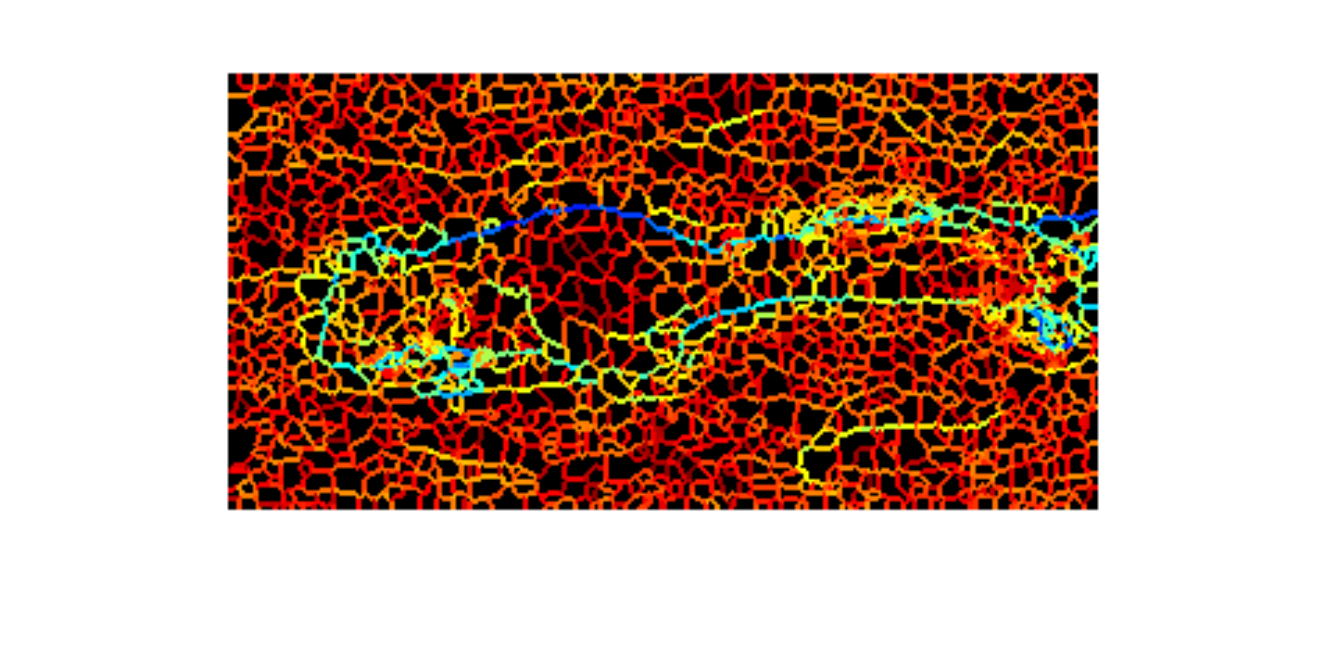}}   	                
              % \subfloat[3)] 
               {\includegraphics[width=0.24\textwidth]{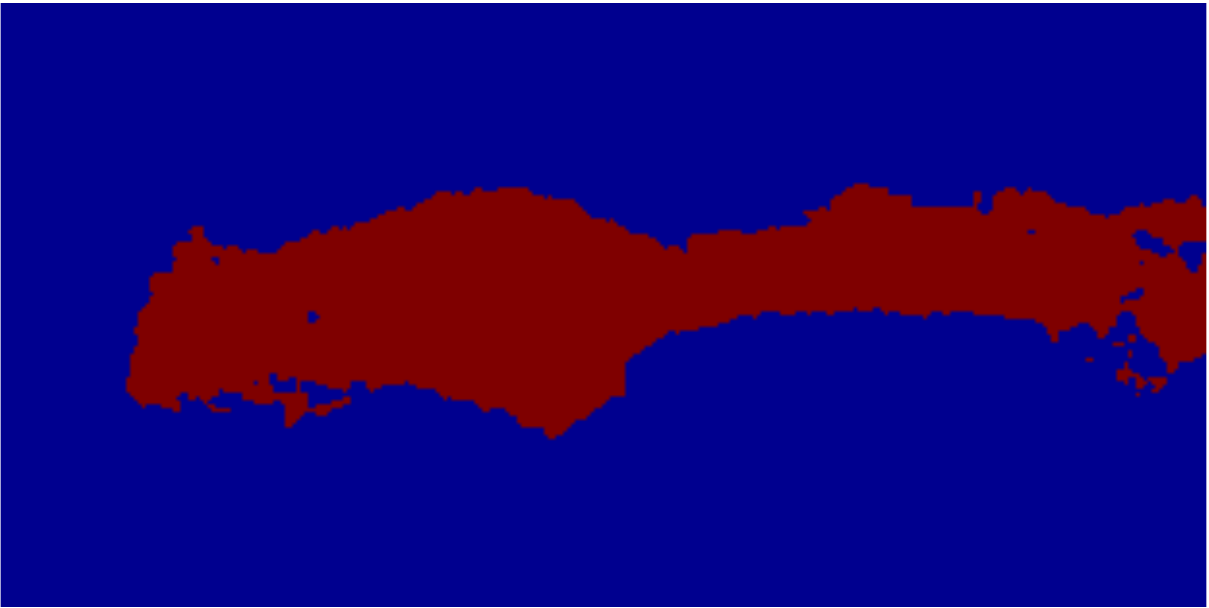}}
			 % \subfloat[4)]
			   {\includegraphics[width=0.24\textwidth]{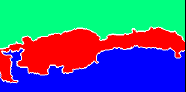}}  
			   %\end{subfloatro
              \end{tabular}
            }
      }
		\caption{\sl Sample frames from \emph{Walking Legs, Couch Color} and \emph{Squirrel4} (first column), affinities between superpixels on the reference frame  (second column), output of our algorithm (third column) and the result of the segmentation method described in \cite{steinSH08} (fourth column) (Figures at fourth column are borrowed from \cite{steinSH08}) under Andrew Stein's authorization.'. }		
		 \label{fig-compare}
\end{figure*}

\begin{figure*}[bth!]
  \centerline 
      {
        \hbox
            {%\\
               \begin{tabular}{cccc}
               {\includegraphics[width=0.24\textwidth]{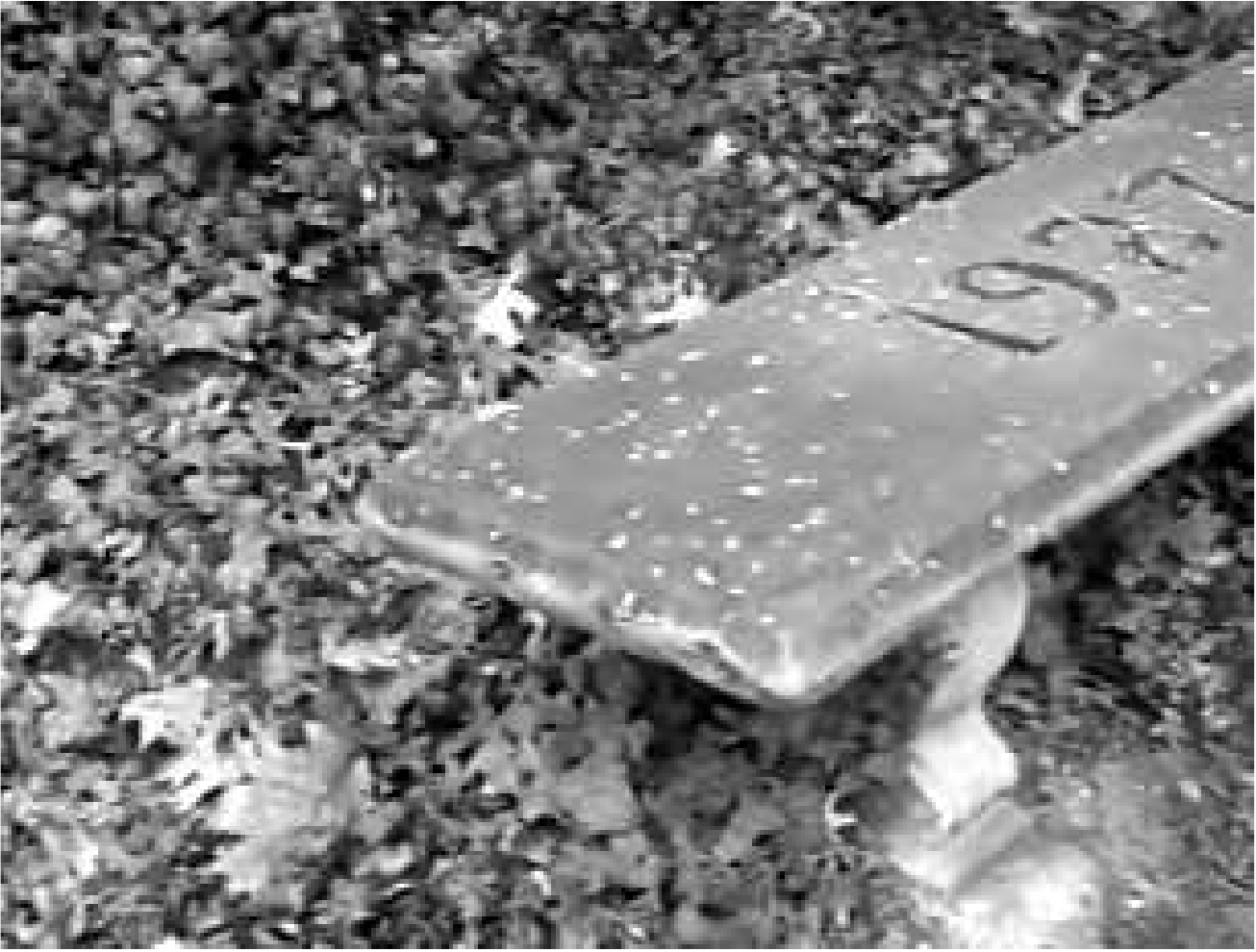}}   	                
               {\includegraphics[width=0.24\textwidth]{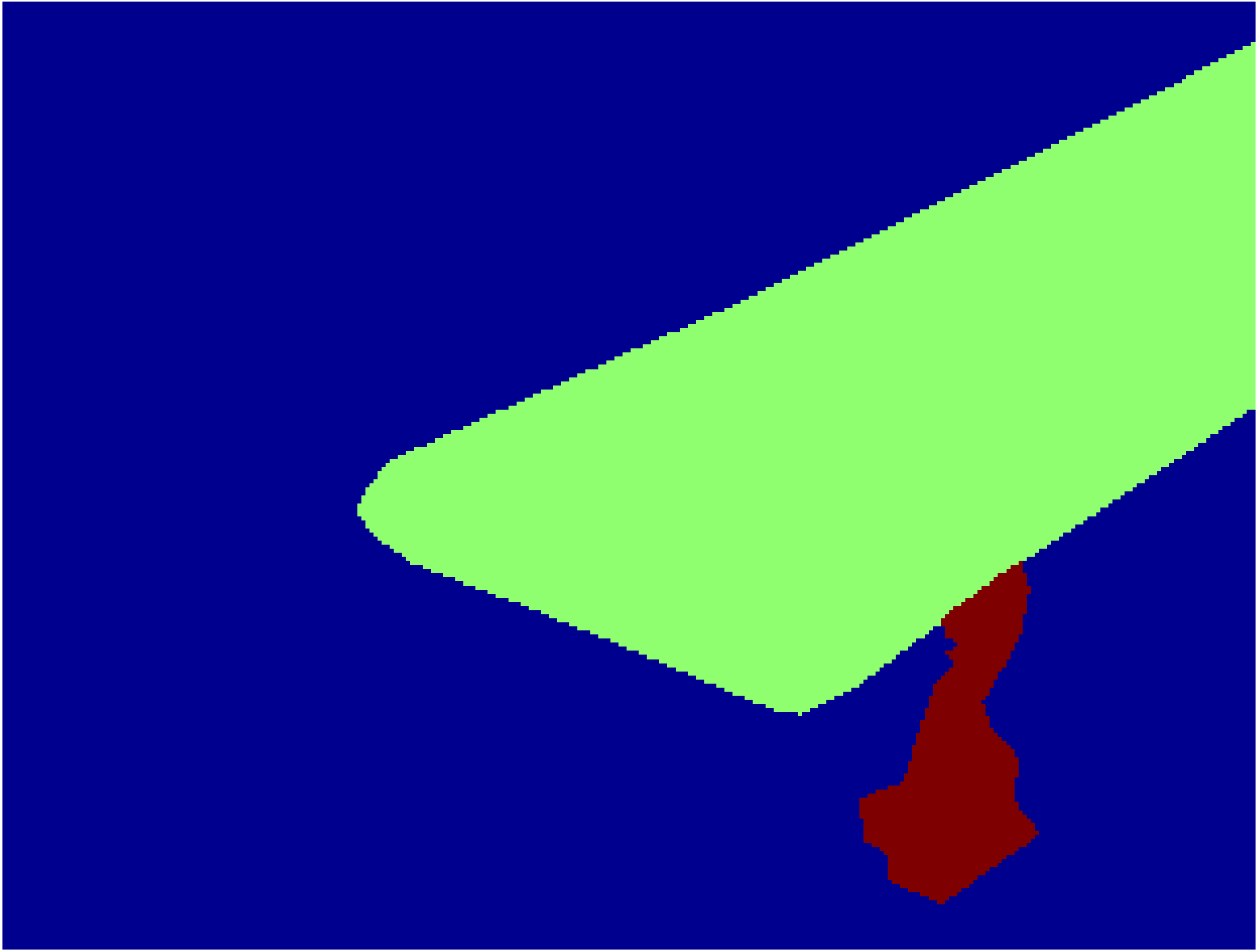}}
			   {\includegraphics[width=0.24\textwidth]{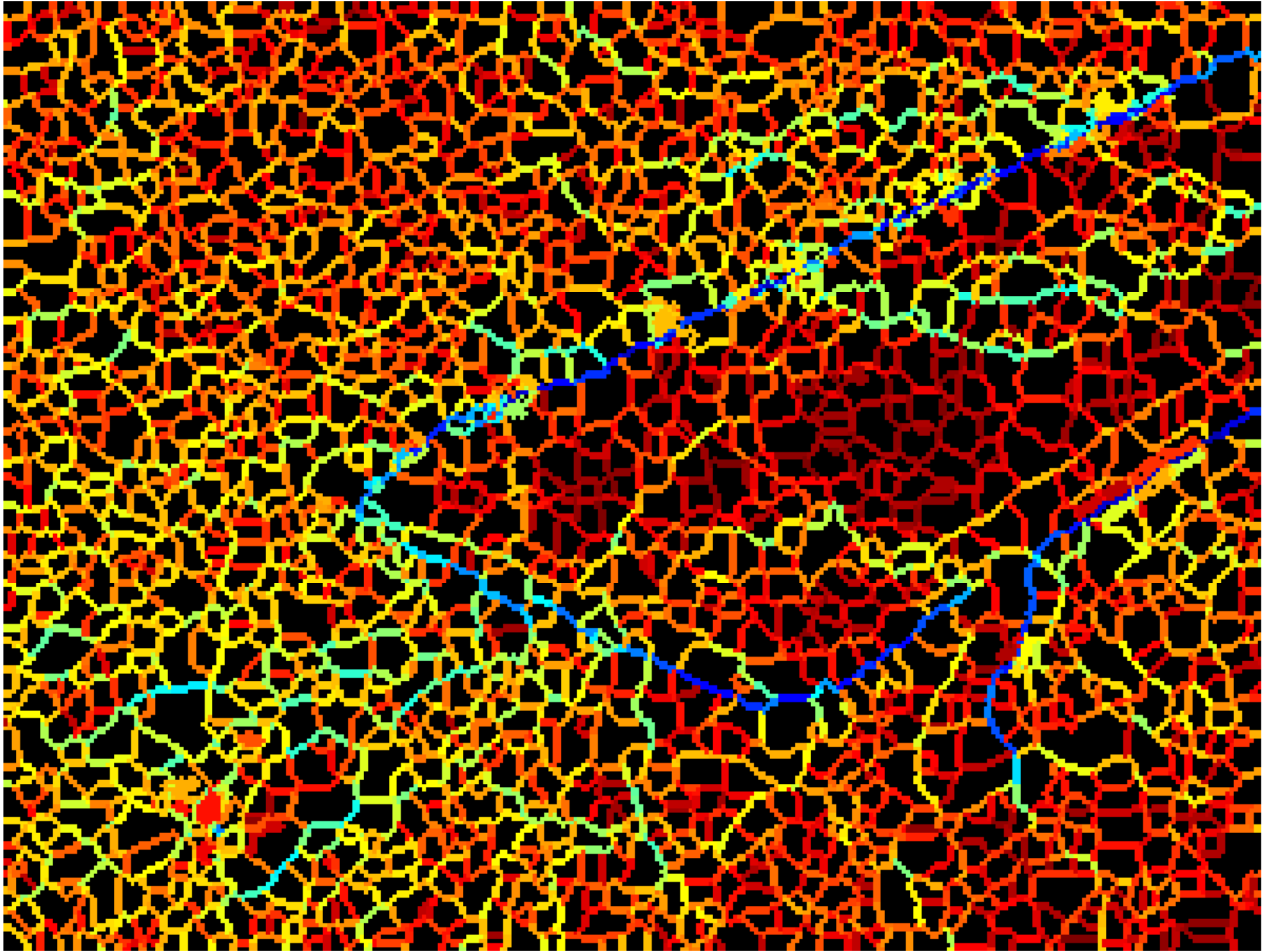}}   	                
               {\includegraphics[width=0.24\textwidth]{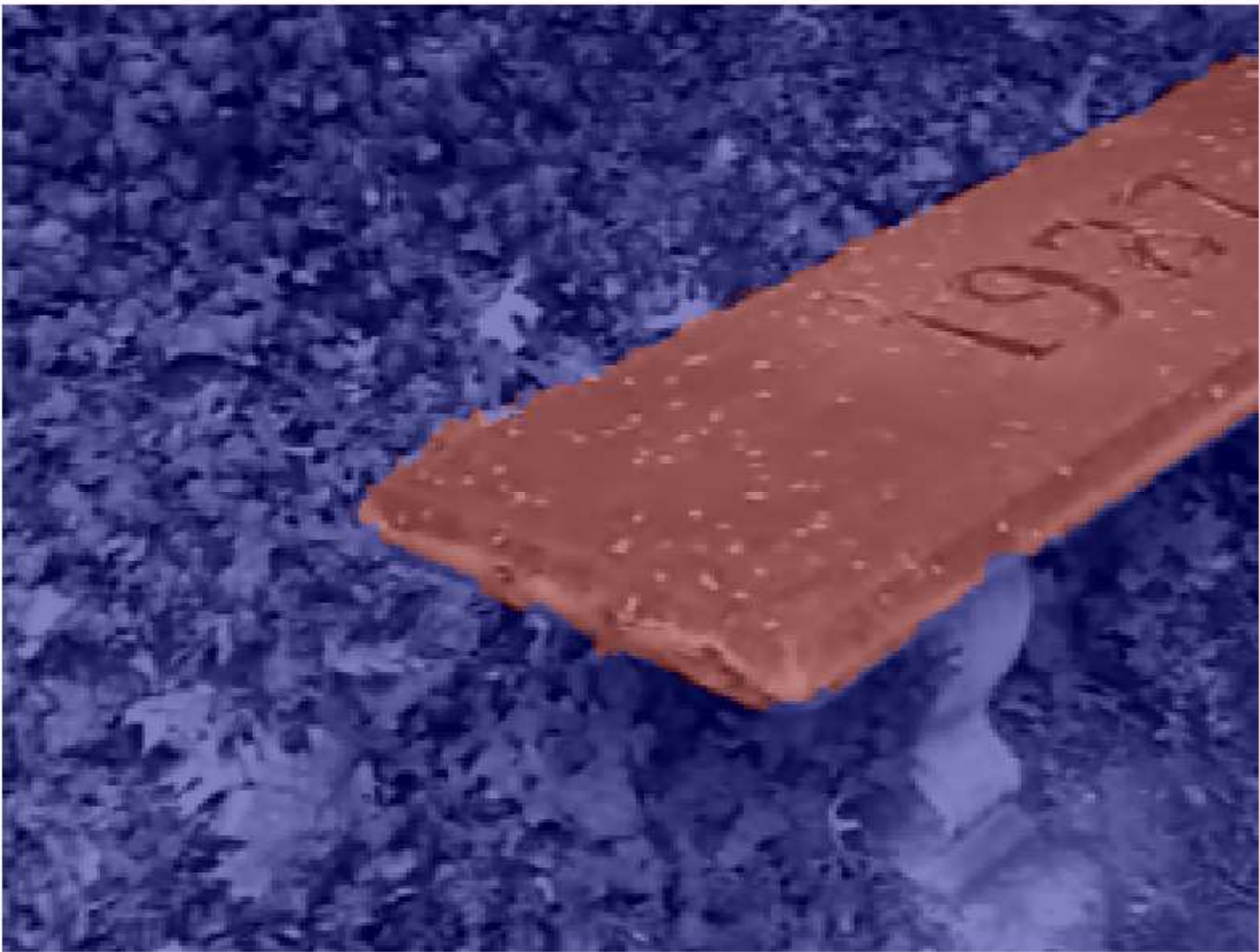}}   \\	                		
			   {\includegraphics[width=0.24\textwidth]{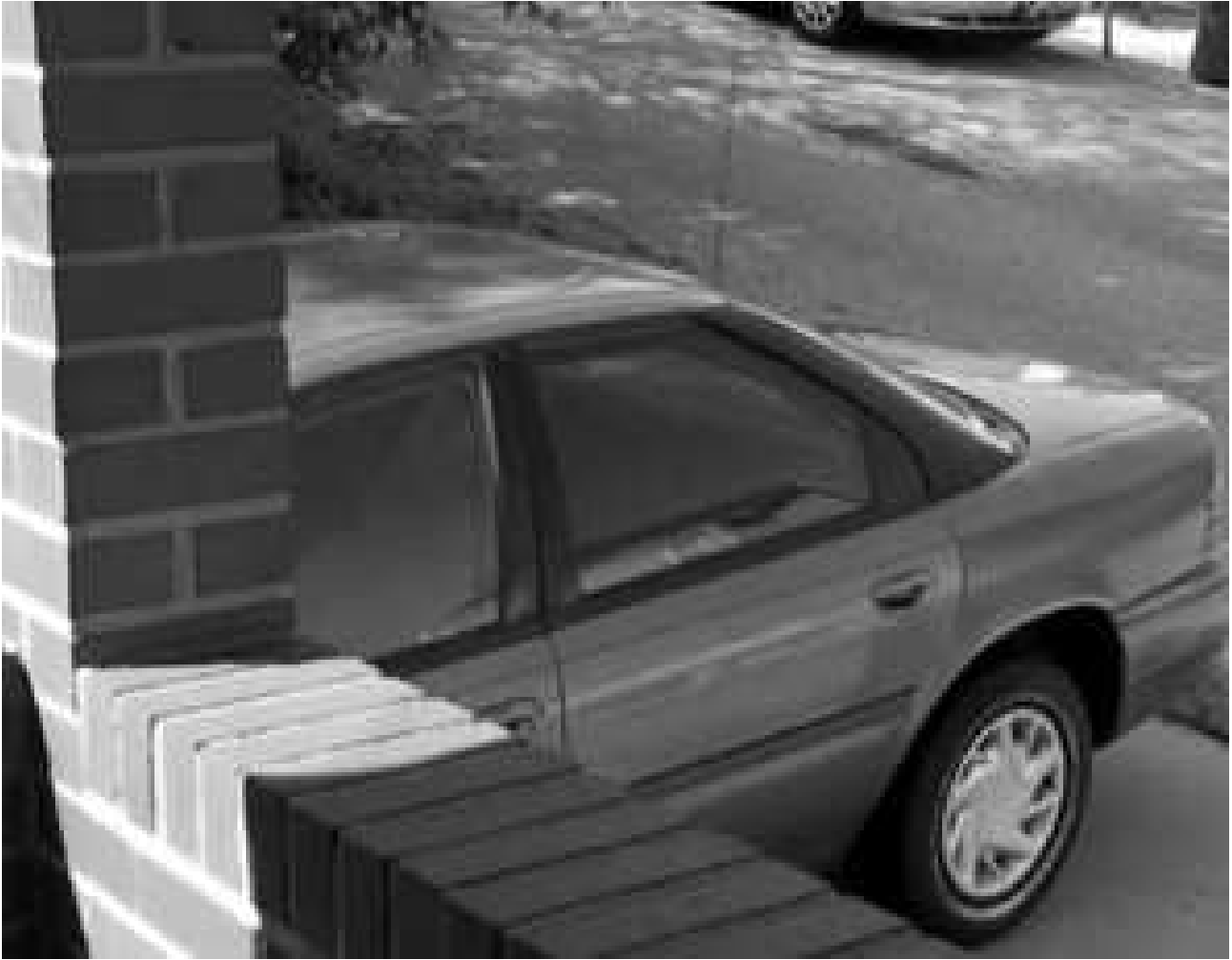}}   	                
               {\includegraphics[width=0.24\textwidth]{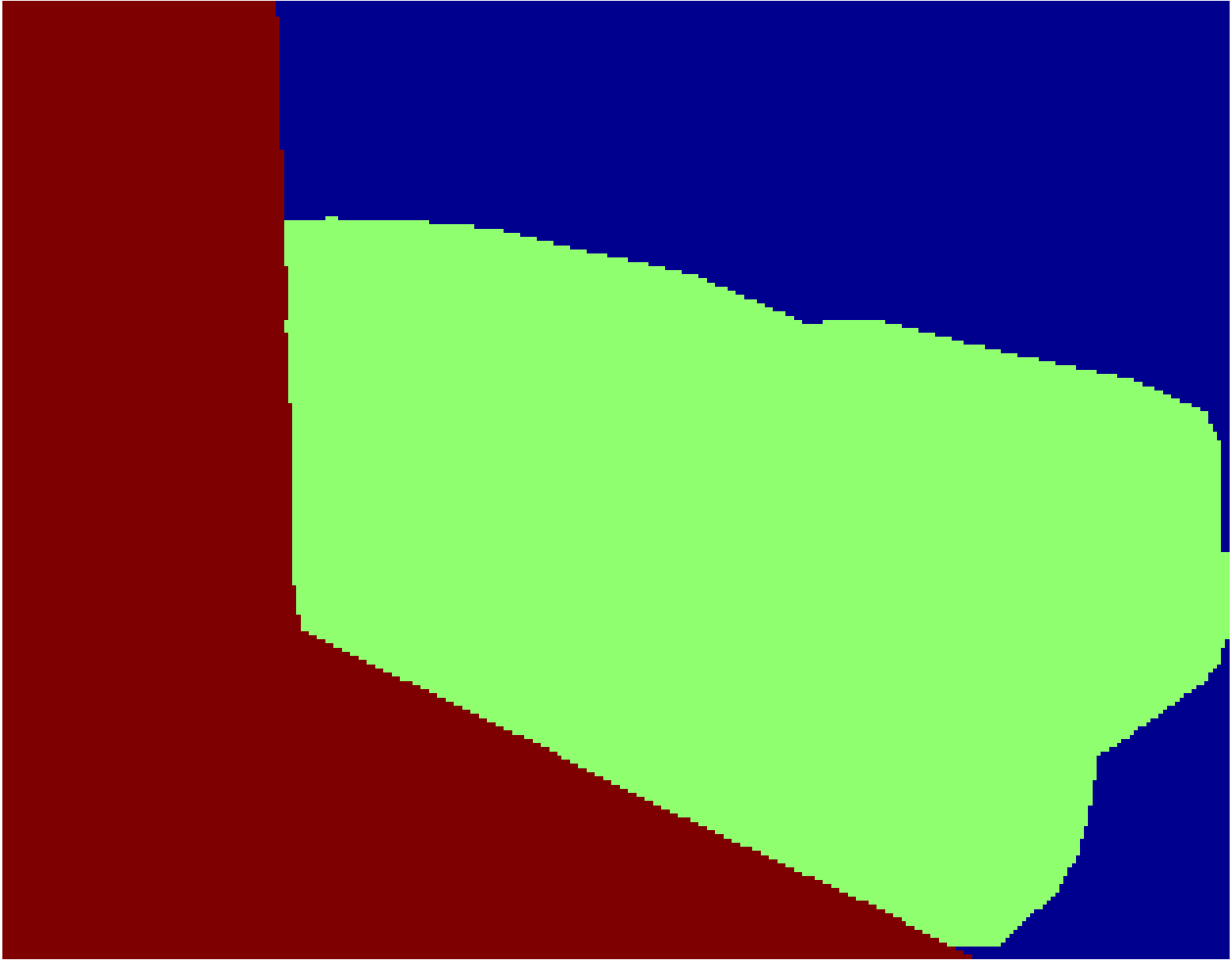}}
			   {\includegraphics[width=0.24\textwidth]{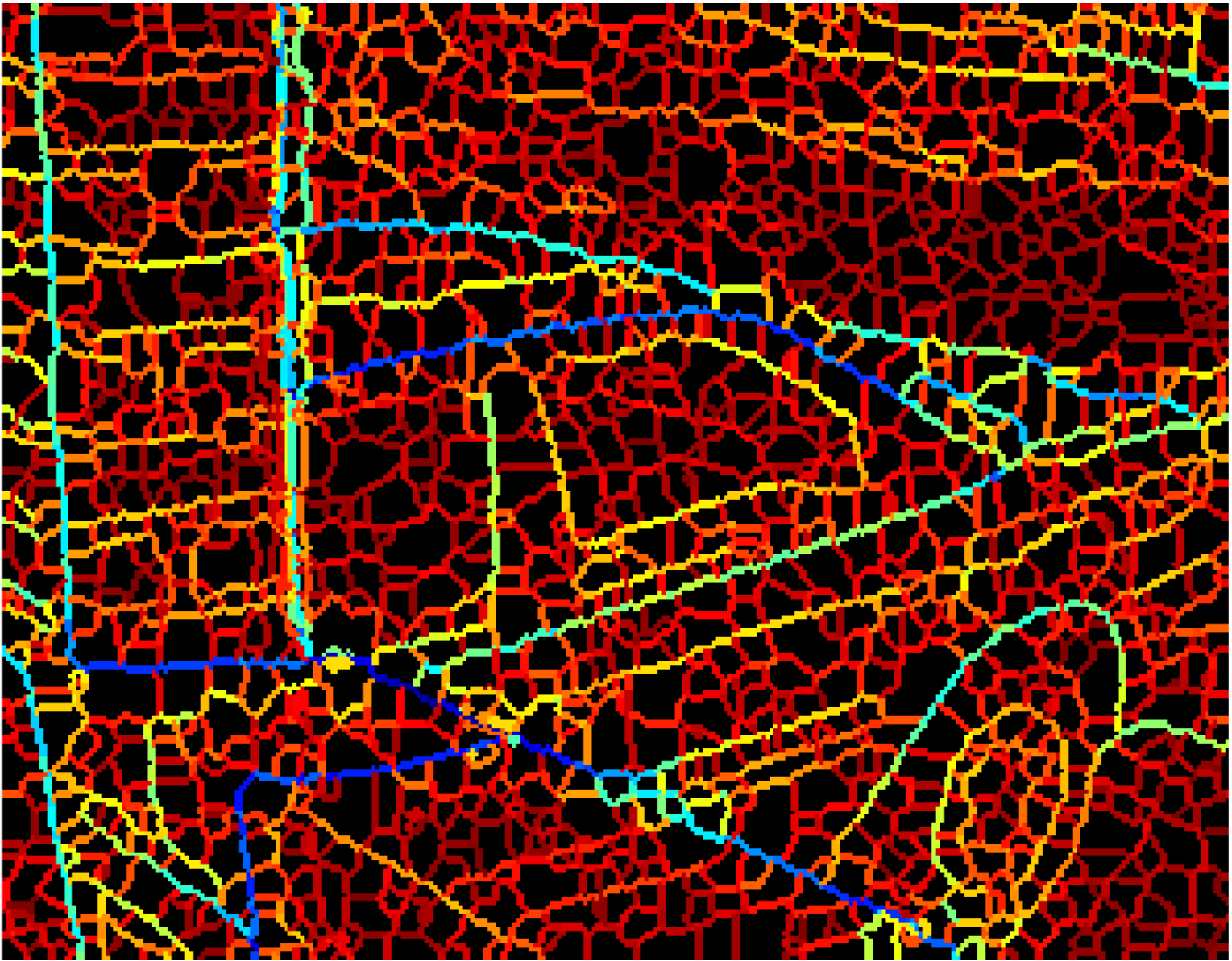}}   	                
               {\includegraphics[width=0.24\textwidth]{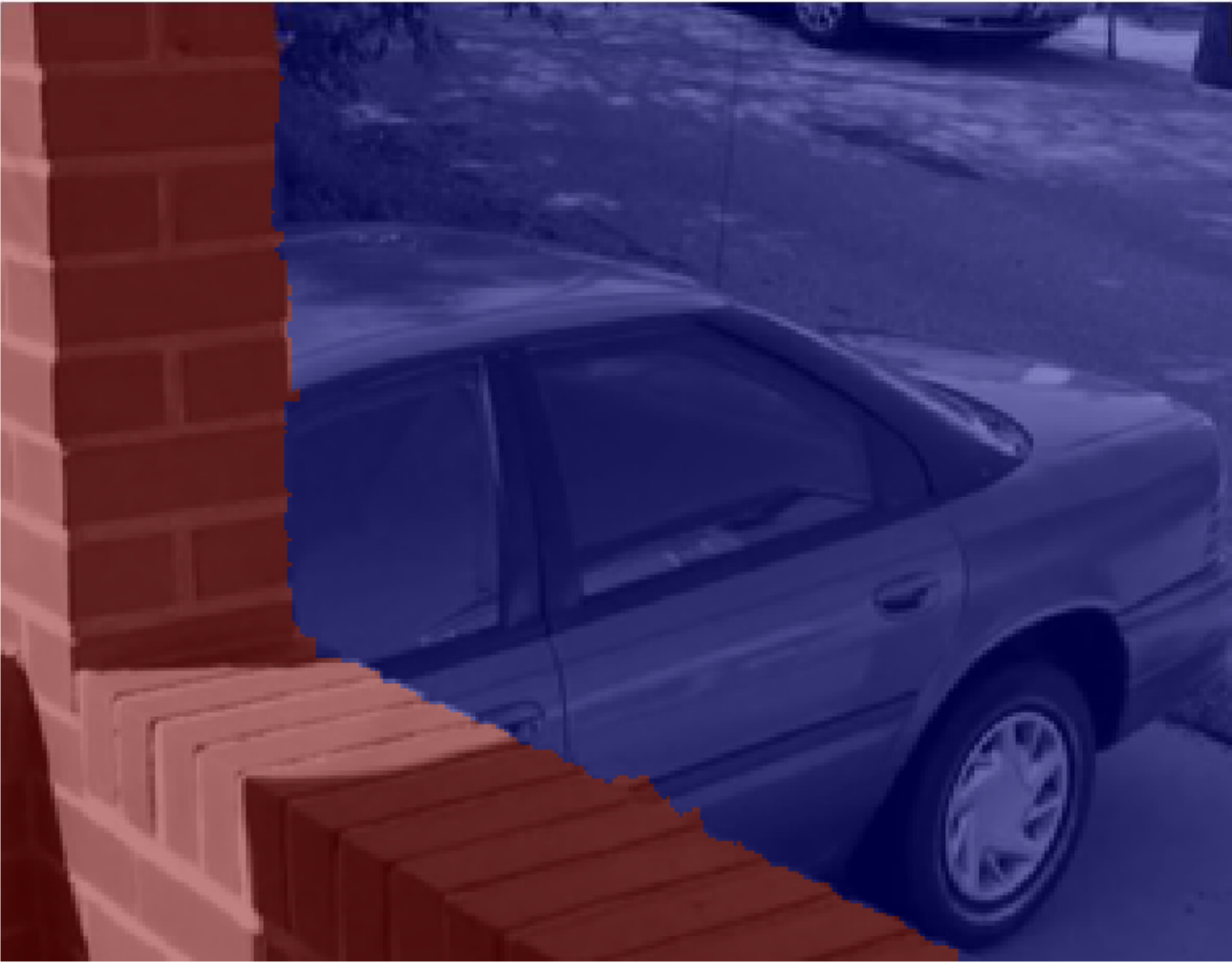}} \\
			   {\includegraphics[width=0.24\textwidth]{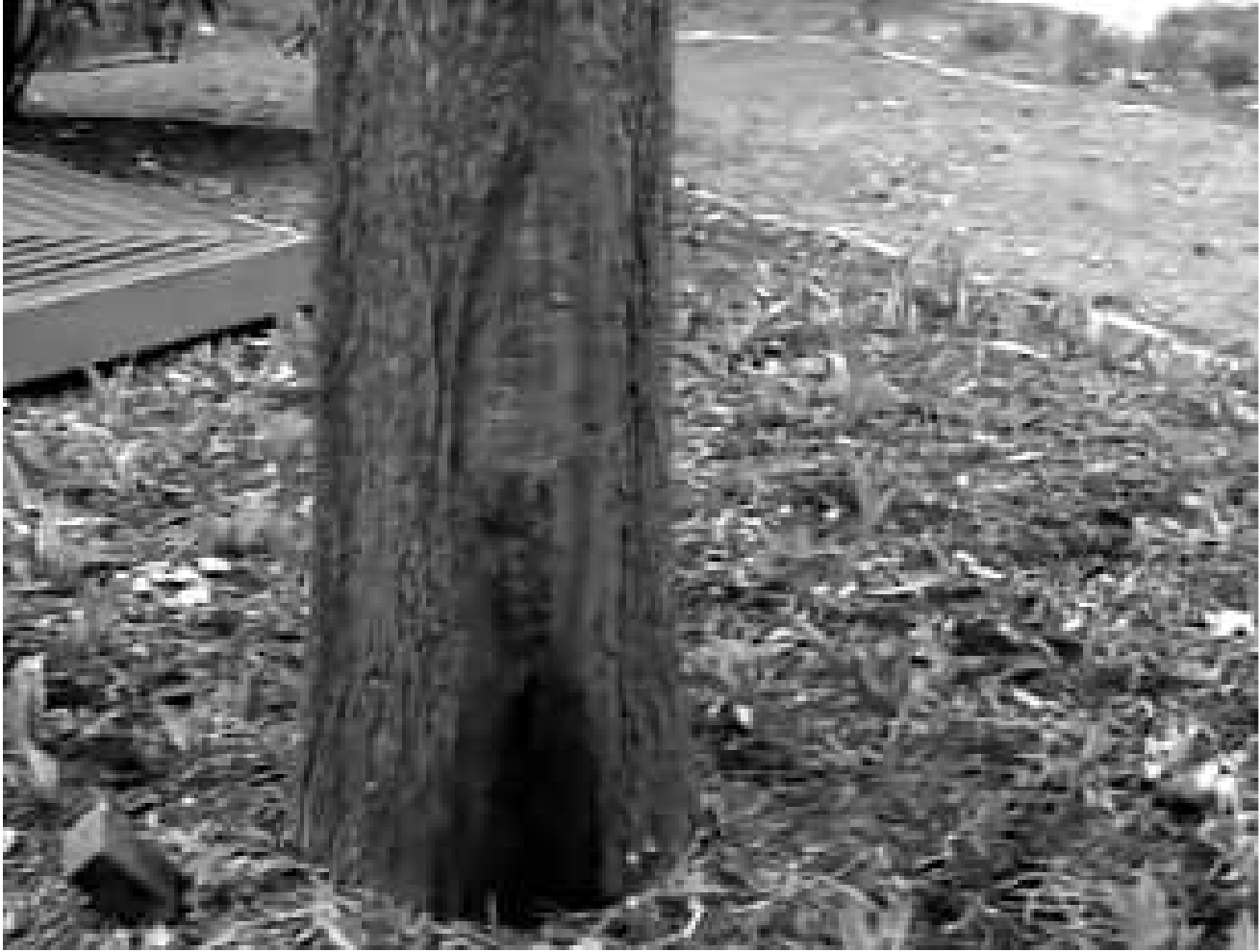}}   	                
               {\includegraphics[width=0.24\textwidth]{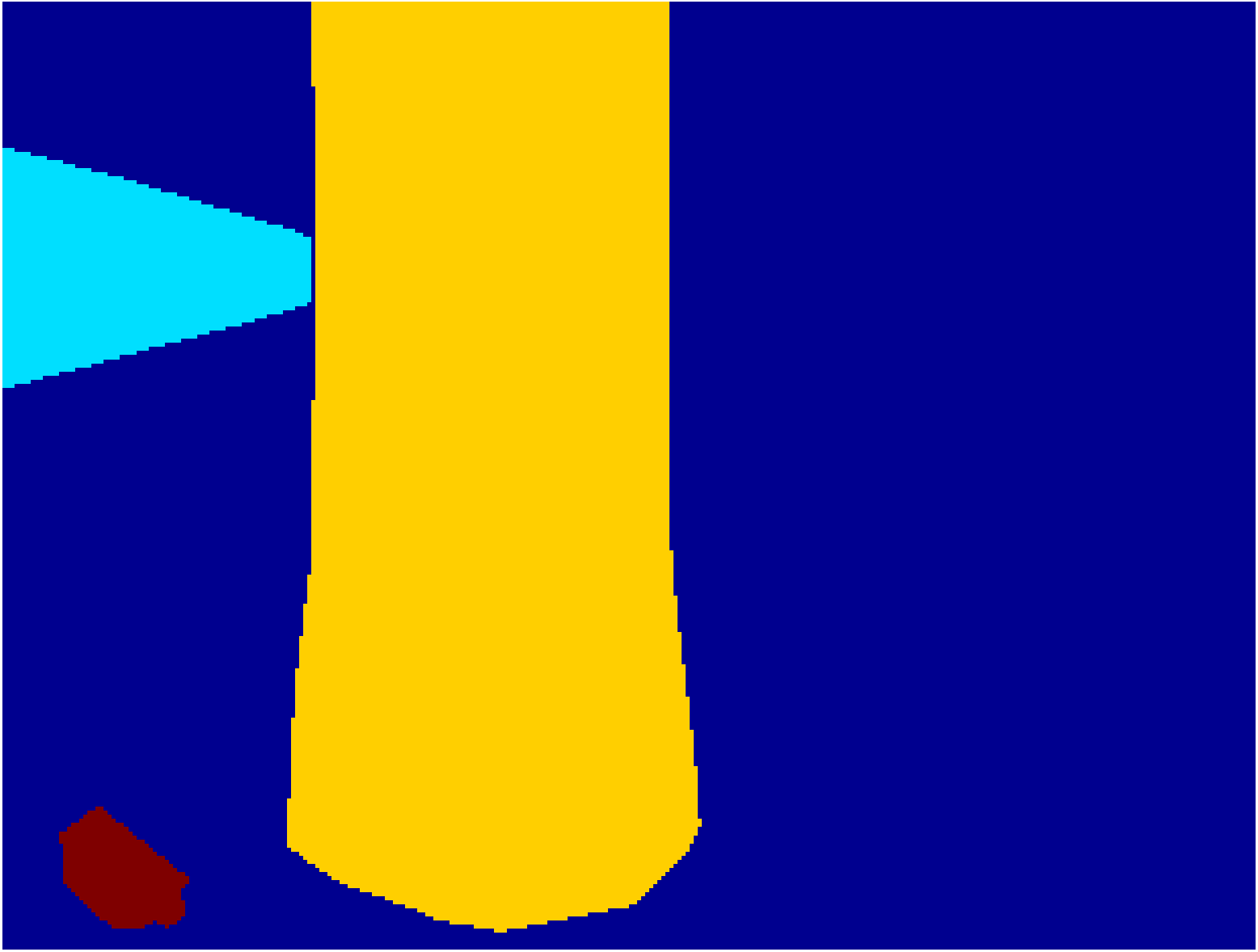}}
			   {\includegraphics[width=0.24\textwidth]{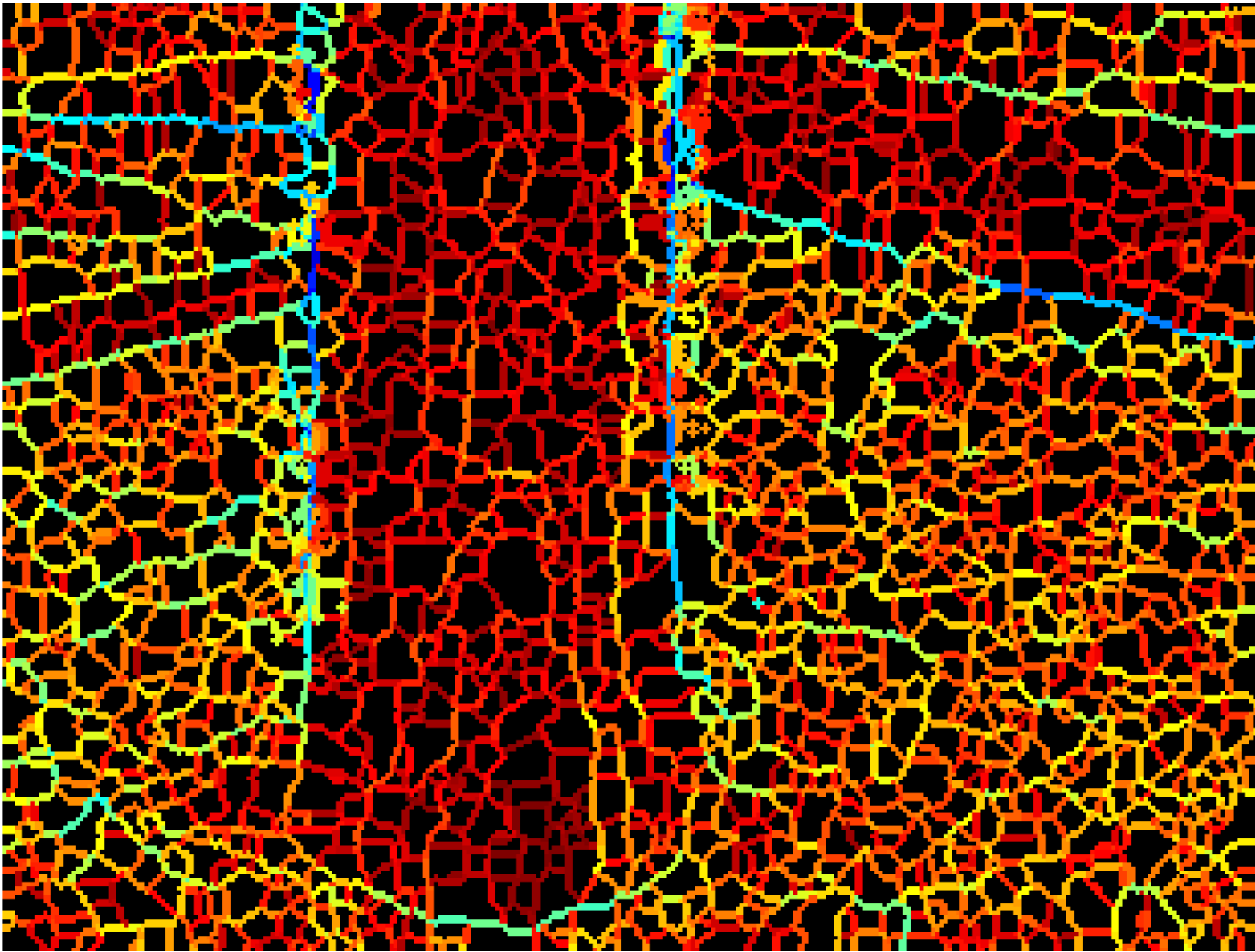}}   	                
               {\includegraphics[width=0.24\textwidth]{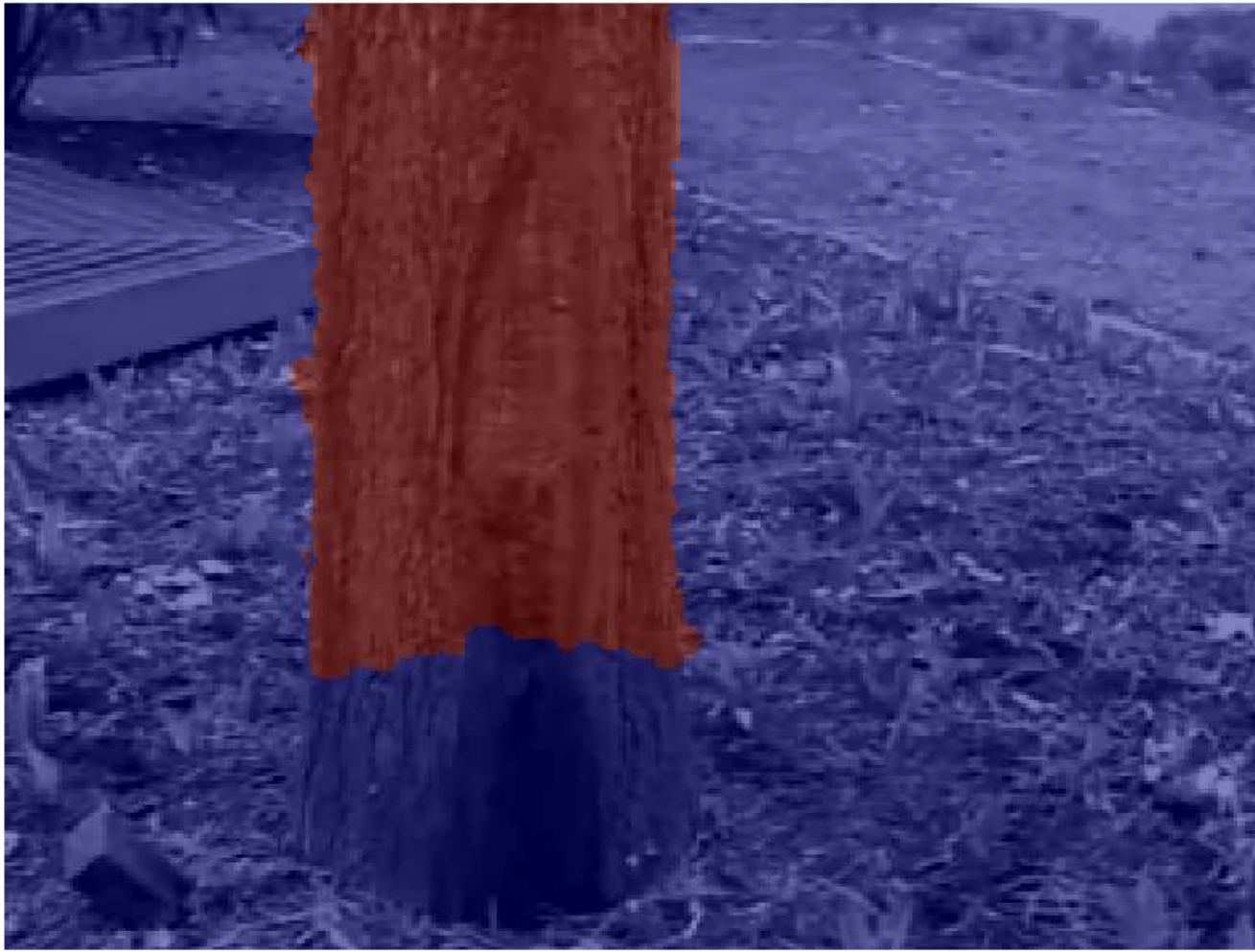}} \\
			   {\includegraphics[width=0.24\textwidth]{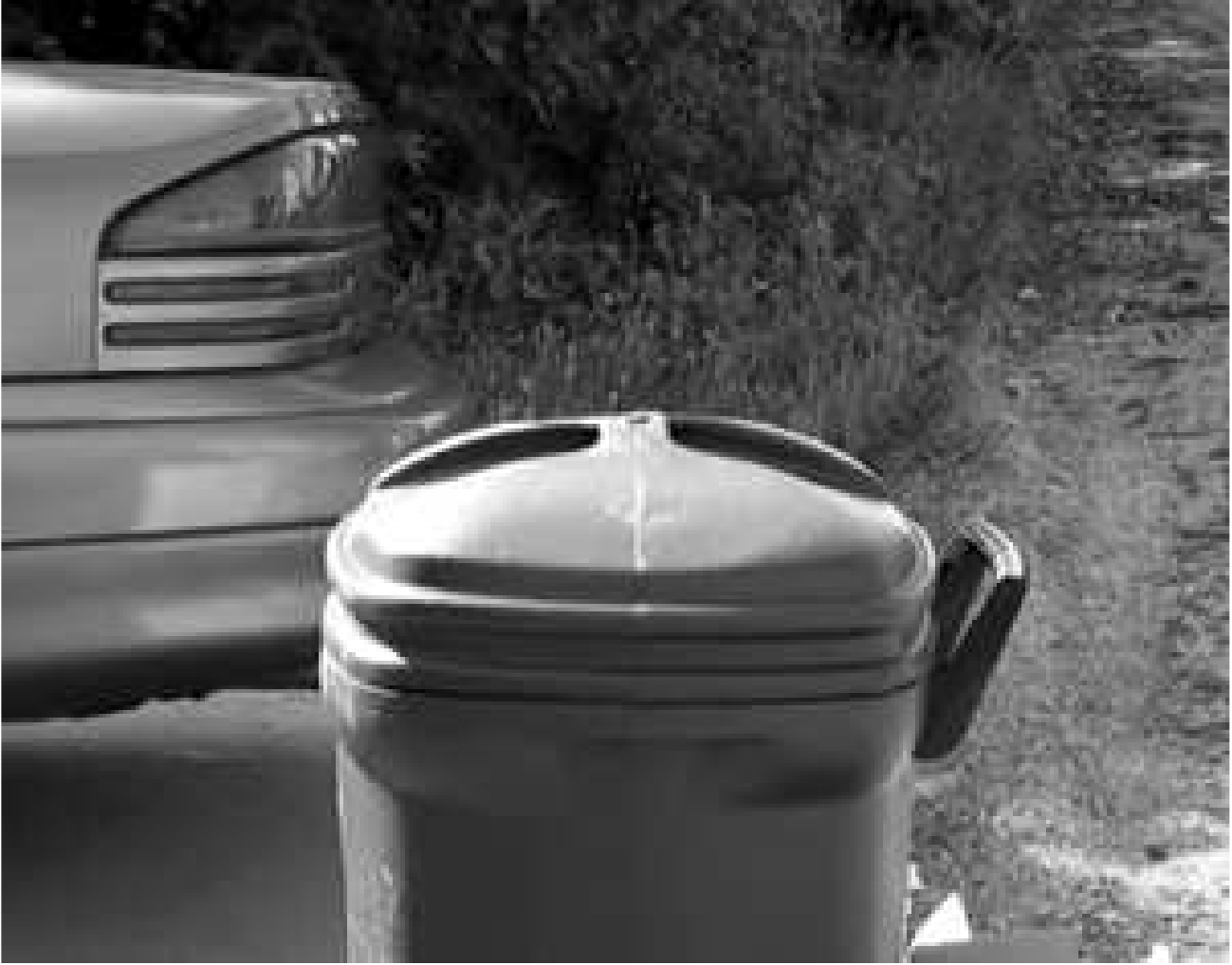}}   	                
               {\includegraphics[width=0.24\textwidth]{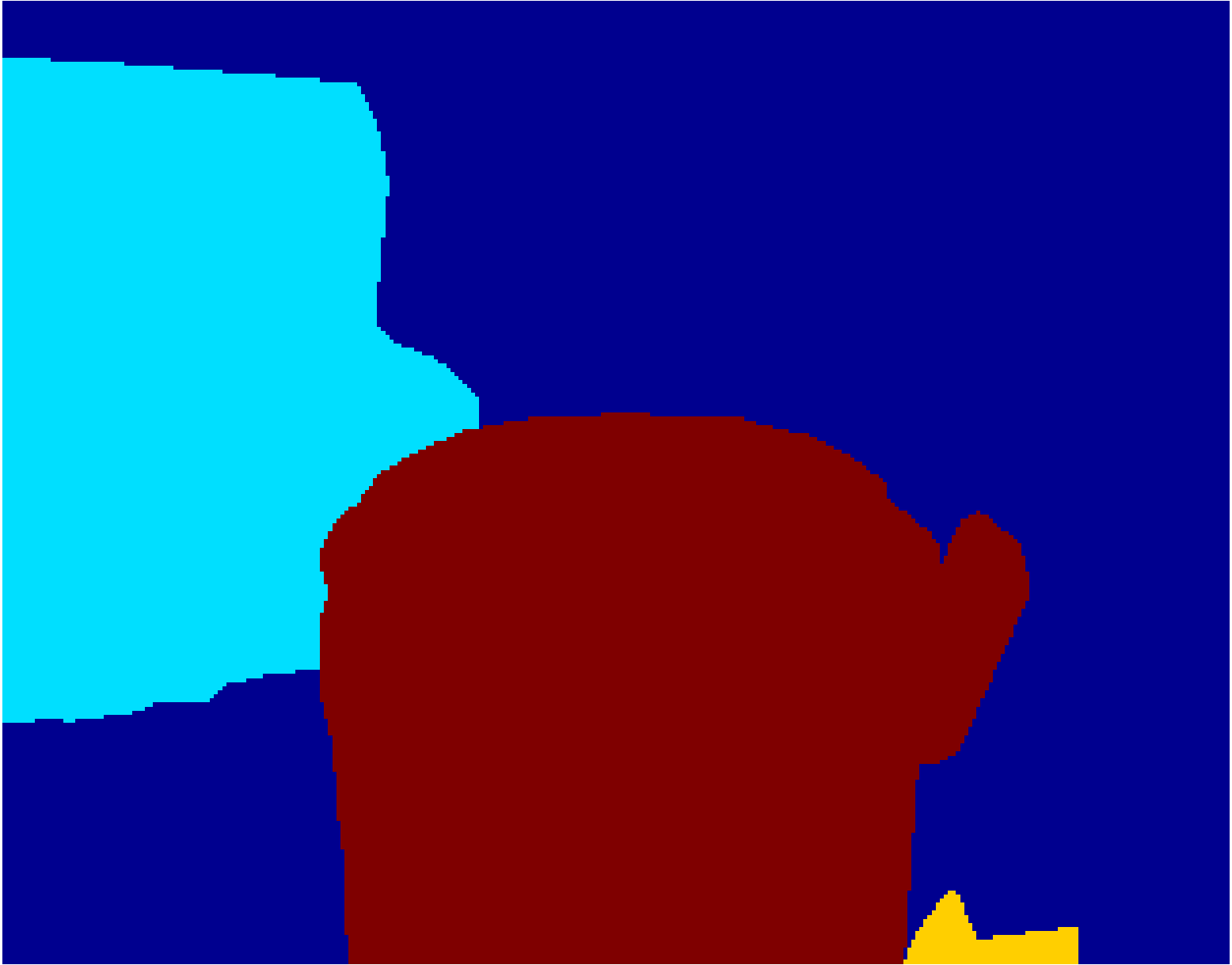}}
			   {\includegraphics[width=0.24\textwidth]{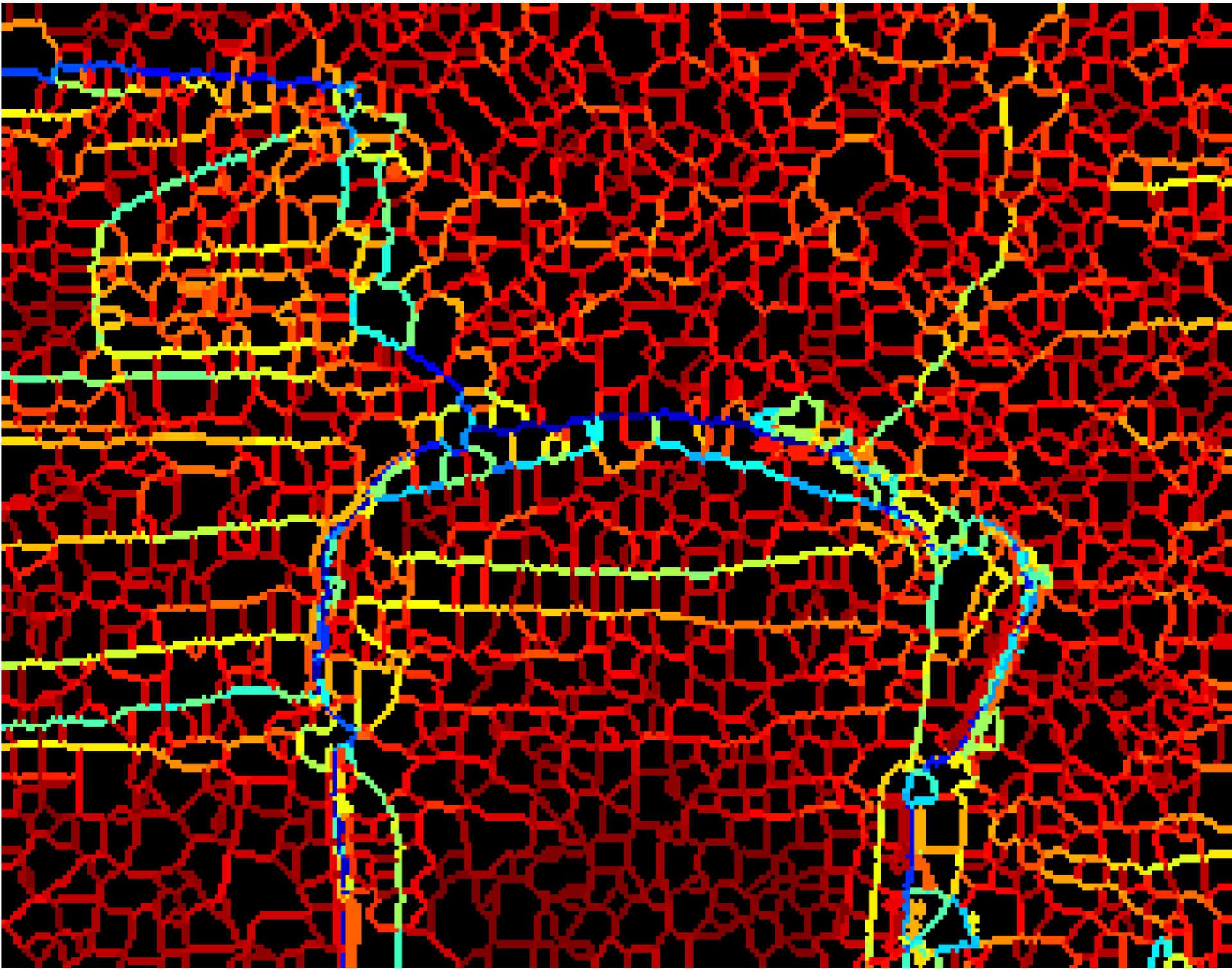}}   	                
               {\includegraphics[width=0.24\textwidth]{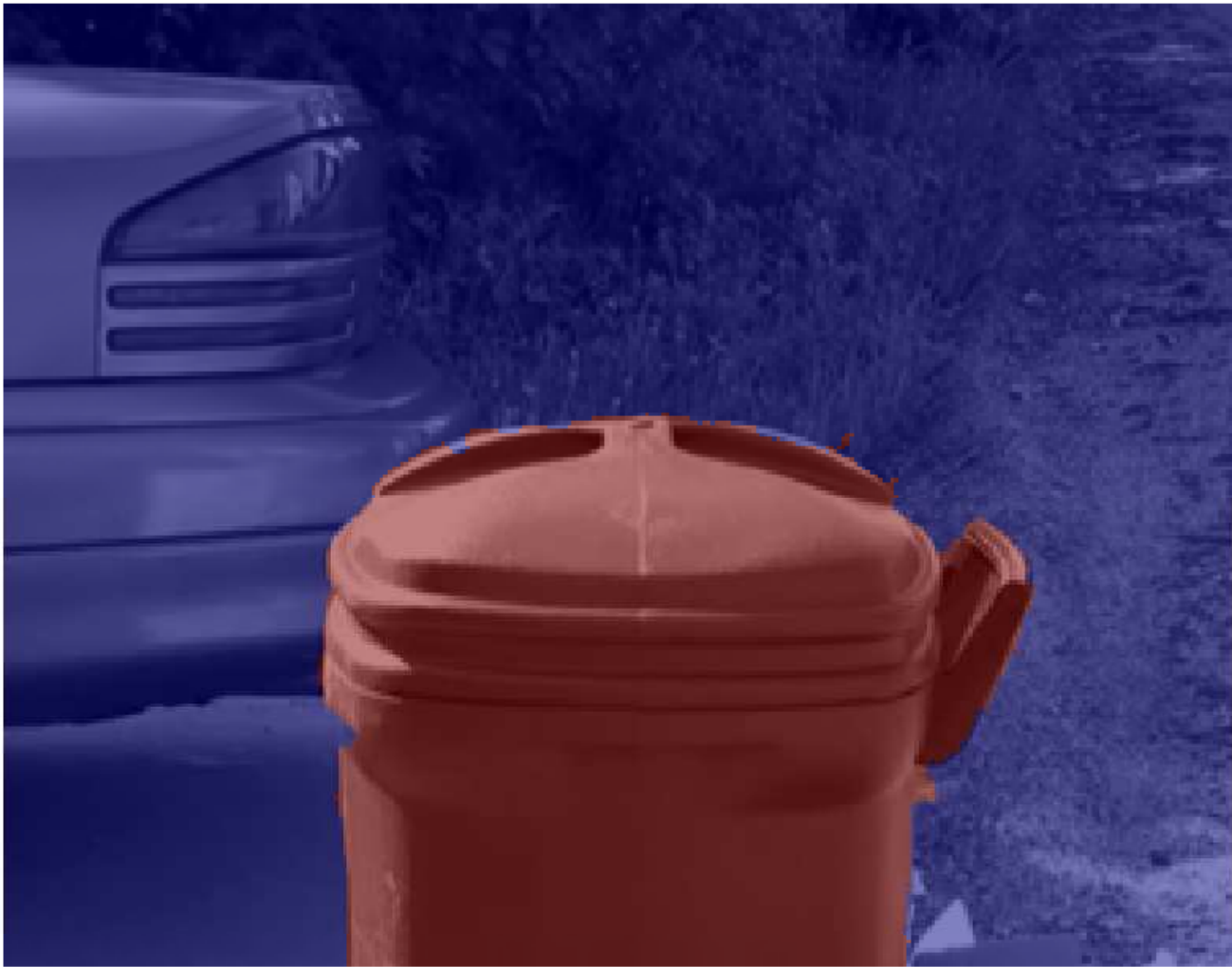}}\\
			   {\includegraphics[width=0.24\textwidth]{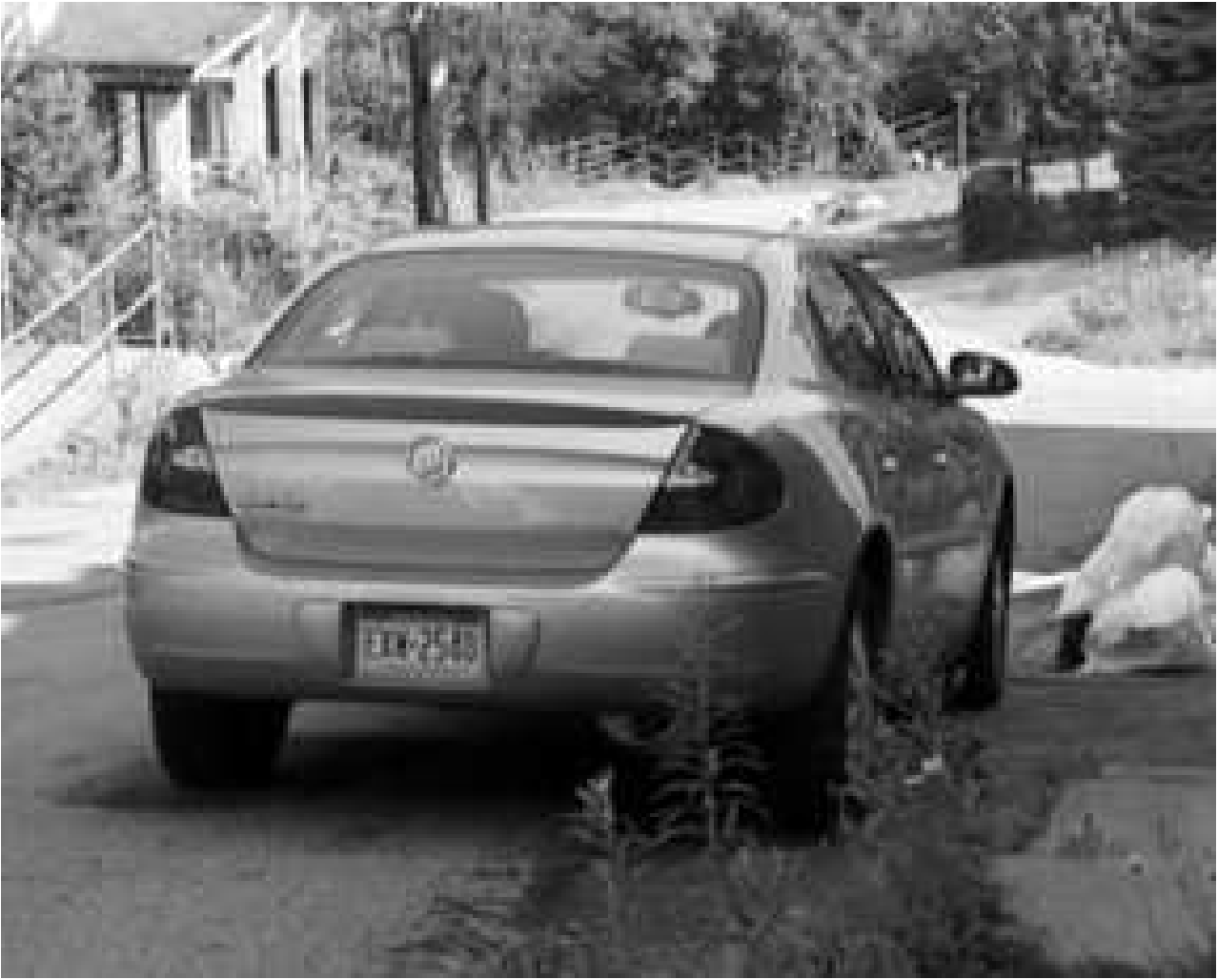}}   	                
               {\includegraphics[width=0.24\textwidth]{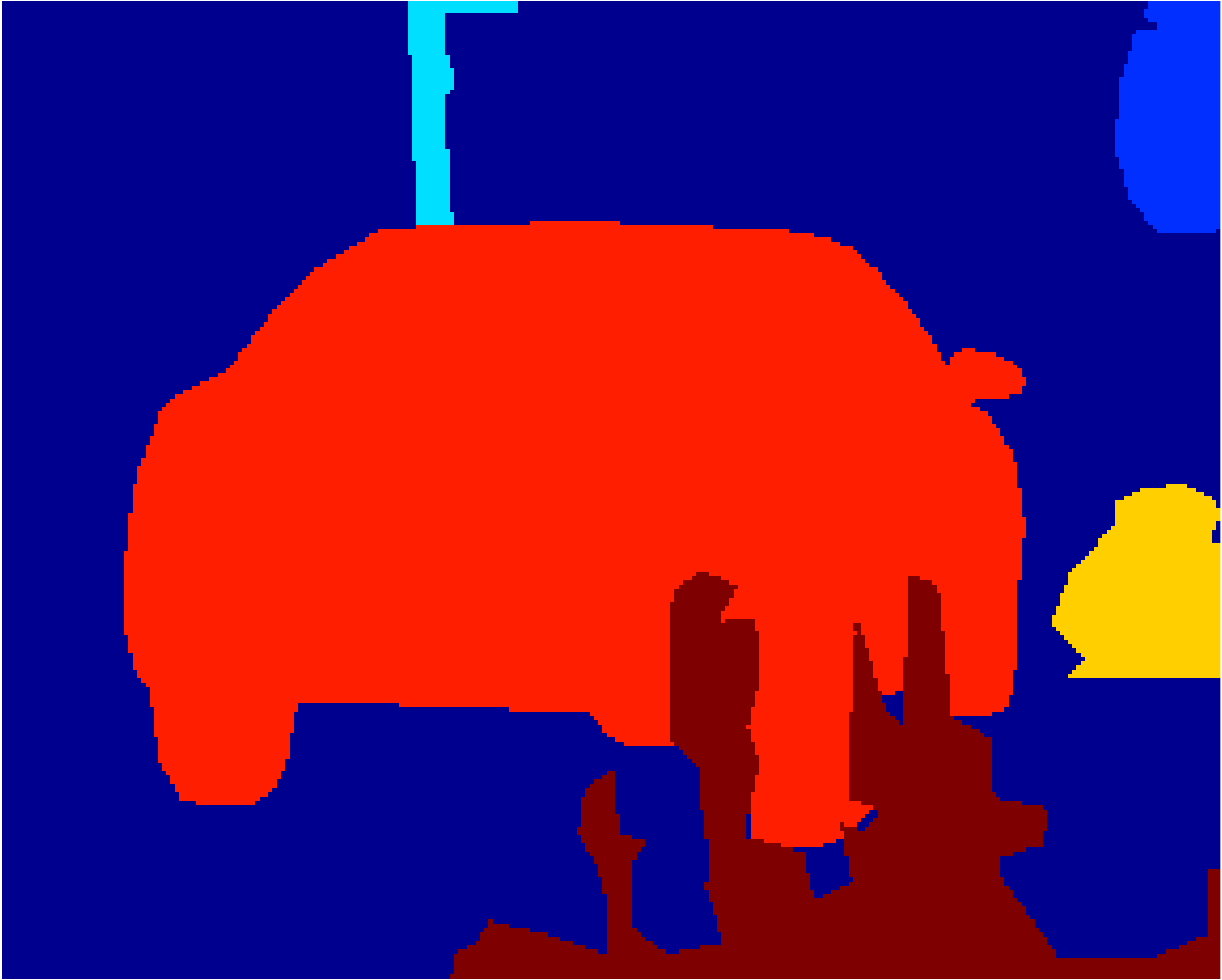}}
			   {\includegraphics[width=0.24\textwidth]{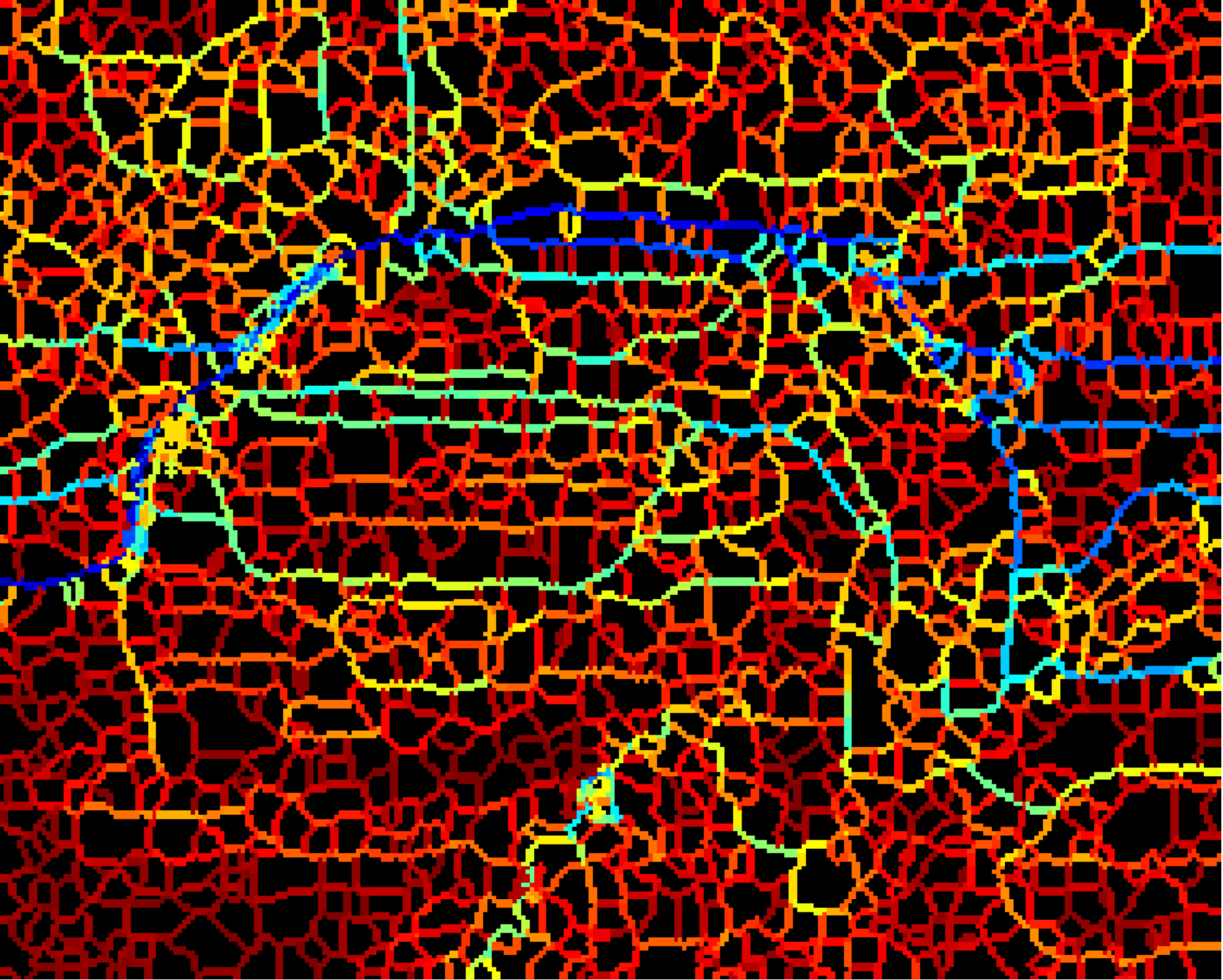}}   	                
               {\includegraphics[width=0.24\textwidth]{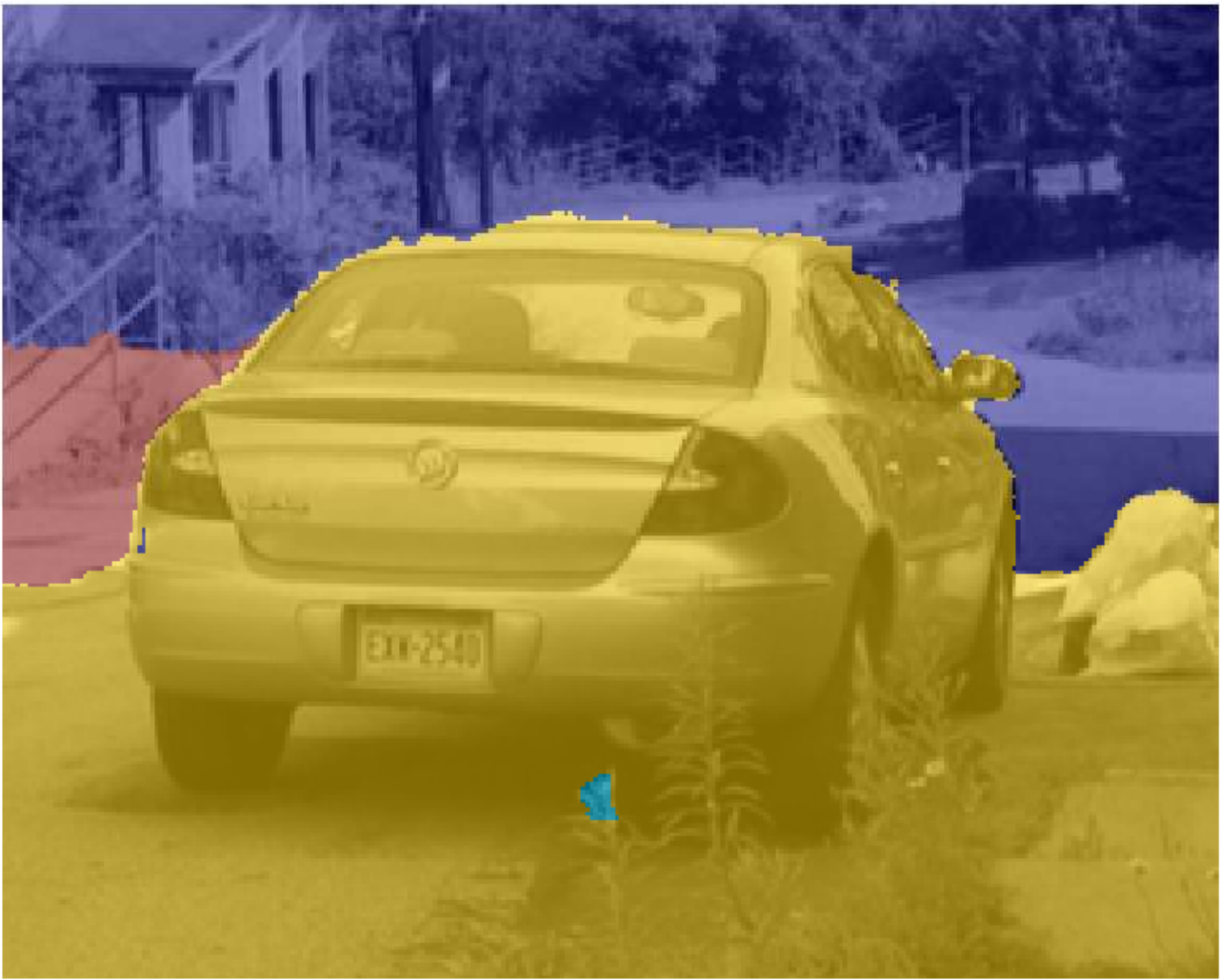}} \\
		   	   {\includegraphics[width=0.24\textwidth]{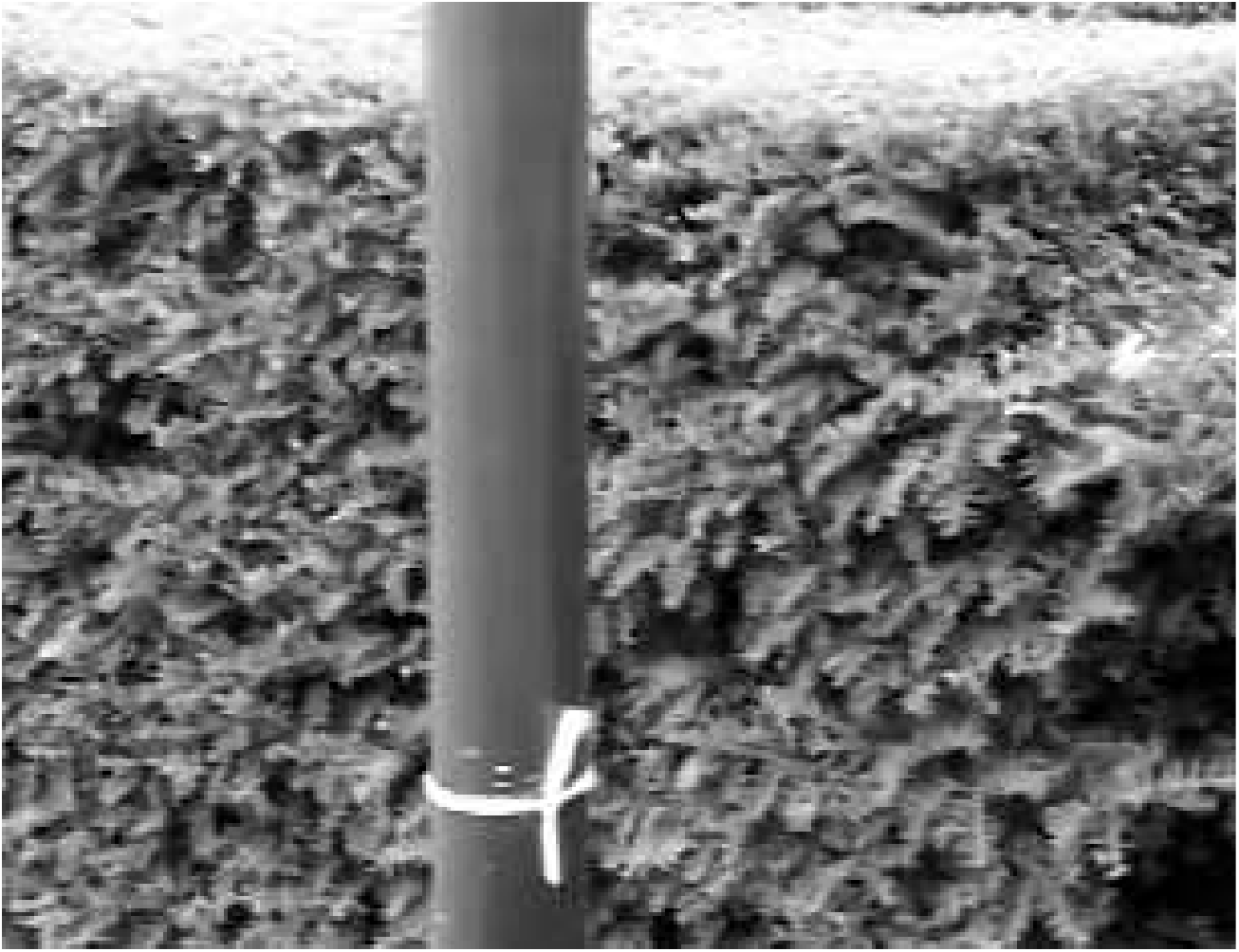}}   	                
               {\includegraphics[width=0.24\textwidth]{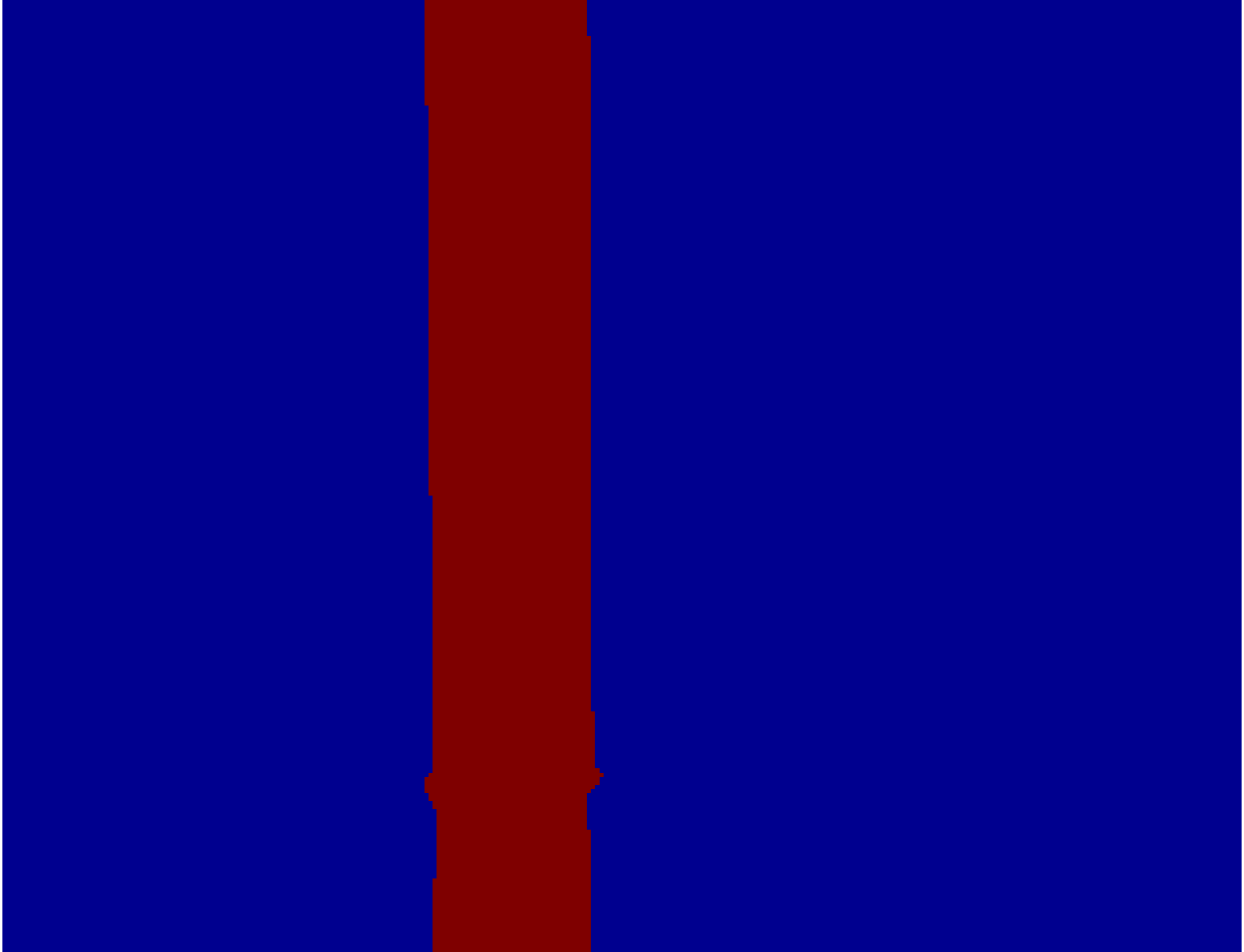}}
			   {\includegraphics[width=0.24\textwidth]{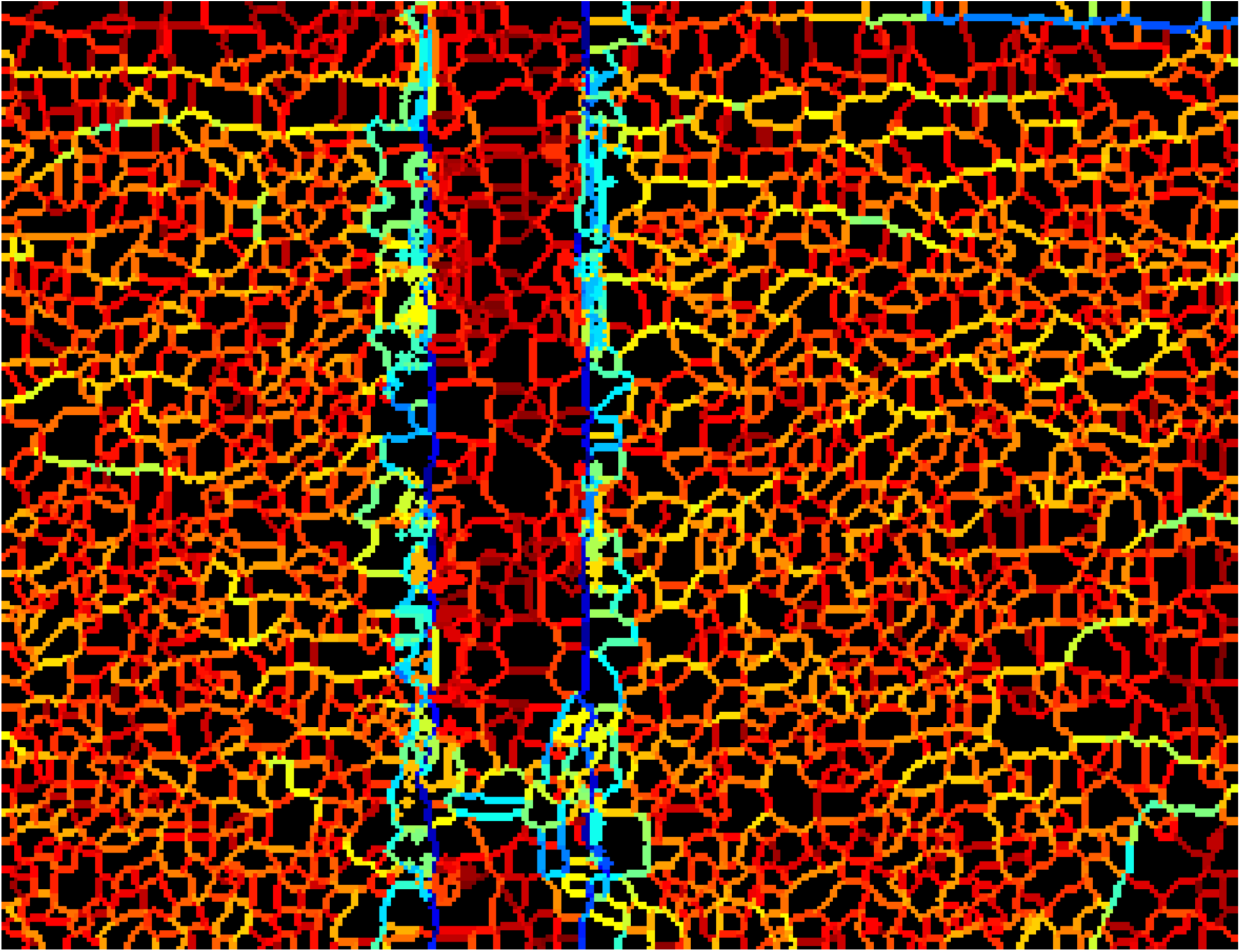}}   	                
               {\includegraphics[width=0.24\textwidth]{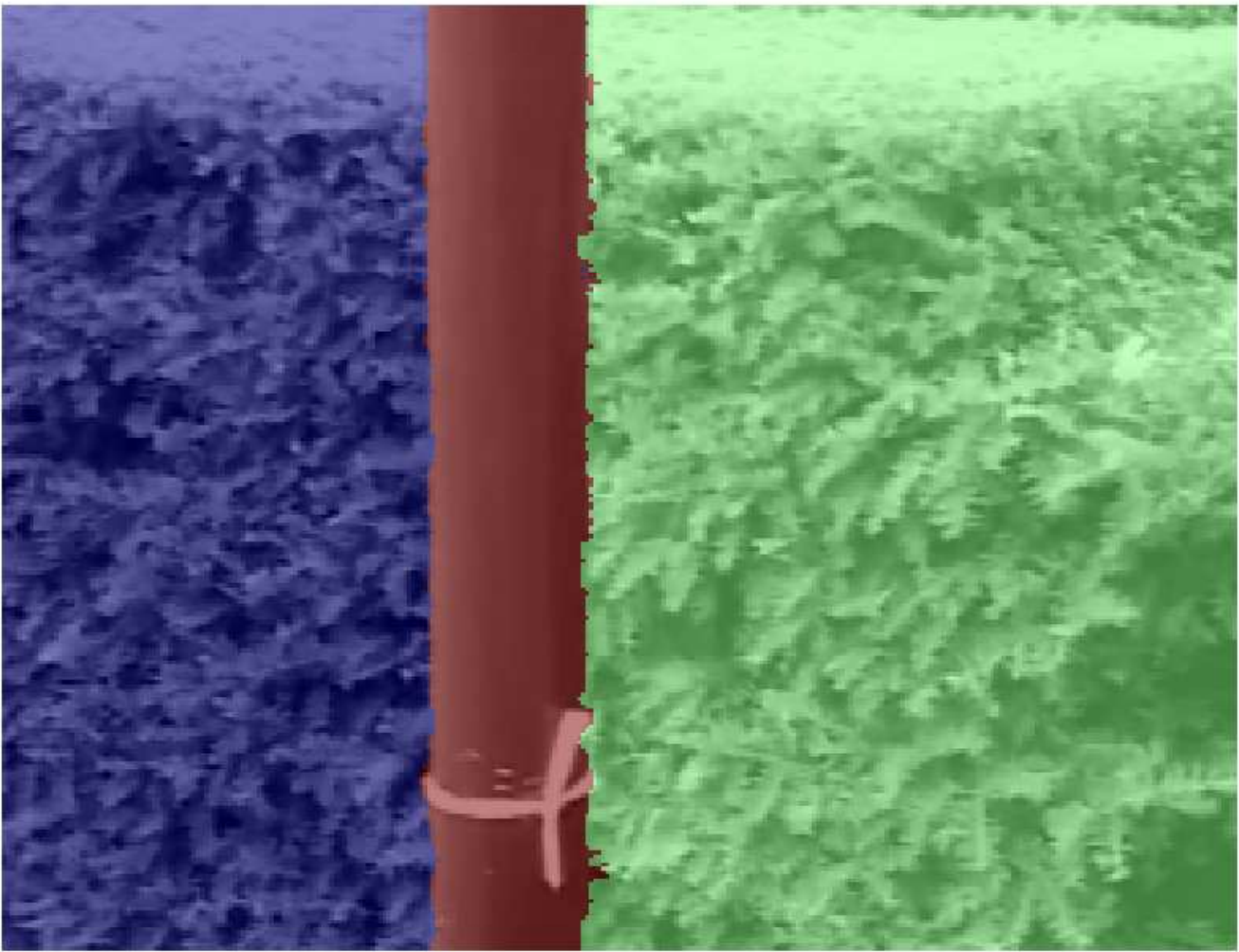}}
              \end{tabular}
            }
      }
     \caption{\sl Additional samples from the CMU dataset (first column), ground truth objects on these sequences (second column)	affinities between superpixels computed on the reference frames (third column), output of our algorithm (fourth column). Note that color coding does not represent the layers rather the distinct components on the layer map. Failures are related to small motion and miss detection of occluded regions.}
	  \label{fig-qualitative2}
\end{figure*}

Representative examples of successful detection are shown in Fig. \ref{fig-compare}, \ref{fig-qualitative} and \ref{fig-qualitative2}. In Fig. \ref{fig-compare}, a hiker and his hand are detected as separate detachable objects. The support region (left foot) is attributed to the ground plane, as expected, since there is insufficient evidence (photometric discontinuity) to separate it from the ground. The same goes for the squirrel (Fig. \ref{fig-compare} bottom row), and the cat in Fig. \ref{fig-qualitative} (fifth row). Note that the arm in Fig. \ref{fig-compare} appears as a detached object, since its support region (where it is attached) is outside the field of view, and therefore it appears to be floating in midair.  The same goes for the chair, couch, hand, and horse in Fig. \ref{fig-qualitative} (first to fourth row, respectively).  

\begin{figure}[!bh]
  \centerline 
      {
        \hbox
            {%\\
               \begin{tabular}{c}
                 %\subfloat[]
                 {\includegraphics[height=3.5cm,width=6.5cm]{painted-child-0271-im}}       
                 %\subfloat[]            
                 {\includegraphics[height=3.5cm,width=6.5cm]{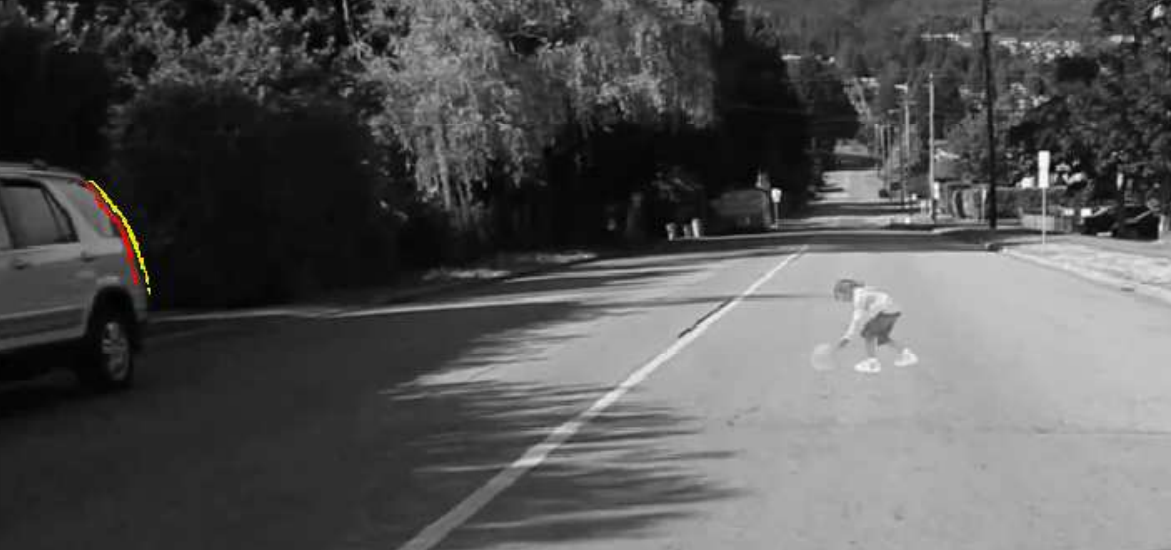}}  \\
                 %\subfloat[]
                 {\includegraphics[height=3.5cm,width=6.5cm]{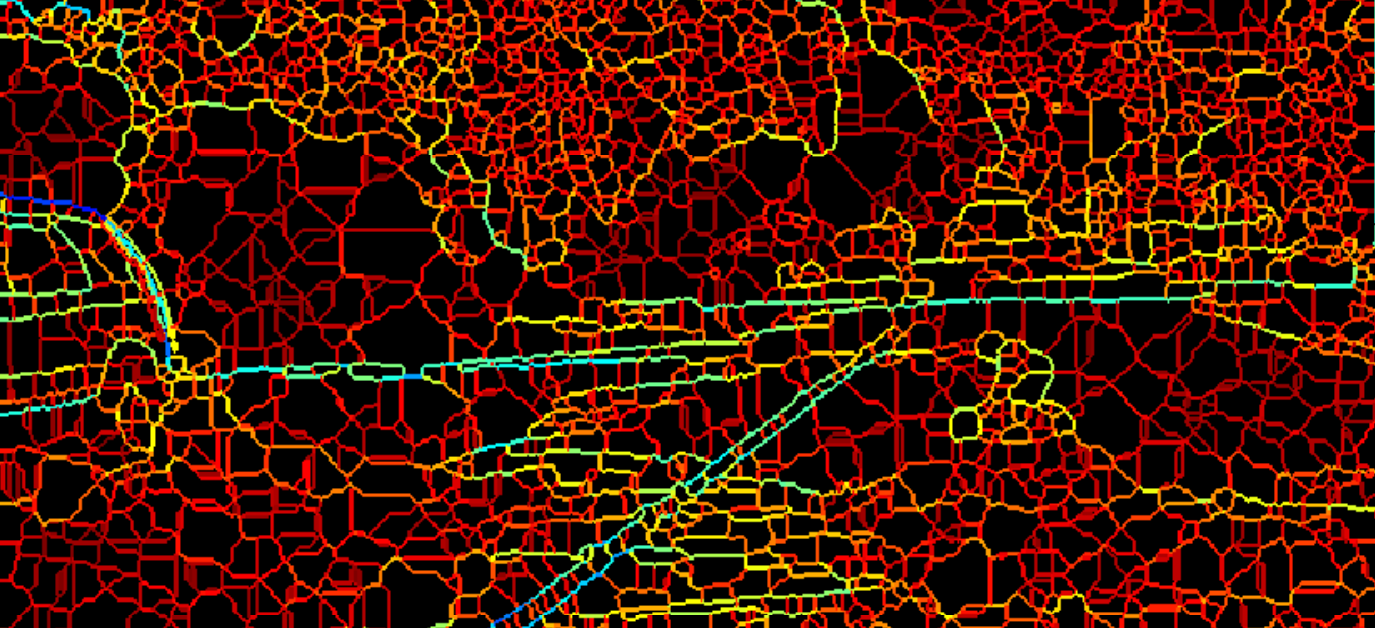}}                   
                % \subfloat[]
                 {\includegraphics[height=3.5cm,width=6.5cm]{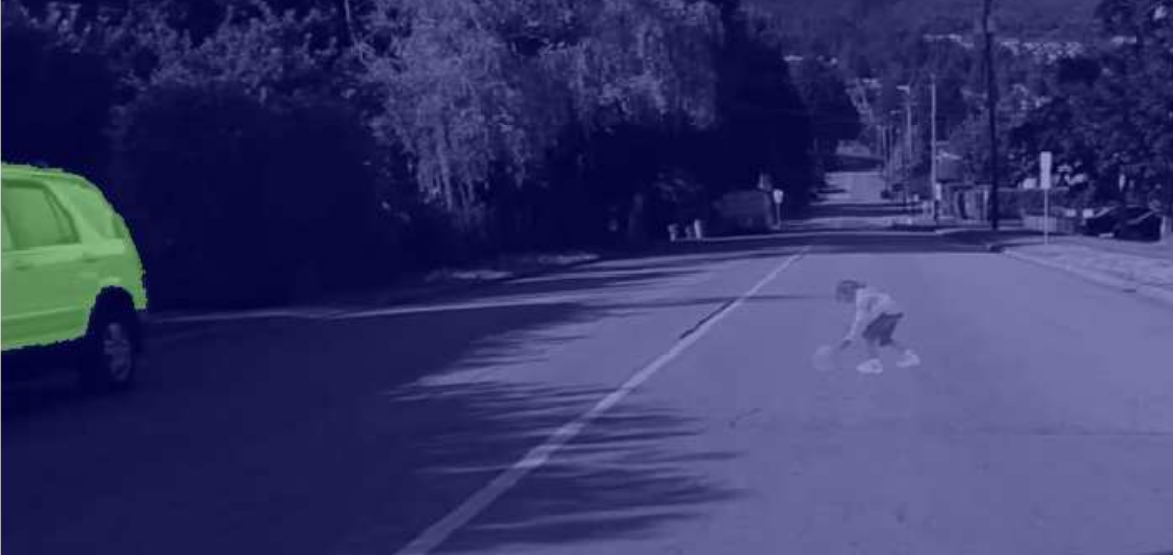}}                  
              \end{tabular}
            }
        }
     \caption{\sl A child figure is painted on the road in West Vancouver (a). Unlike a real pedestrian or a car, this drawing does not cause any occlusion (b). Therefore, given the image and motion features (c), our algorithm does not detect it as a detachable object while it segments the nearby car (d). The original sequence can be  seen at http://reviews.cnet.com/8301-13746\_7-20016169-48.html. }
\label{fig-painted-child}
\end{figure}

The sequence in Fig. \ref{fig-compare}, from \cite{steinH09}, is too short to capture an entire walking cycle, so the person cannot be positively identified as detachable. Using a longer sequence, such as the Soccer scene in Fig. \ref{fig-occlusions}, shows that we can successfully aggregate the entire person into one segment, and therefore positively detect him as a detachable object. 
The sequence in Fig. \ref{fig-painted-child} is taken at an intersection where a child figure is painted on the road. Unlike a real pedestrian or a car, this does not trigger occlusions, and is therefore not detected as a detachable object, unlike the nearby car.

\subsection{Failure modes}

\begin{figure*}[bth!]
  \centerline 
      {
        \hbox
            {%\\
               \begin{tabular}{cccc}
               {\includegraphics[width=0.24\textwidth]{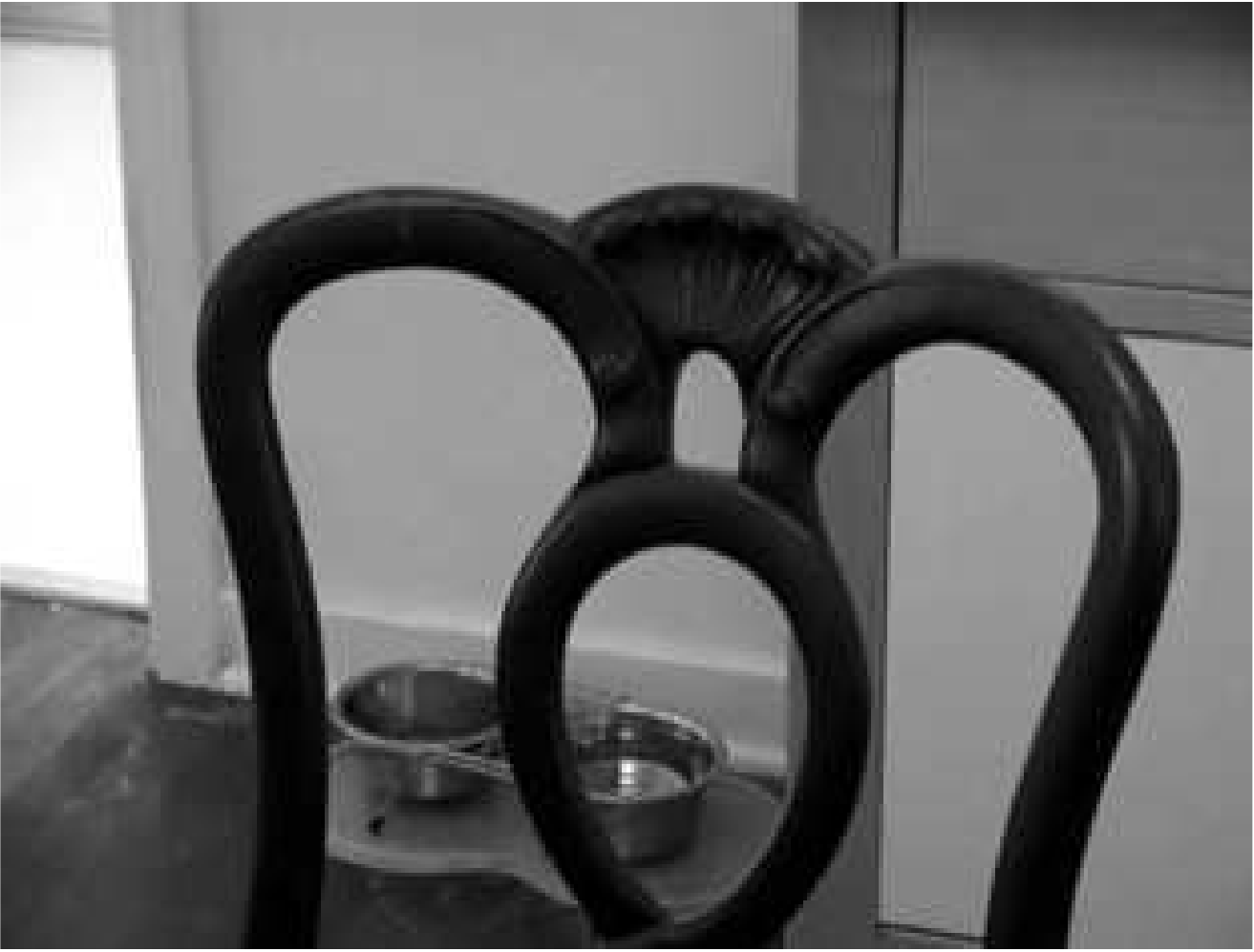}}   	                
               {\includegraphics[width=0.24\textwidth]{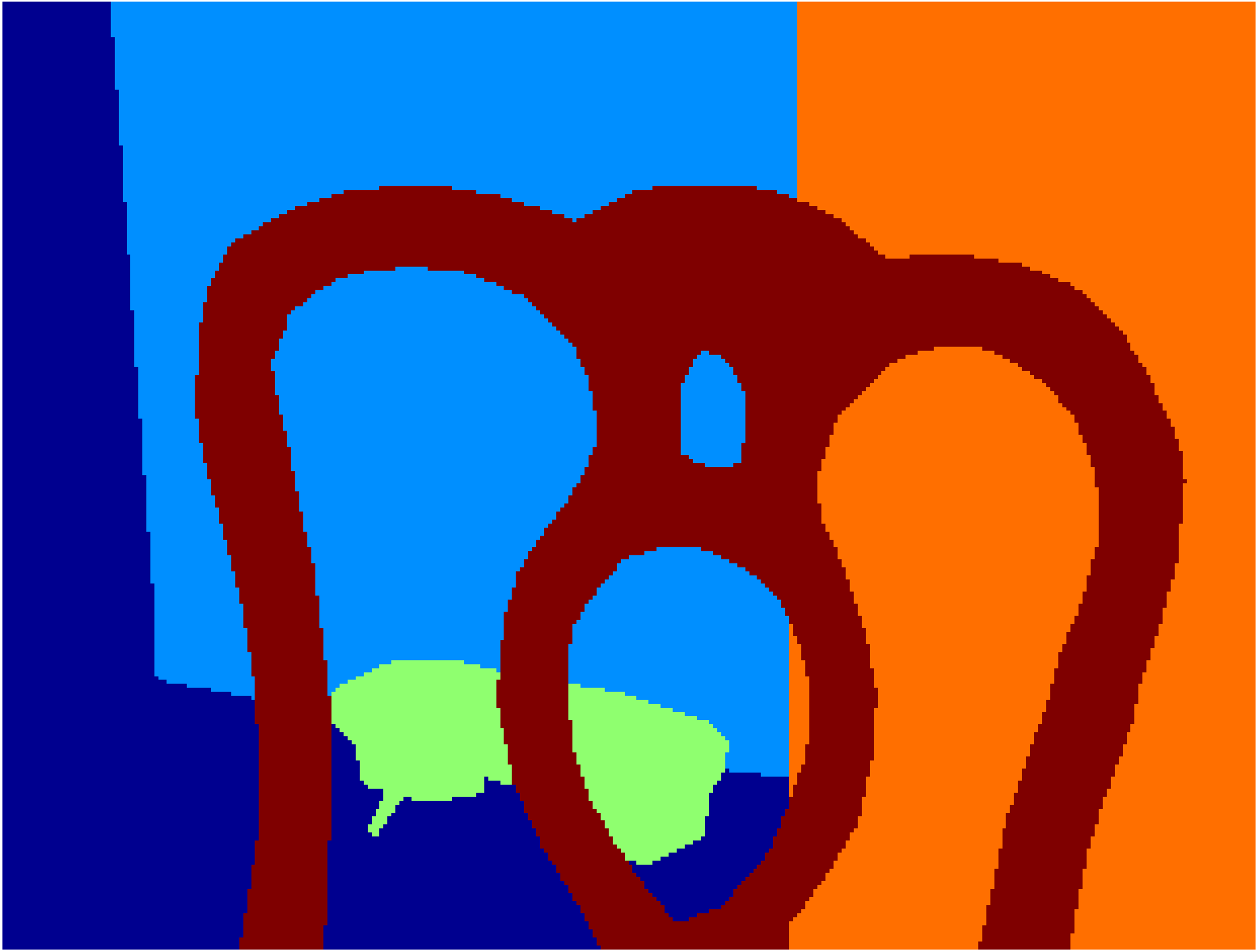}}
			   {\includegraphics[width=0.24\textwidth]{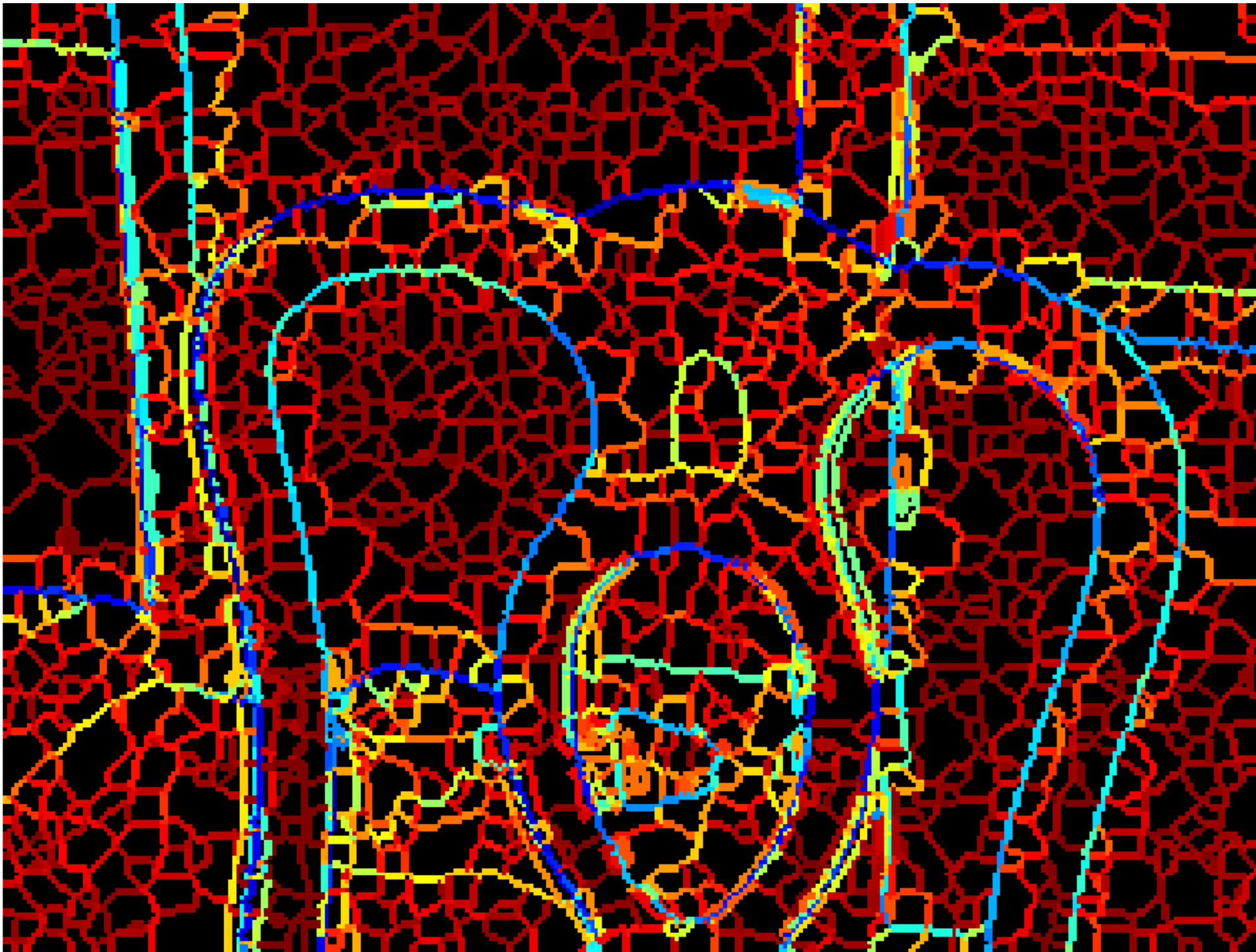}}   	                
               {\includegraphics[width=0.24\textwidth]{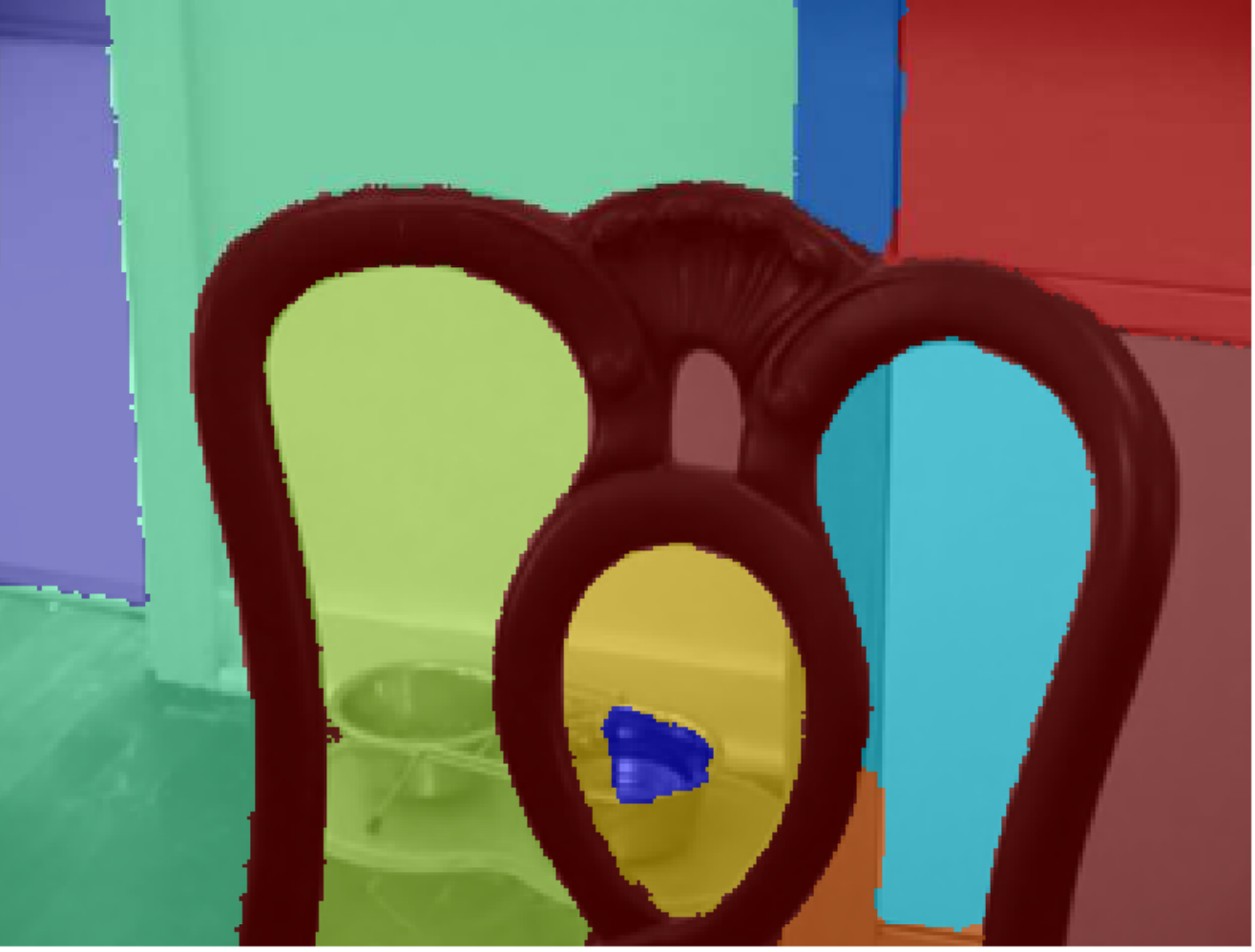}}   \\	                		
			   {\includegraphics[width=0.24\textwidth]{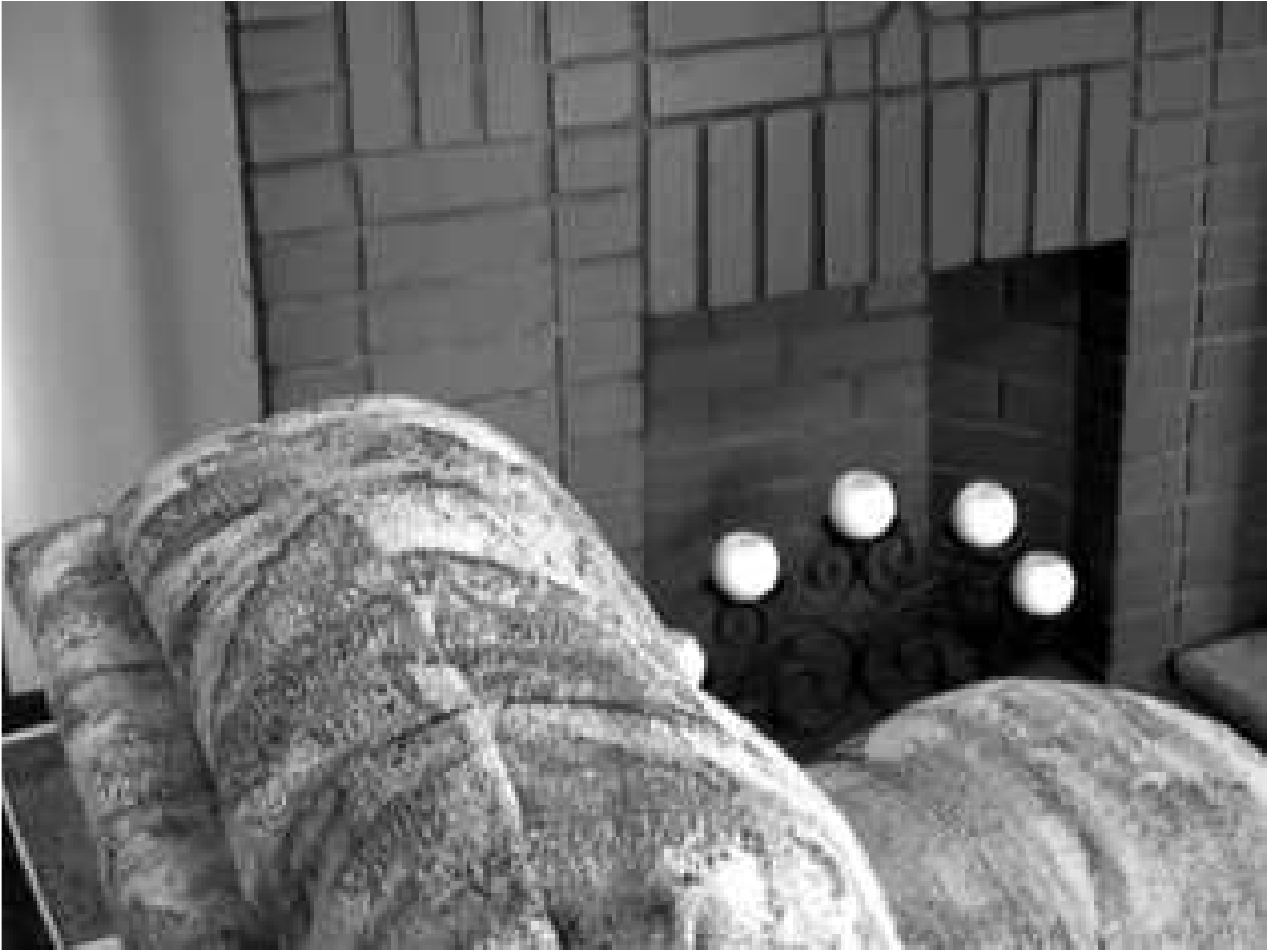}}   	                
               {\includegraphics[width=0.24\textwidth]{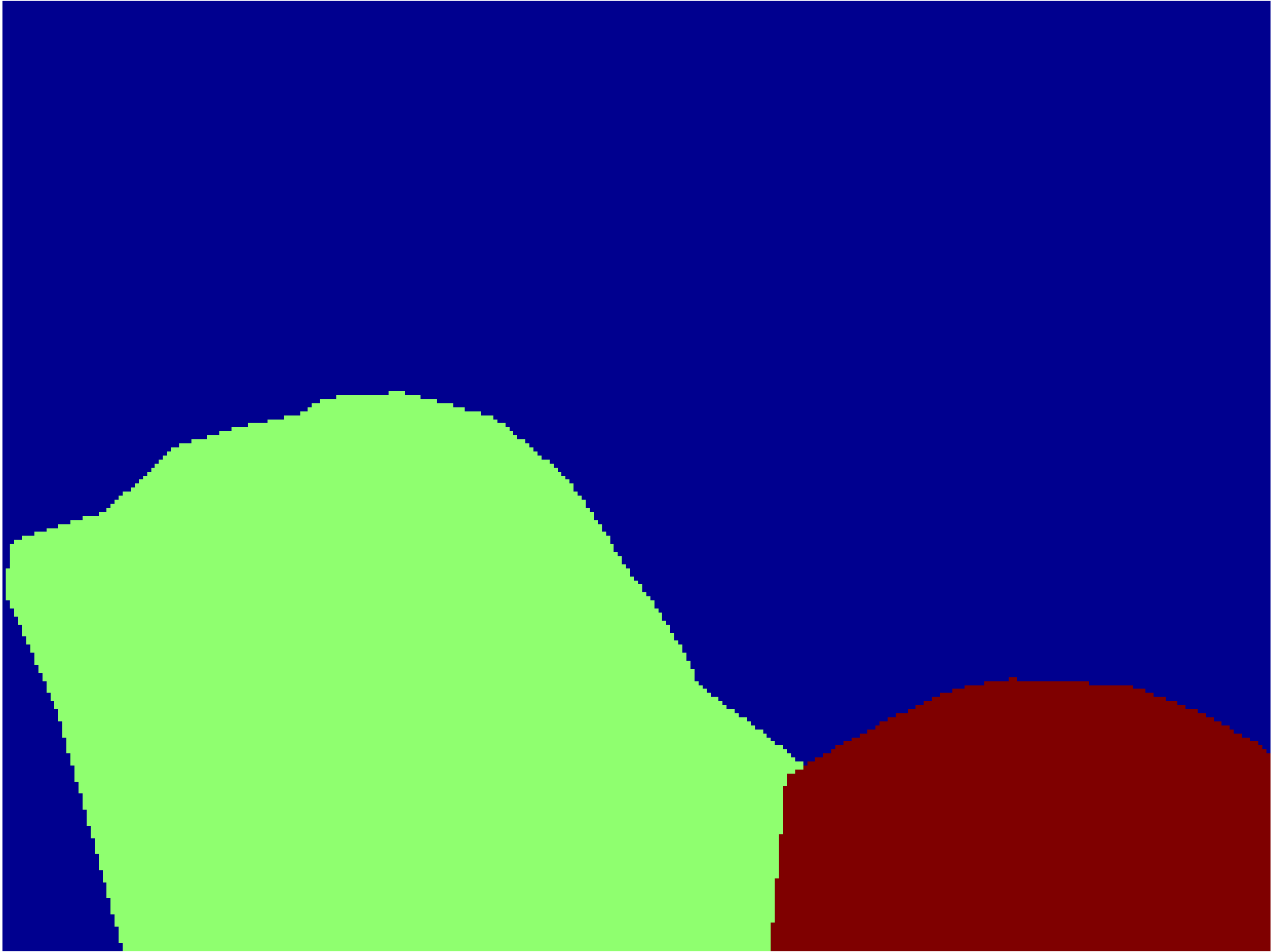}}
			   {\includegraphics[width=0.24\textwidth]{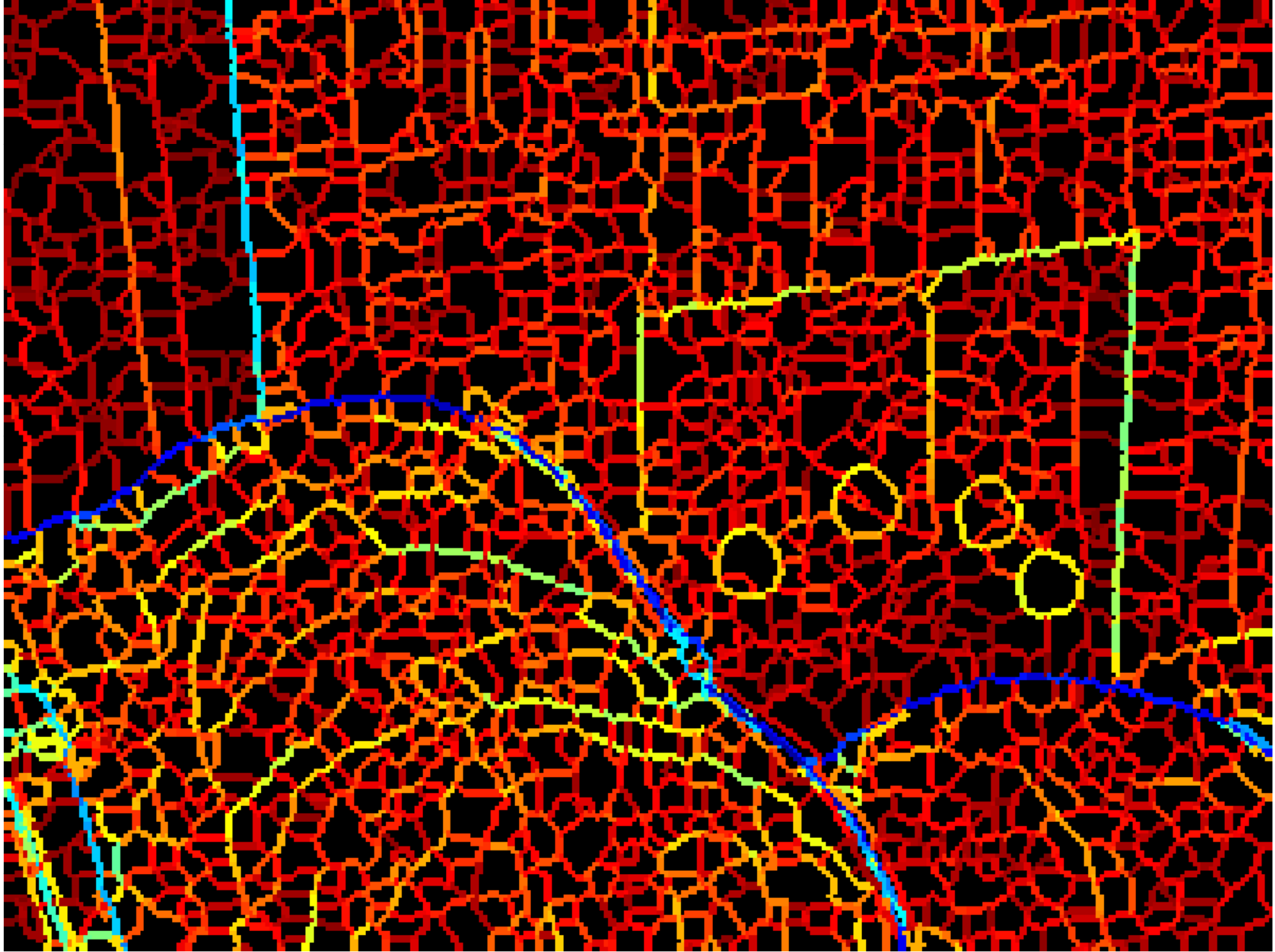}}   	                
               {\includegraphics[width=0.24\textwidth]{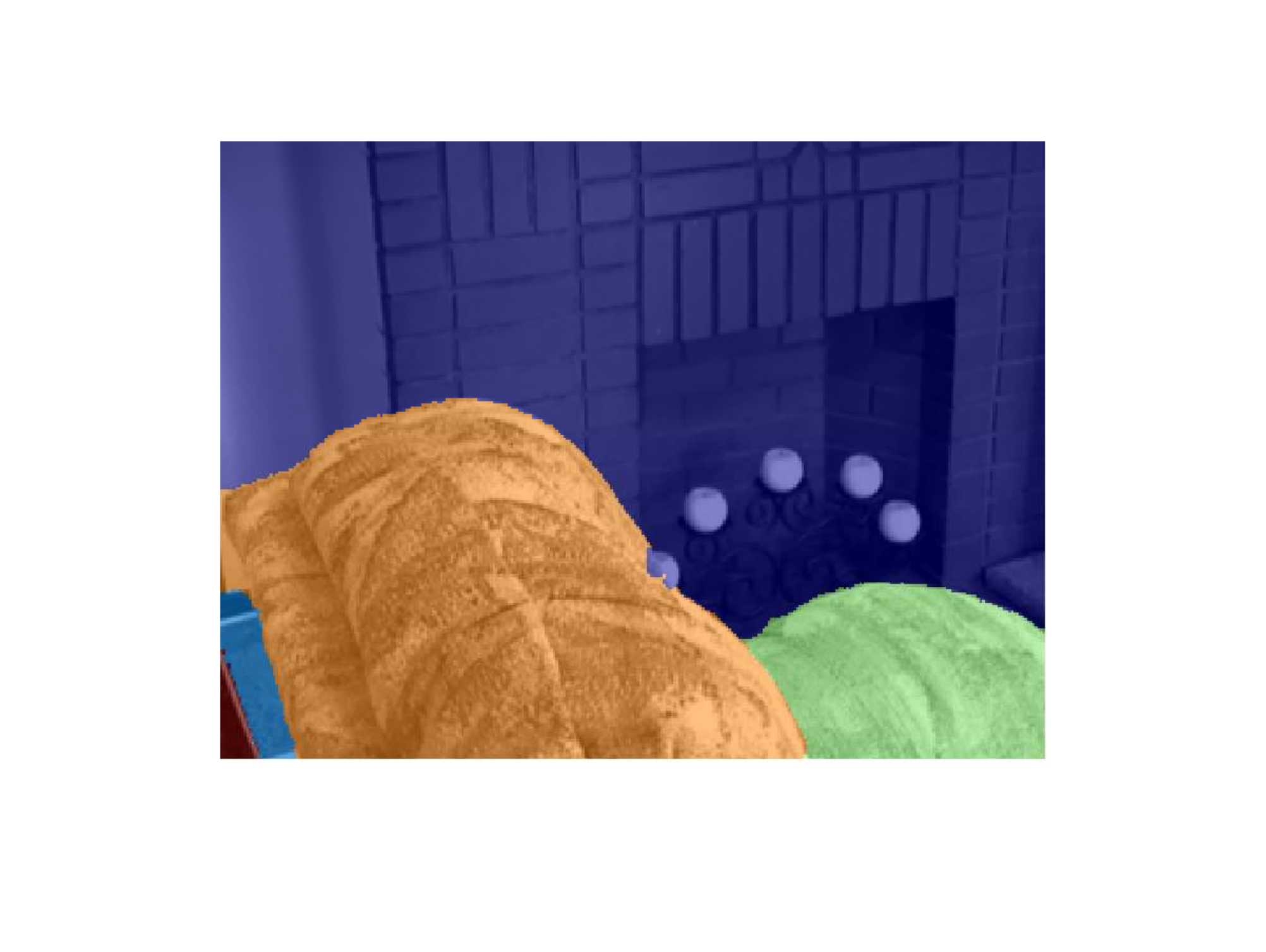}} \\
			   {\includegraphics[width=0.24\textwidth]{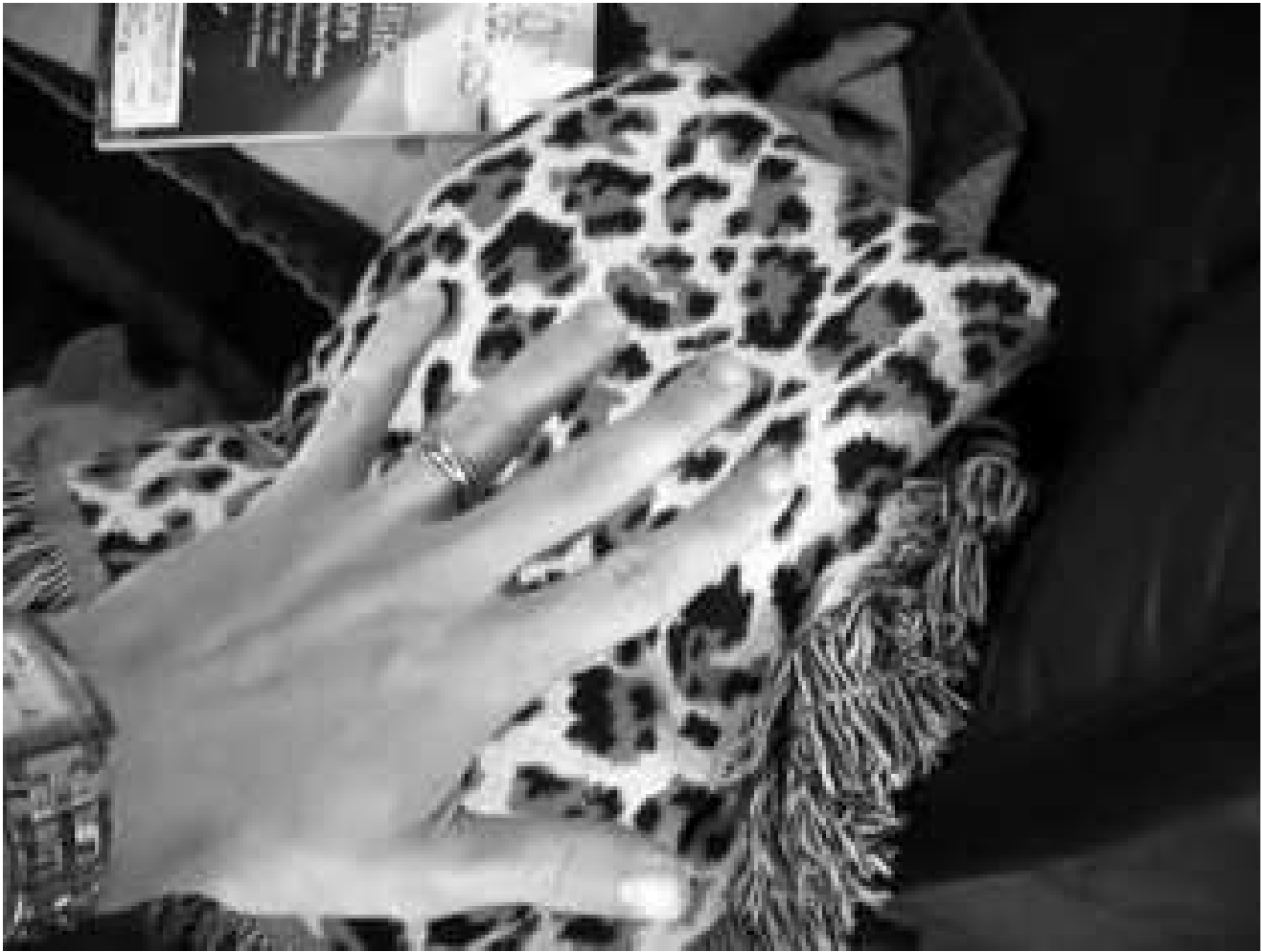}}   	                
               {\includegraphics[width=0.24\textwidth]{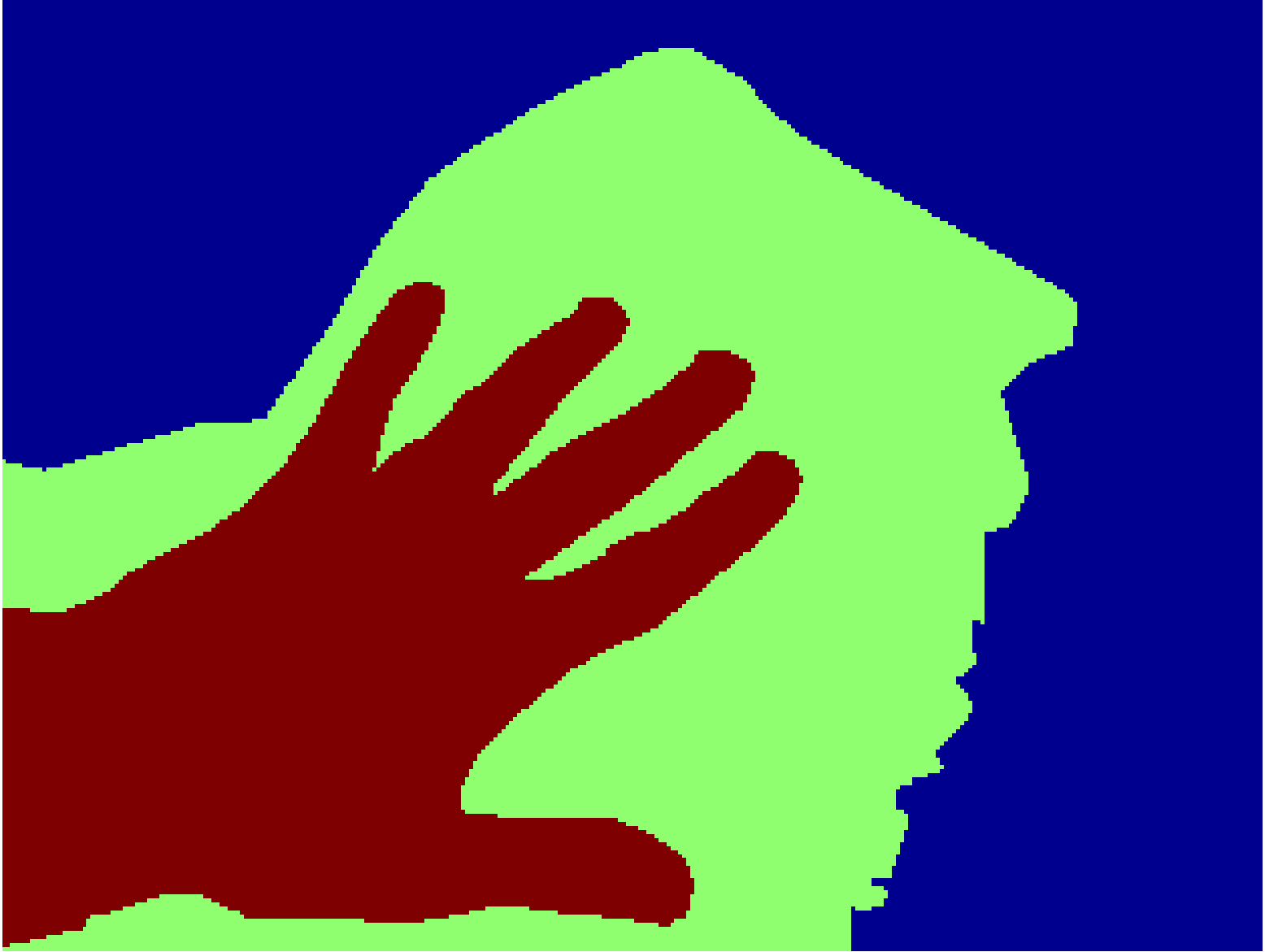}}
			   {\includegraphics[width=0.24\textwidth]{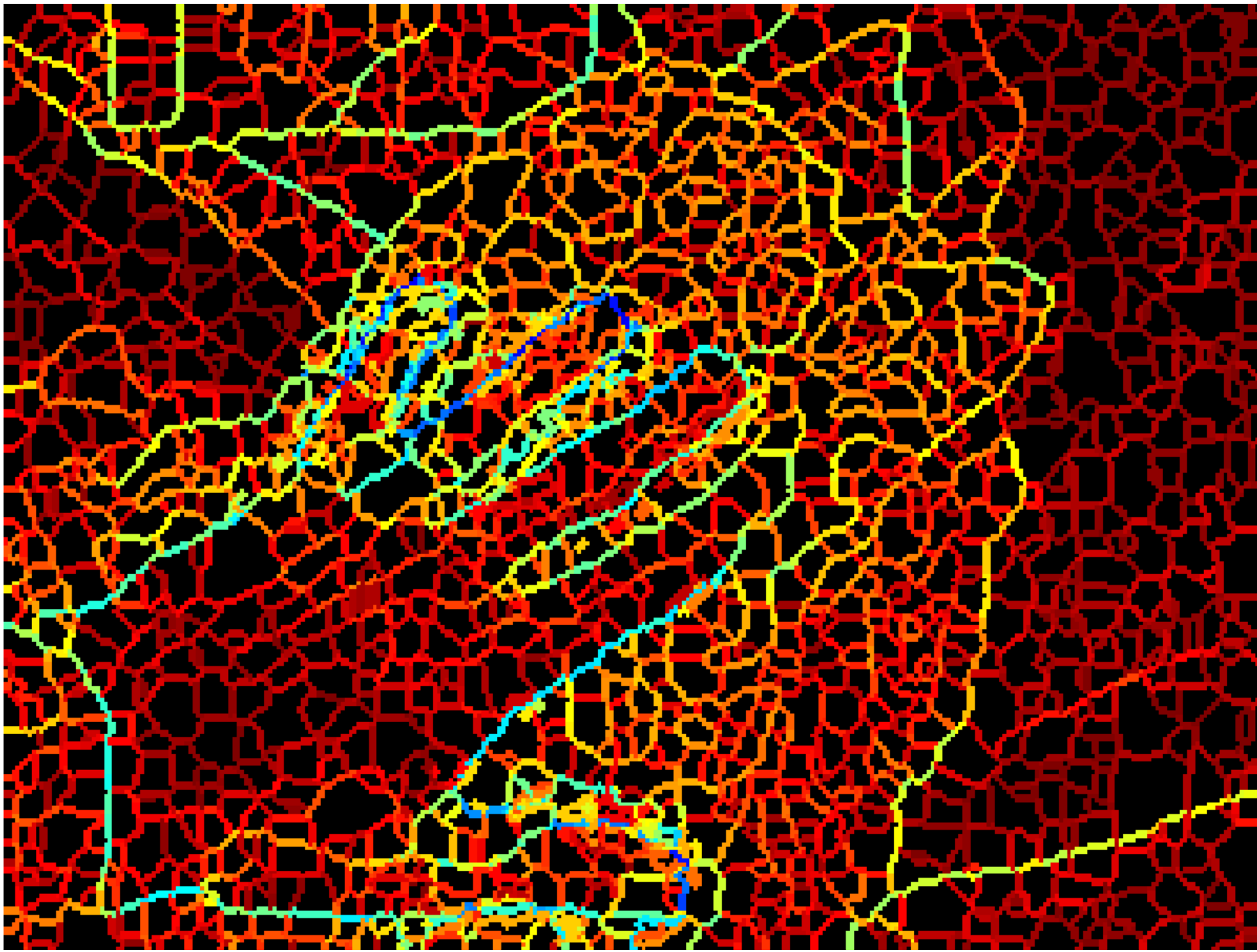}}   	                
               {\includegraphics[width=0.24\textwidth]{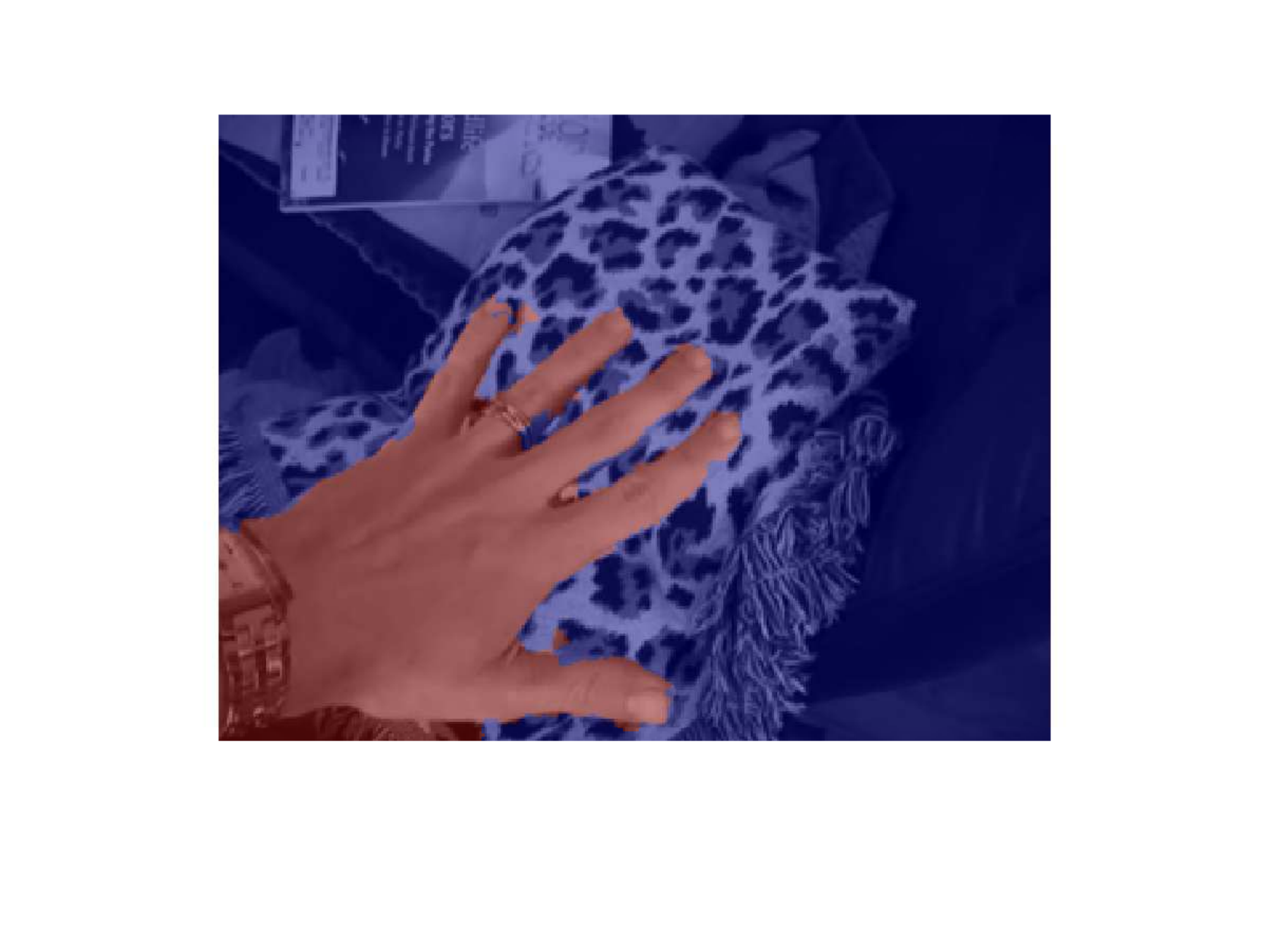}} \\
			   {\includegraphics[width=0.24\textwidth]{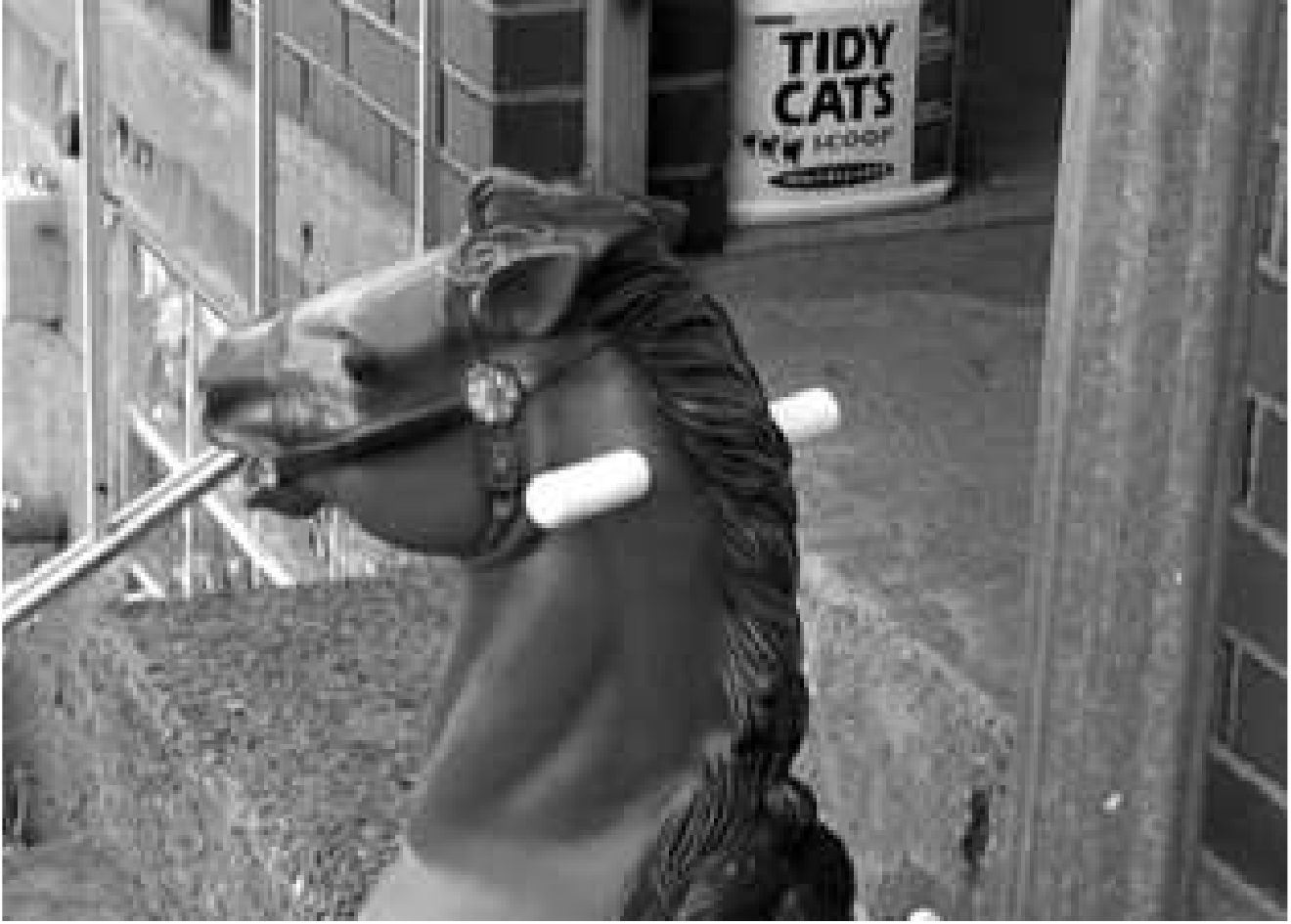}}   	                
               {\includegraphics[width=0.24\textwidth]{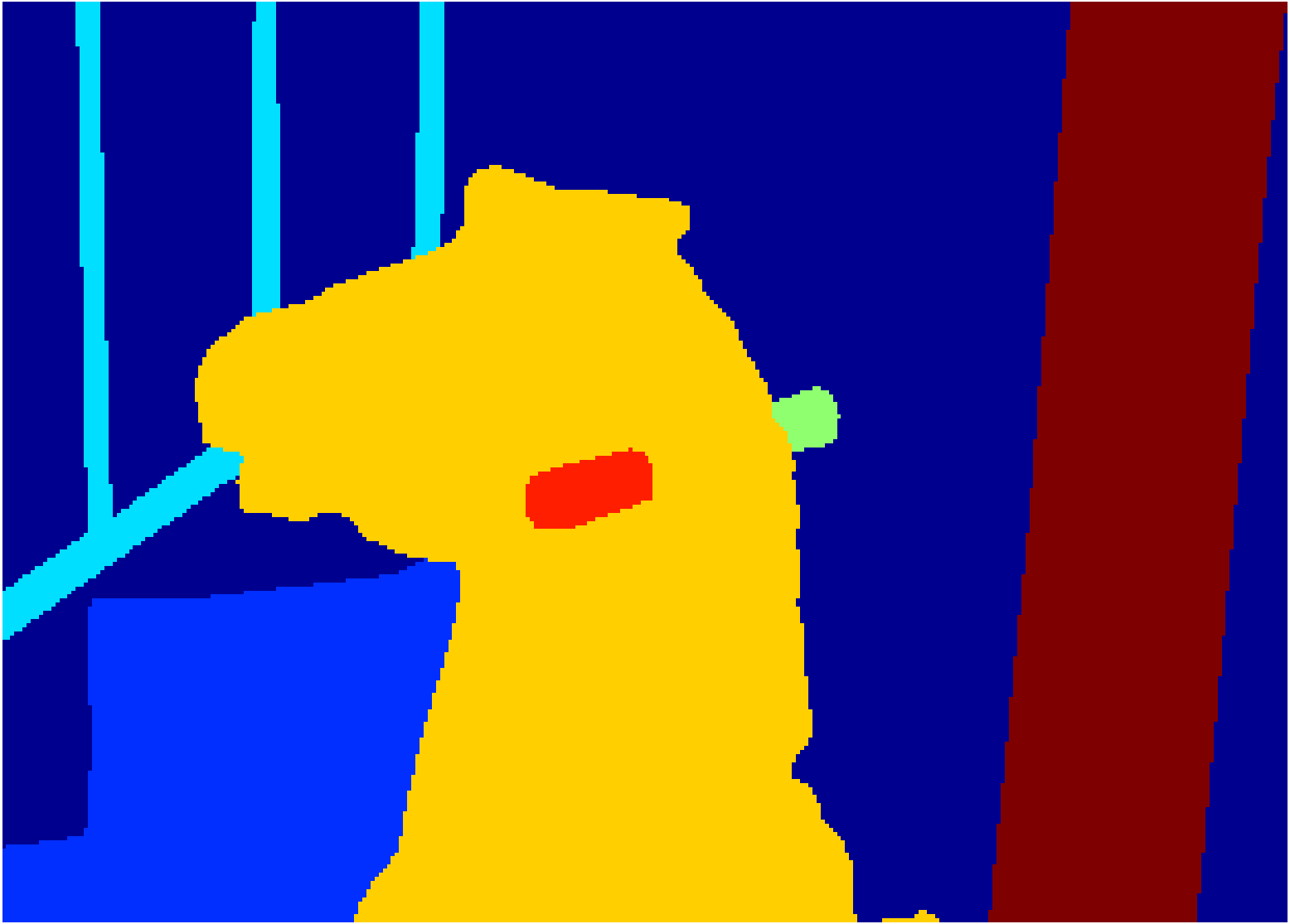}}
			   {\includegraphics[width=0.24\textwidth]{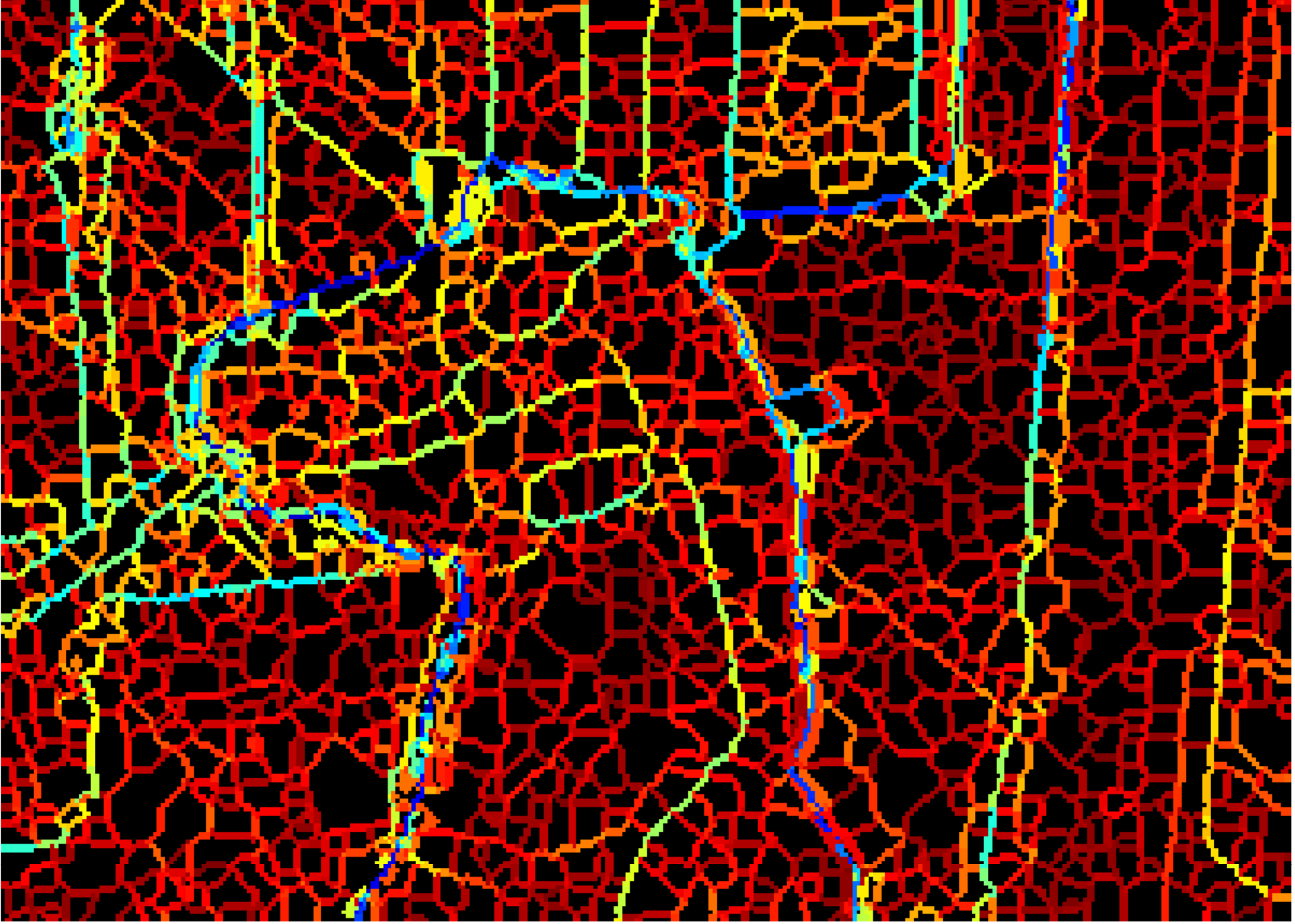}}   	                
               {\includegraphics[width=0.24\textwidth]{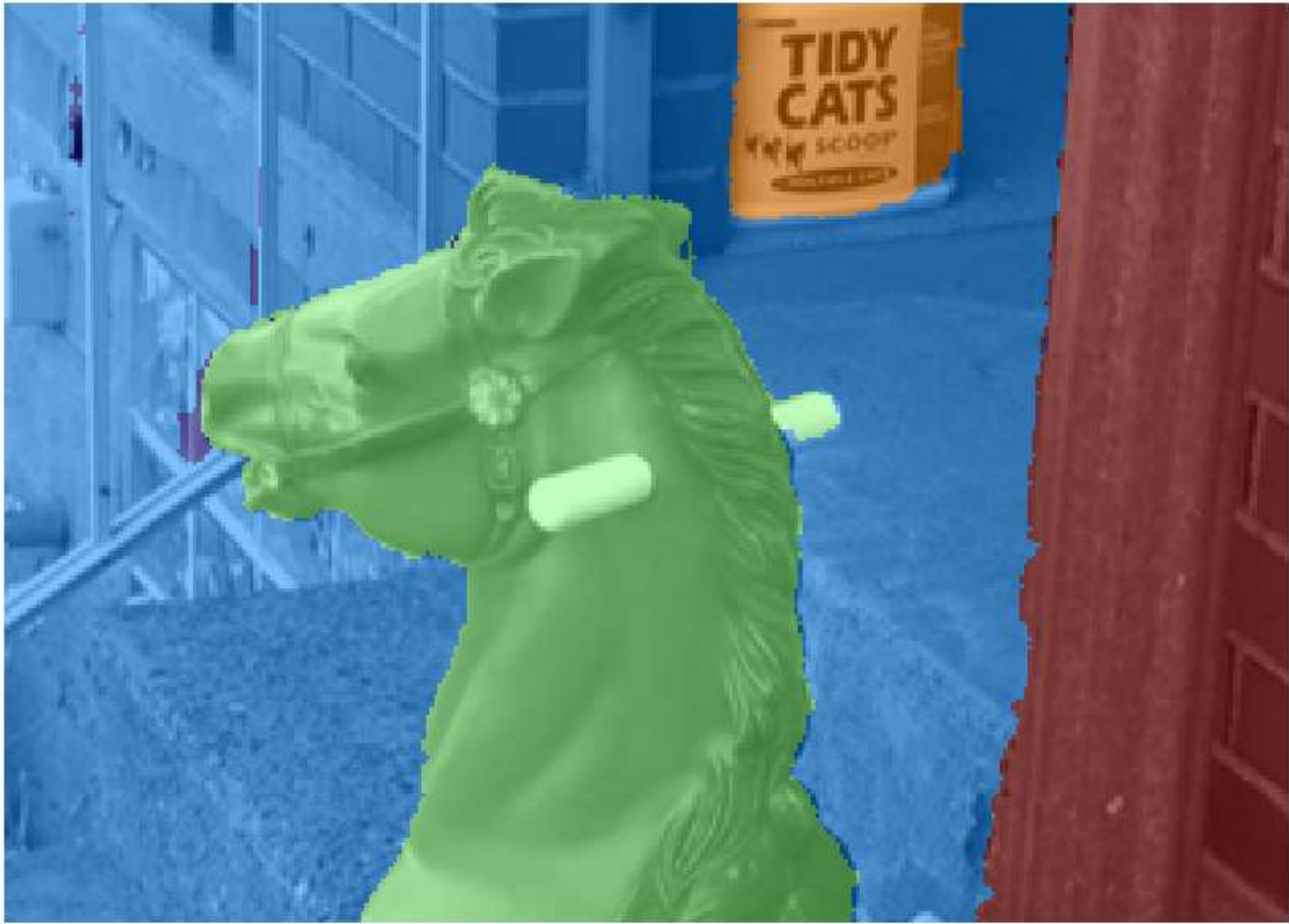}}\\
			   {\includegraphics[width=0.24\textwidth]{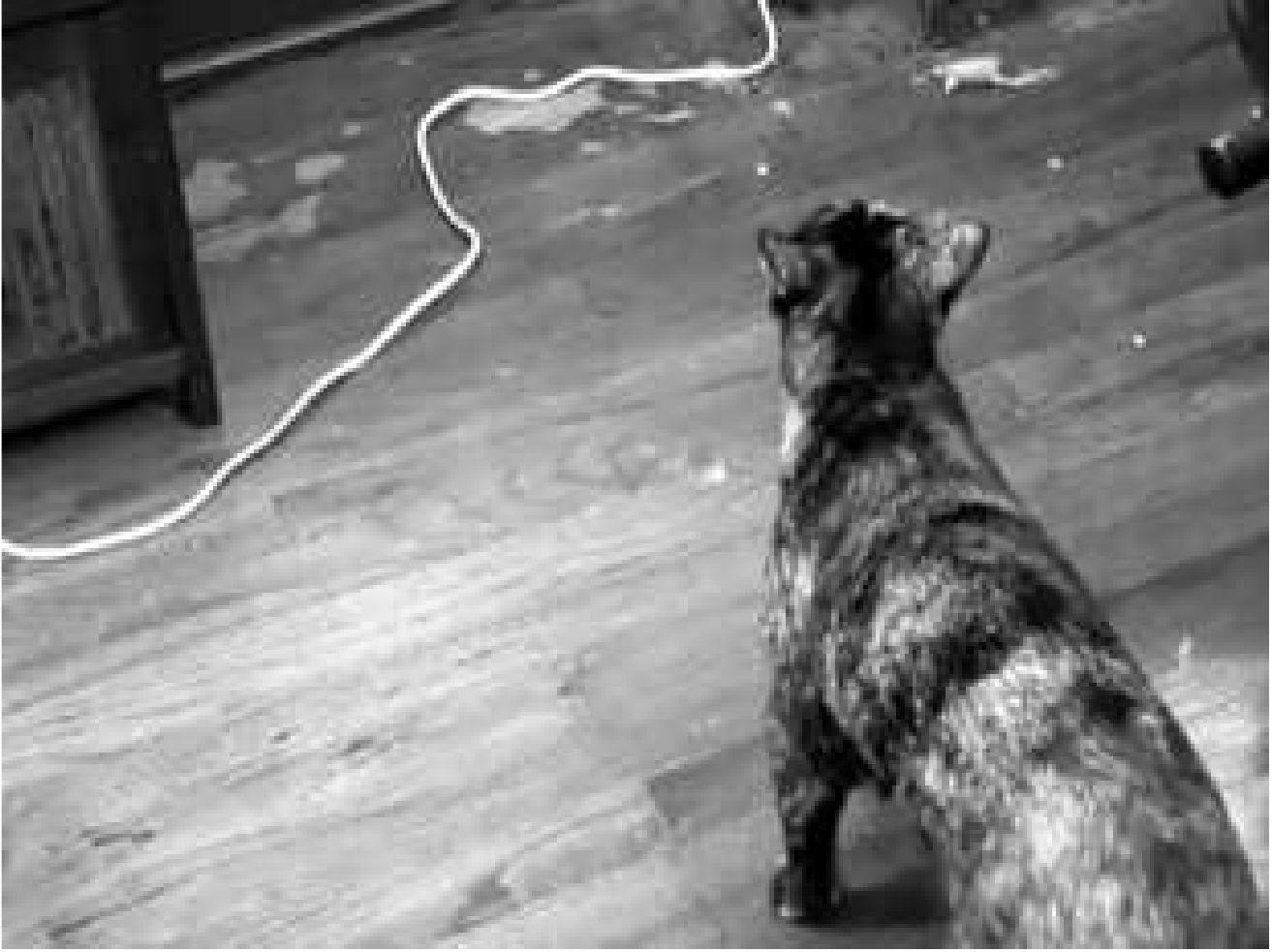}}   	                
               {\includegraphics[width=0.24\textwidth]{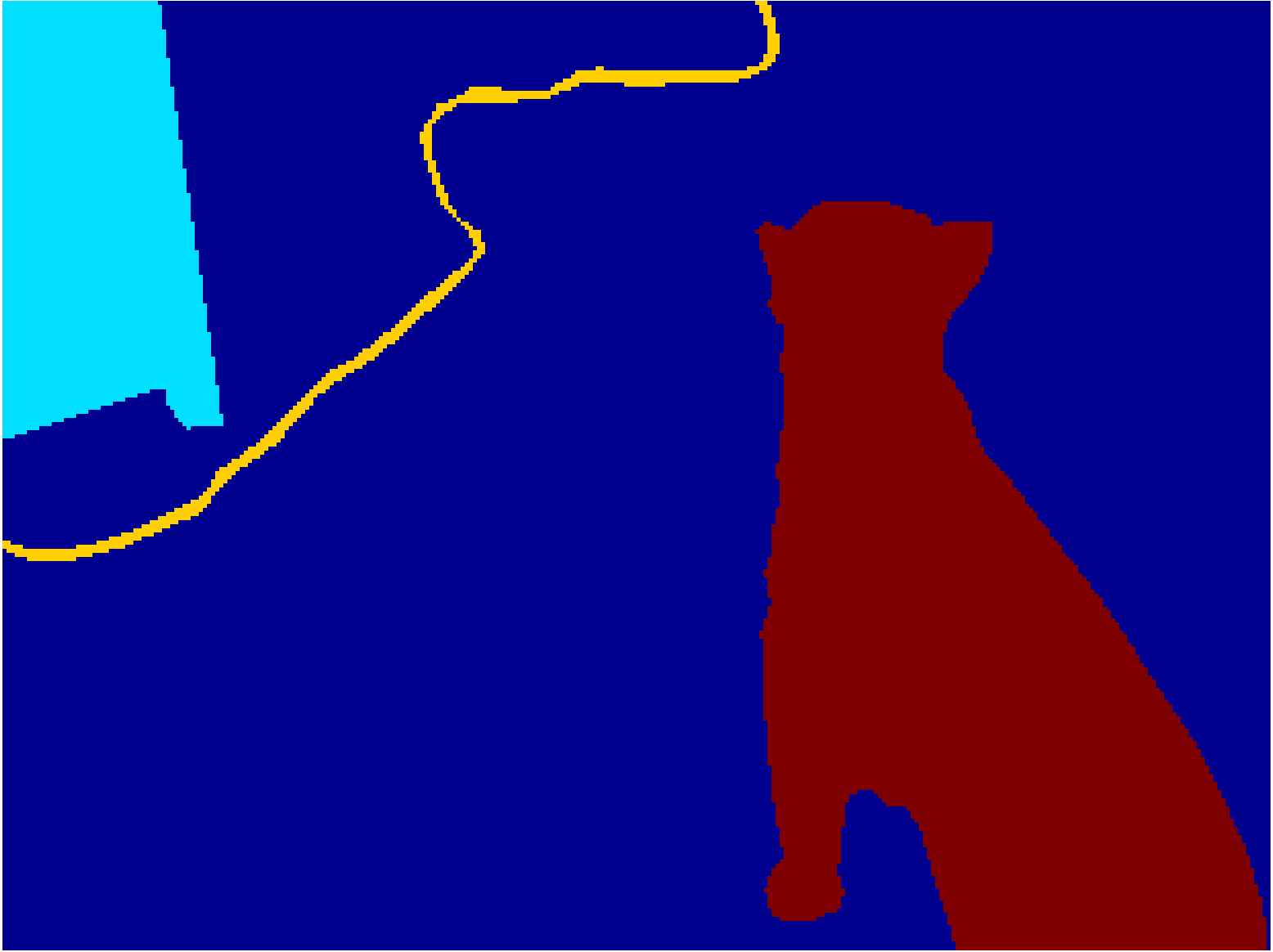}}
			   {\includegraphics[width=0.24\textwidth]{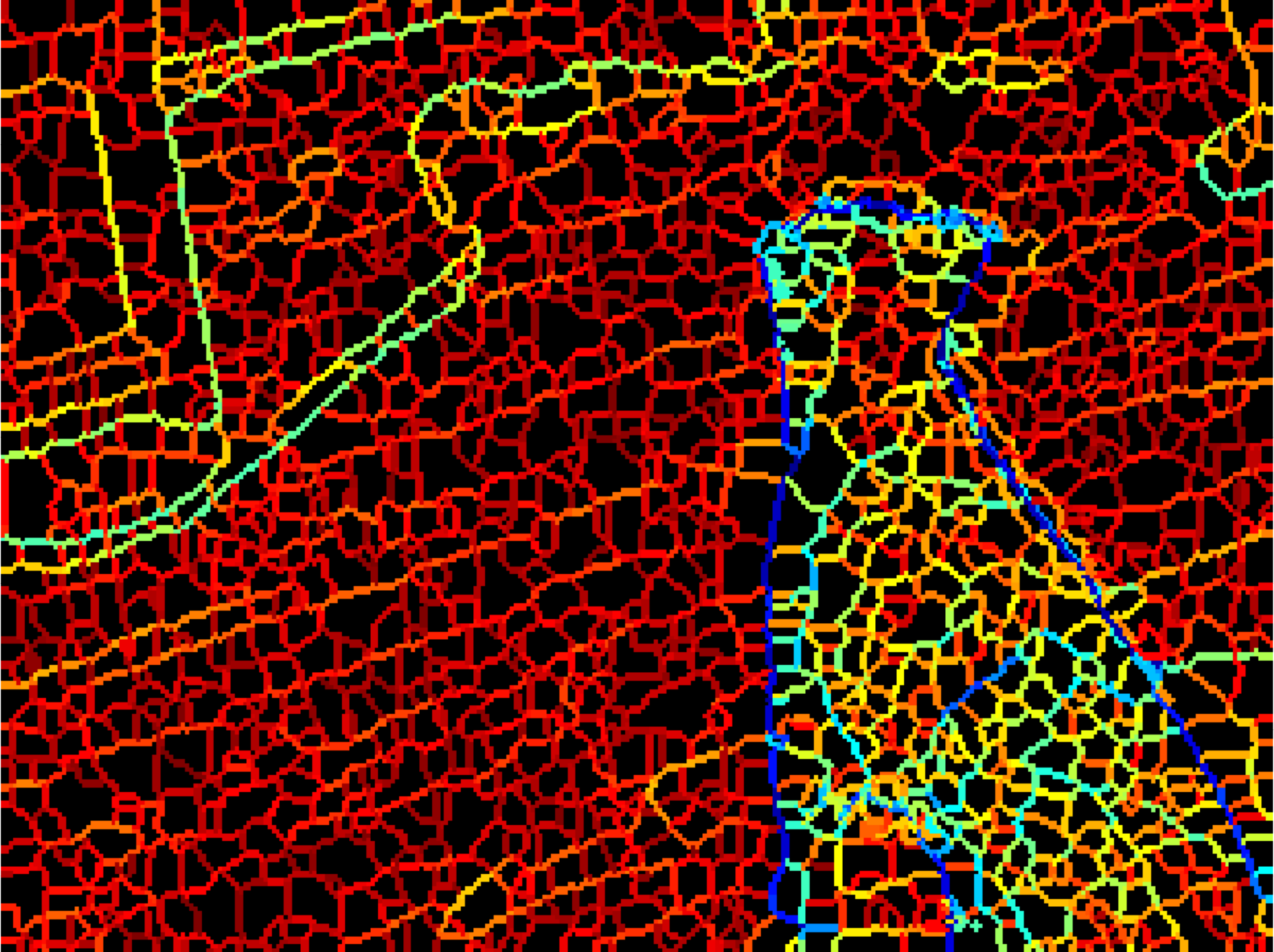}}   	                
               {\includegraphics[width=0.24\textwidth]{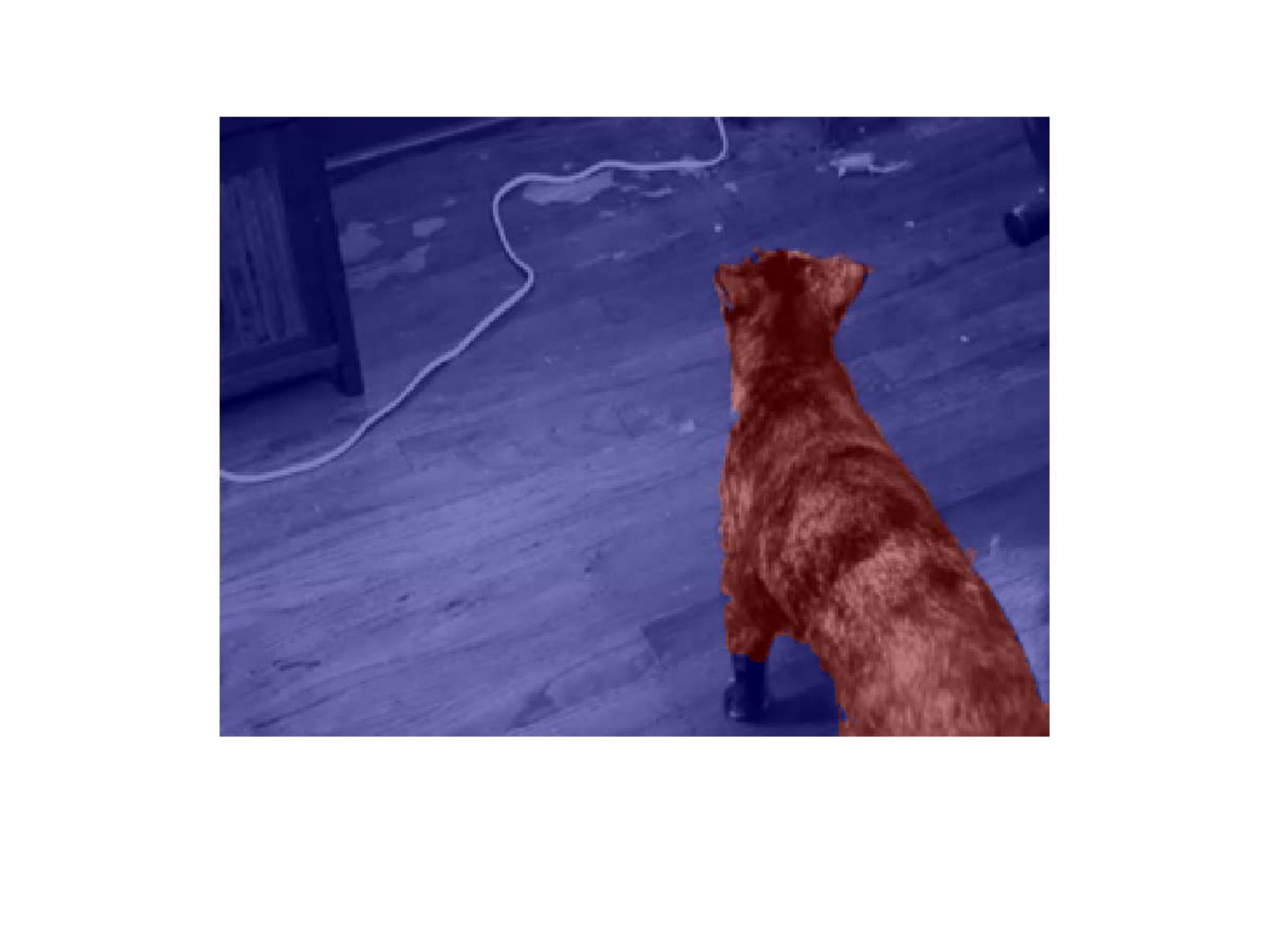}} \\
		   	   {\includegraphics[width=0.24\textwidth]{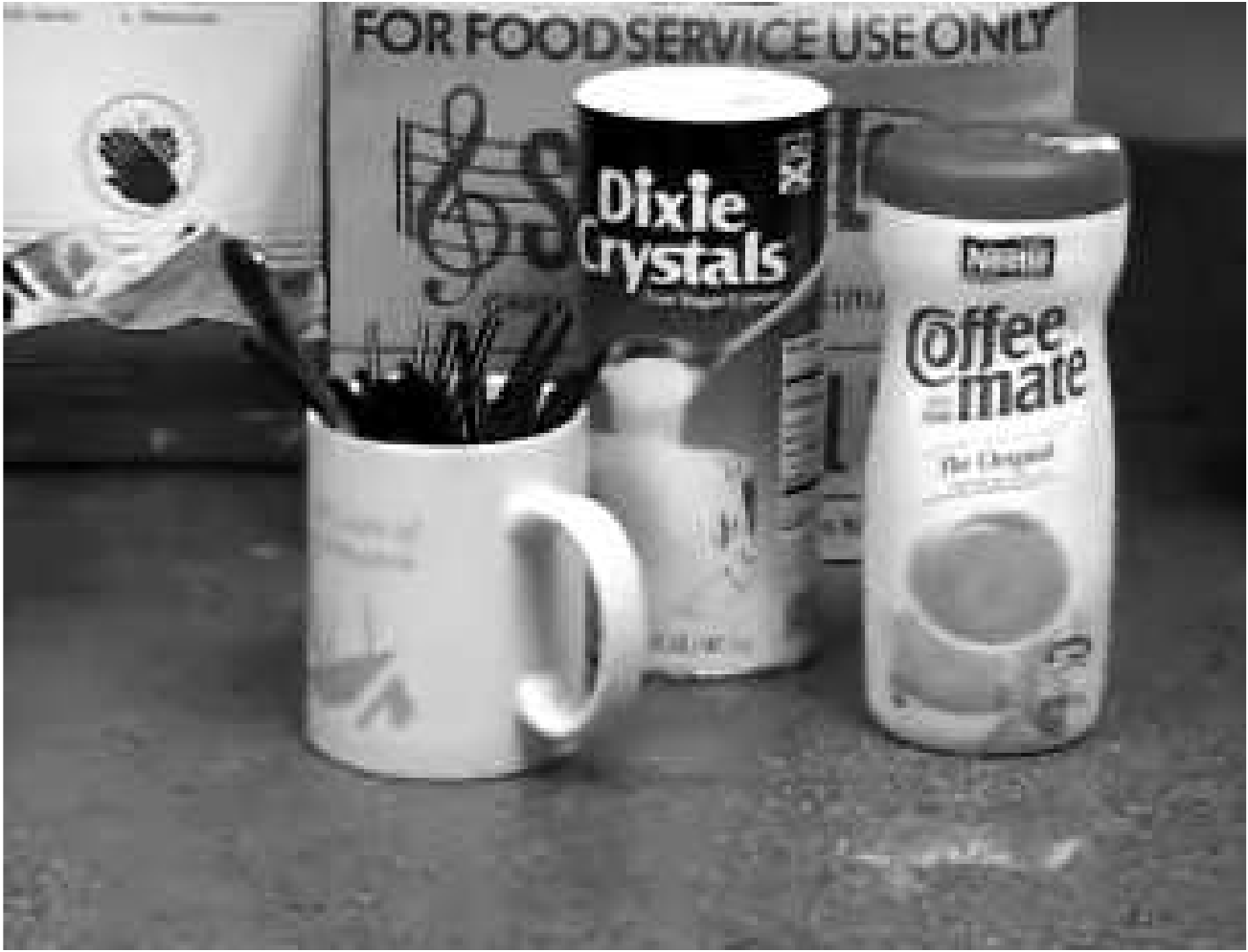}}   	                
               {\includegraphics[width=0.24\textwidth]{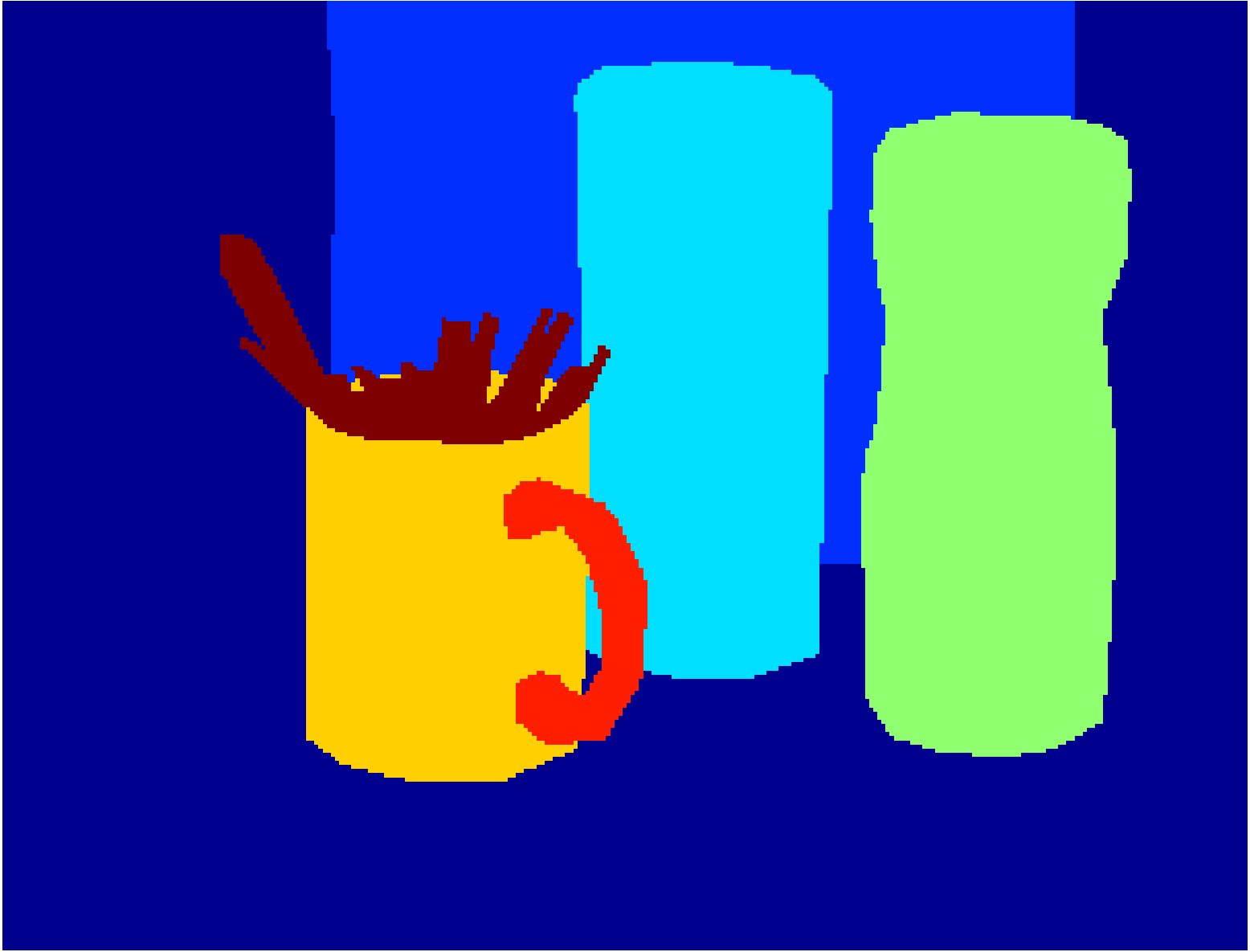}}
			   {\includegraphics[width=0.24\textwidth]{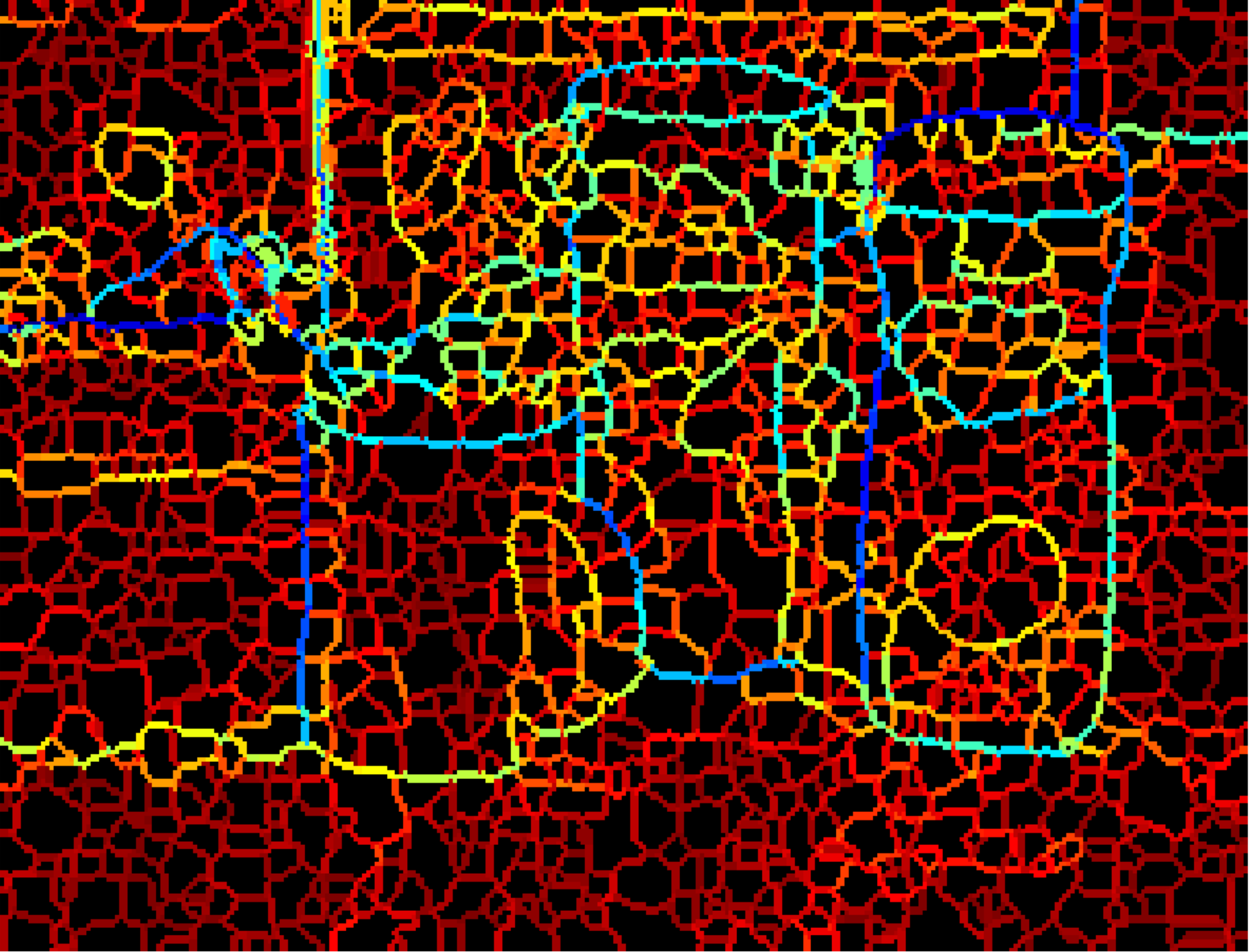}}   	                
               {\includegraphics[width=0.24\textwidth]{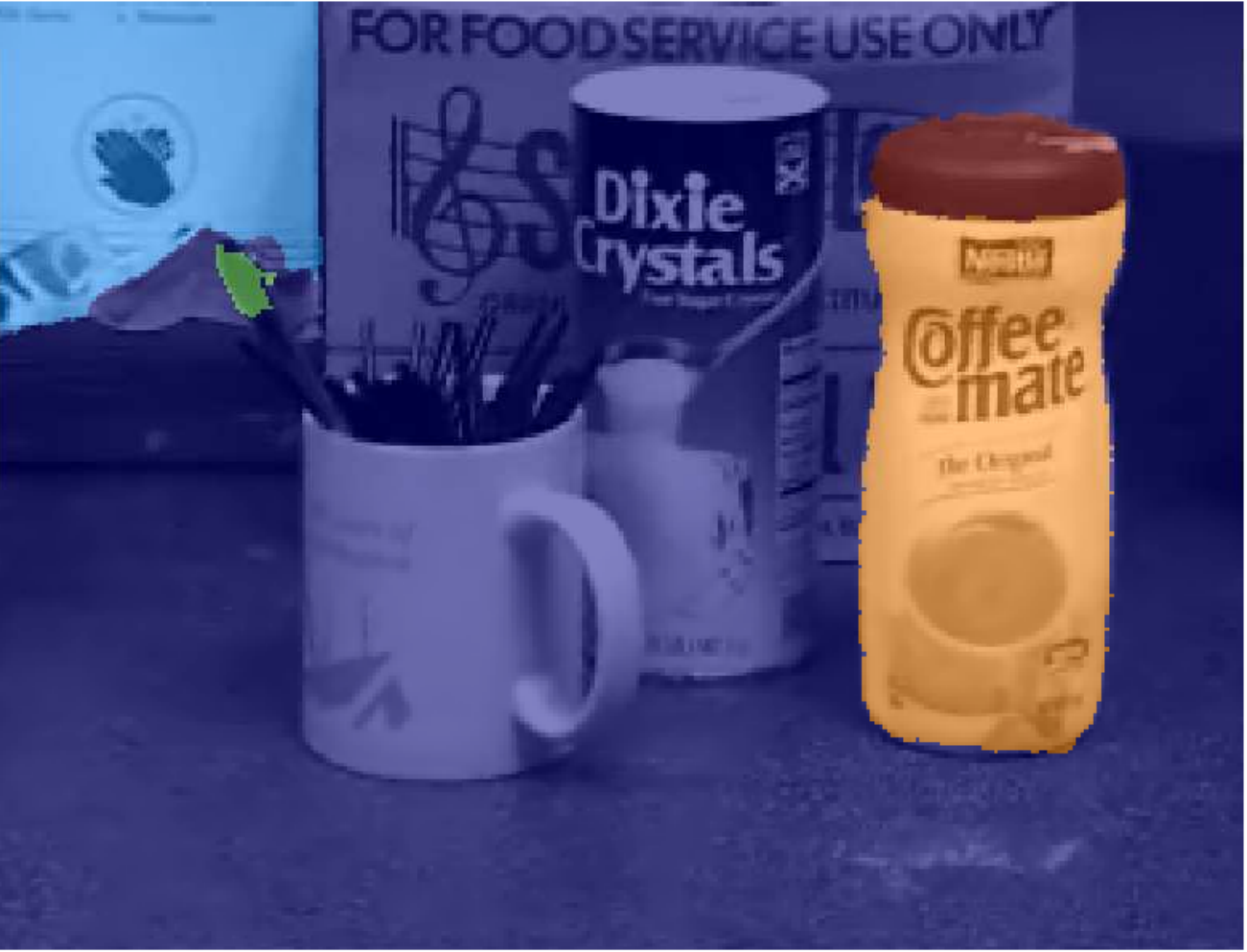}}
              \end{tabular}
            }
      }
     \caption{\sl {Sample results from the CMU dataset (first column), ground truth objects on these sequences (second column) affinities between superpixels computed on the reference frames (third column), output of our algorithm (fourth column). Note that color coding does not represent the layers rather the distinct components on the layer map.}}
	  \label{fig-qualitative}
\end{figure*}

Our method does not always work. Representative examples are shown in Fig. \ref{fig-compare}, where the closest box (second row) is not detected as a detachable object. This is because its support region is large and its occluding boundaries are not very salient. In order to detect it as a detachable object, one would have to see the box under sufficient parallax, for instance by moving around it, or to see it moved relative to its support base. One could also determine that the object is detachable by considering different images of the same object in different contexts, without any temporal continuity, but this would require (wide-baseline) co-segmentation \cite{rotherMBK06}, that is beyond the scope of this paper. The handles on the toy horse in Fig. \ref{fig-qualitative} (fourth row) are also not detected because the motion is too small and there is no significant motion signal around its boundaries. Longer sequences would easily disambiguate these objects.

Another failure mode can be seen in Fig. \ref{fig-qualitative} (top row), where the inside of the  bowl behind the chair is detected as a detached object,  because the water reflection violates the Lambertian assumption implicit in the occlusion detection functional.

Perhaps the most obvious failure mode is shown in the bottom line of Fig. \ref{fig-qualitative}, where only three of the objects (the Coffee mate container, a pen and the box in the background) are detected as detached. Again, this can be attributed to the small motion that does not elicit significant enough signal to trigger an occlusion detection.

\begin{figure*}[htb]
	\begin{center}
\includegraphics[width=0.24\textwidth]{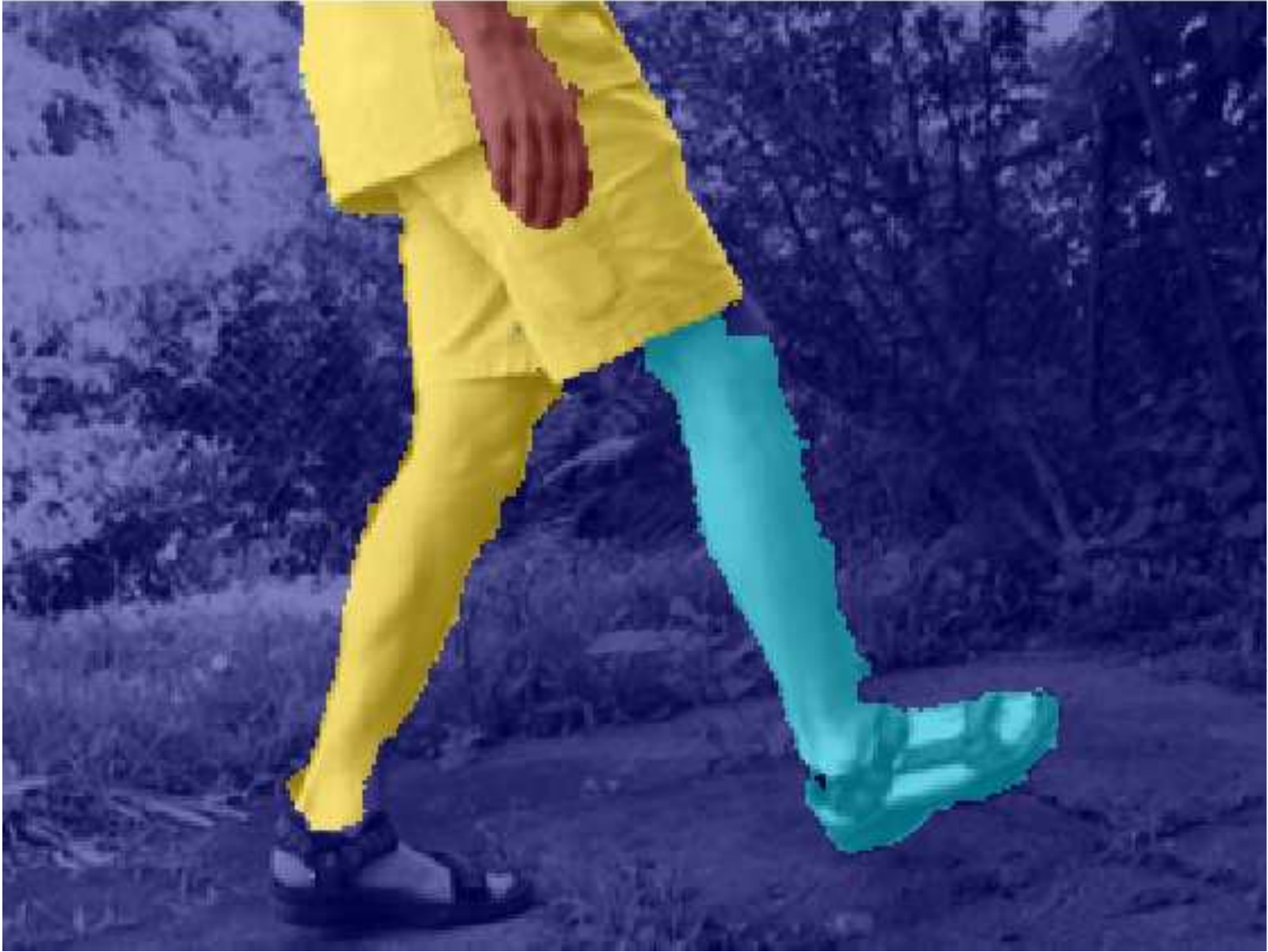}            
\includegraphics[width=0.24\textwidth]{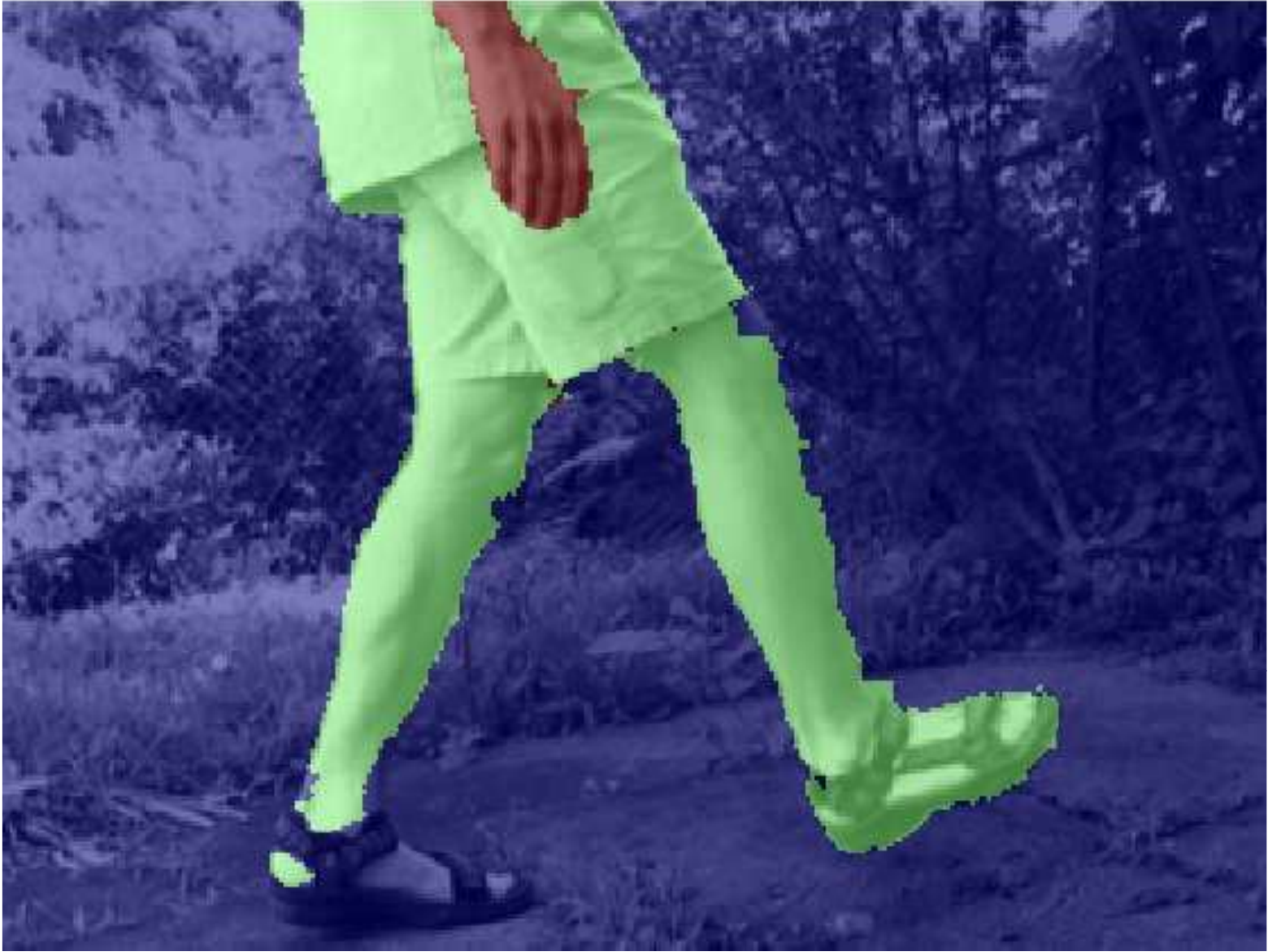}
\includegraphics[width=0.24\textwidth]{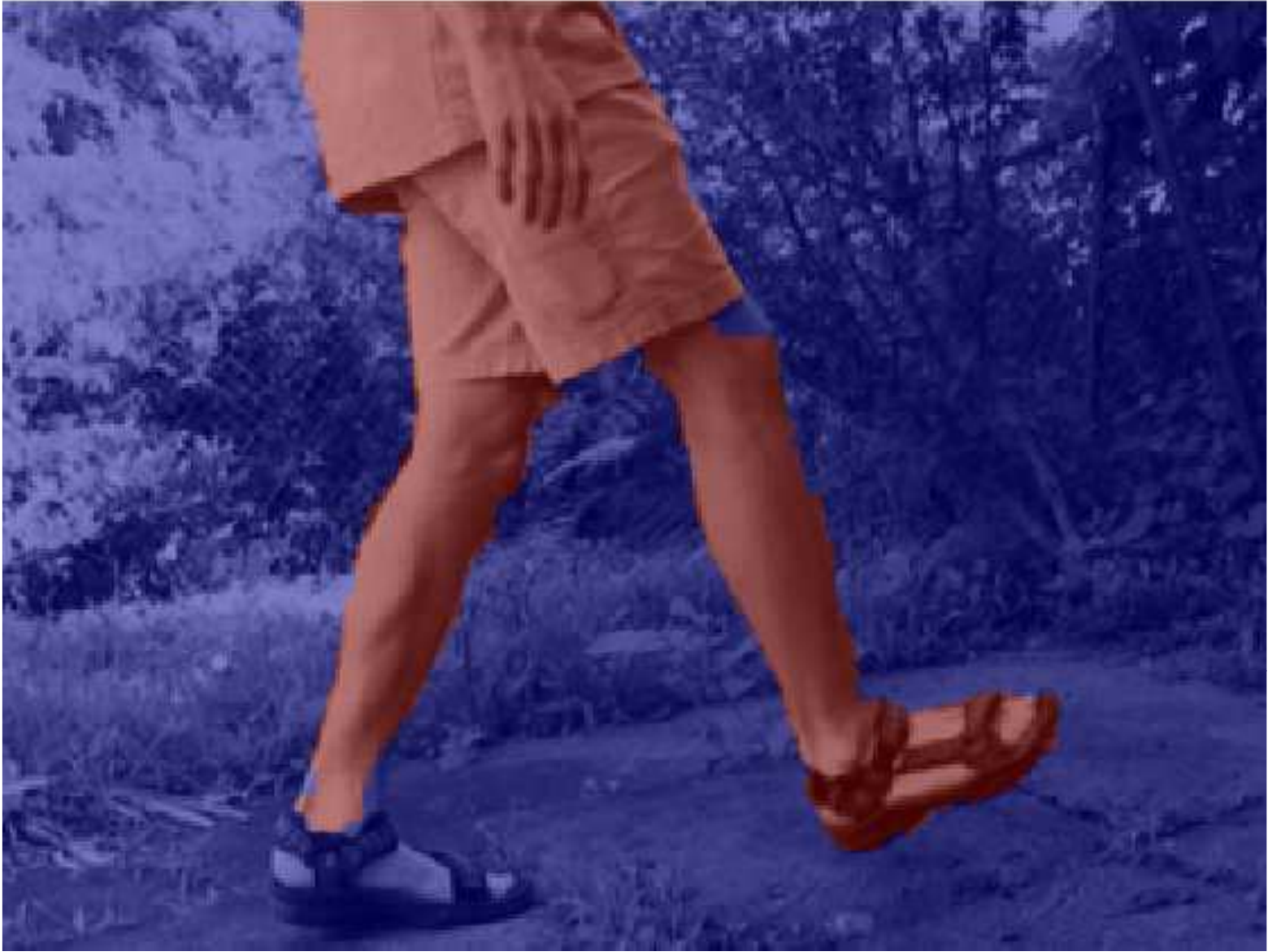} 
      \\
			         %{\includegraphics[width=0.24\textwidth]{rocking_horse-gray}}   	                
               {\includegraphics[width=0.24\textwidth]{rocking_horse-gtruth}}
			         {\includegraphics[width=0.24\textwidth]{rocking_horse-segs-overlay}}   	                
               {\includegraphics[width=0.24\textwidth]{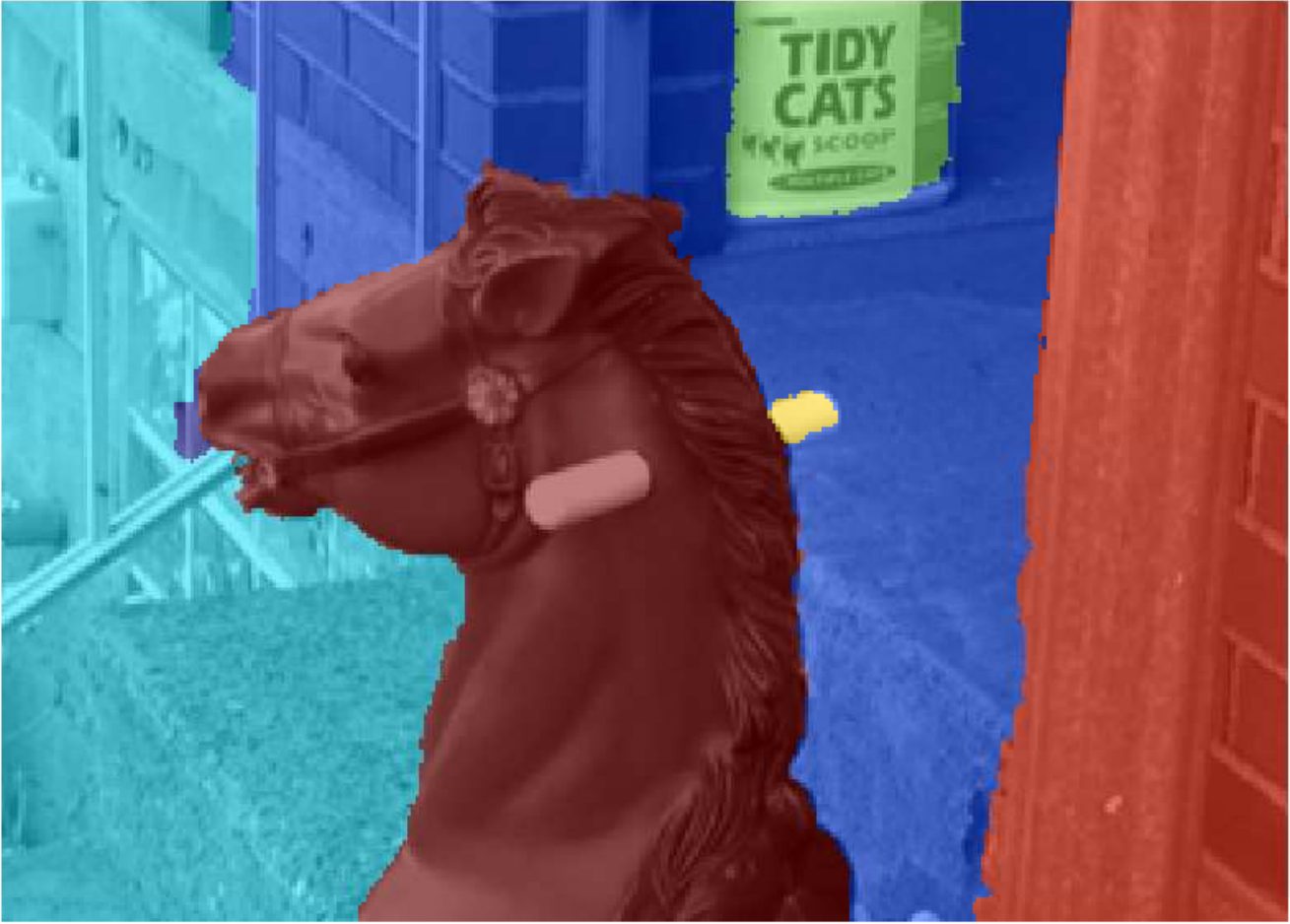}}\\
		   	       %{\includegraphics[width=0.24\textwidth]{couch_color-gray}}   	                
               {\includegraphics[width=0.24\textwidth]{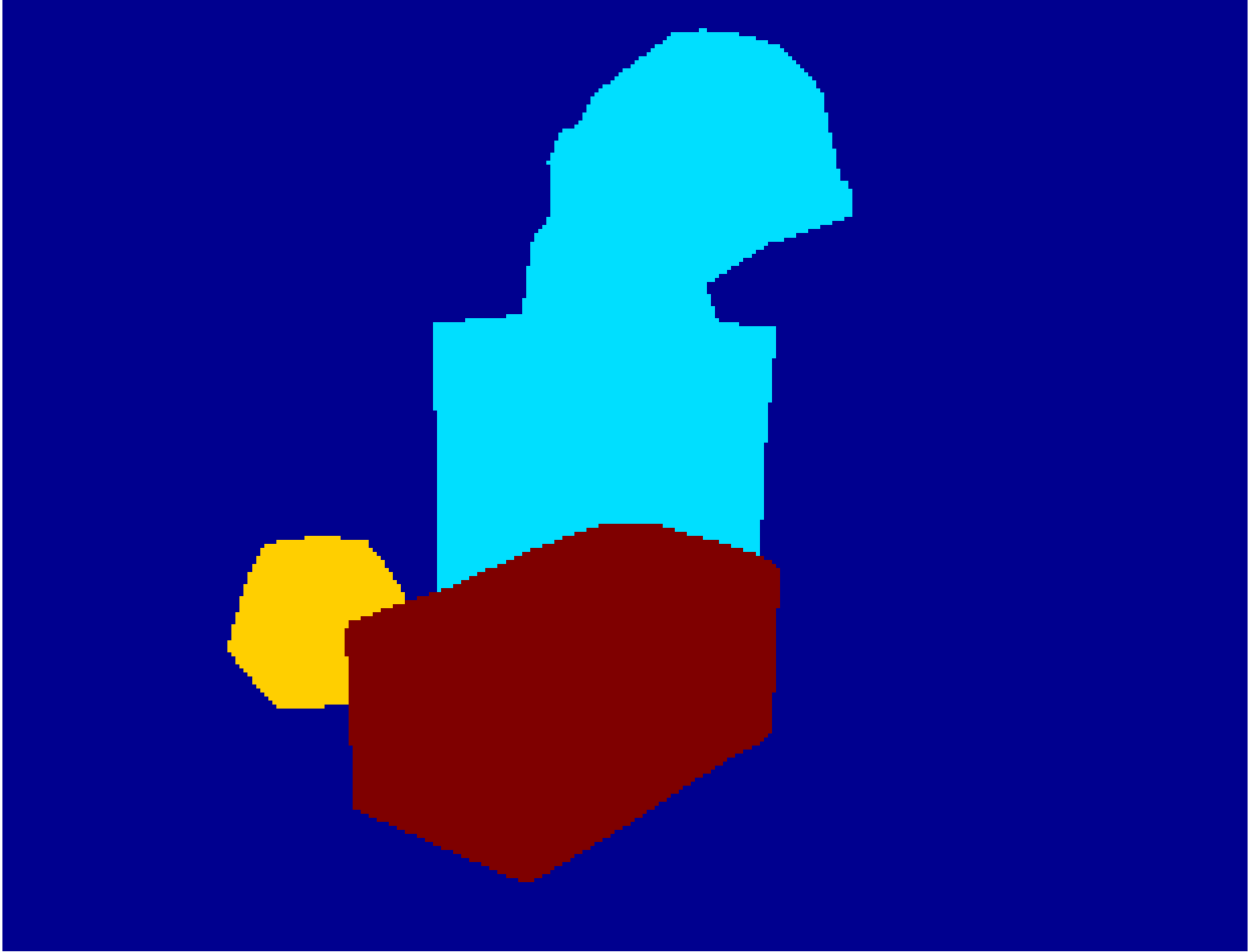}}
			         {\includegraphics[width=0.24\textwidth]{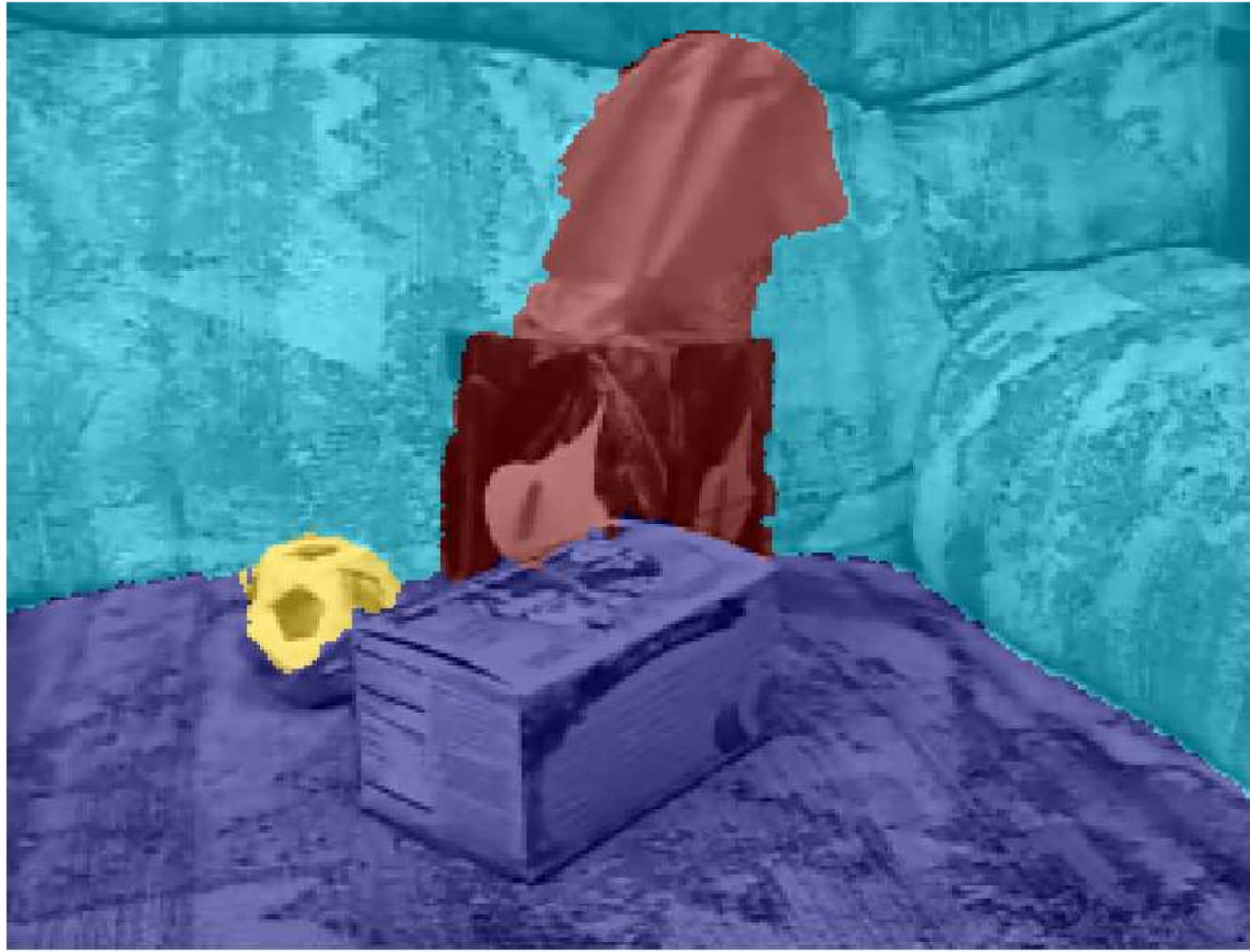}}   	                
               {\includegraphics[width=0.24\textwidth]{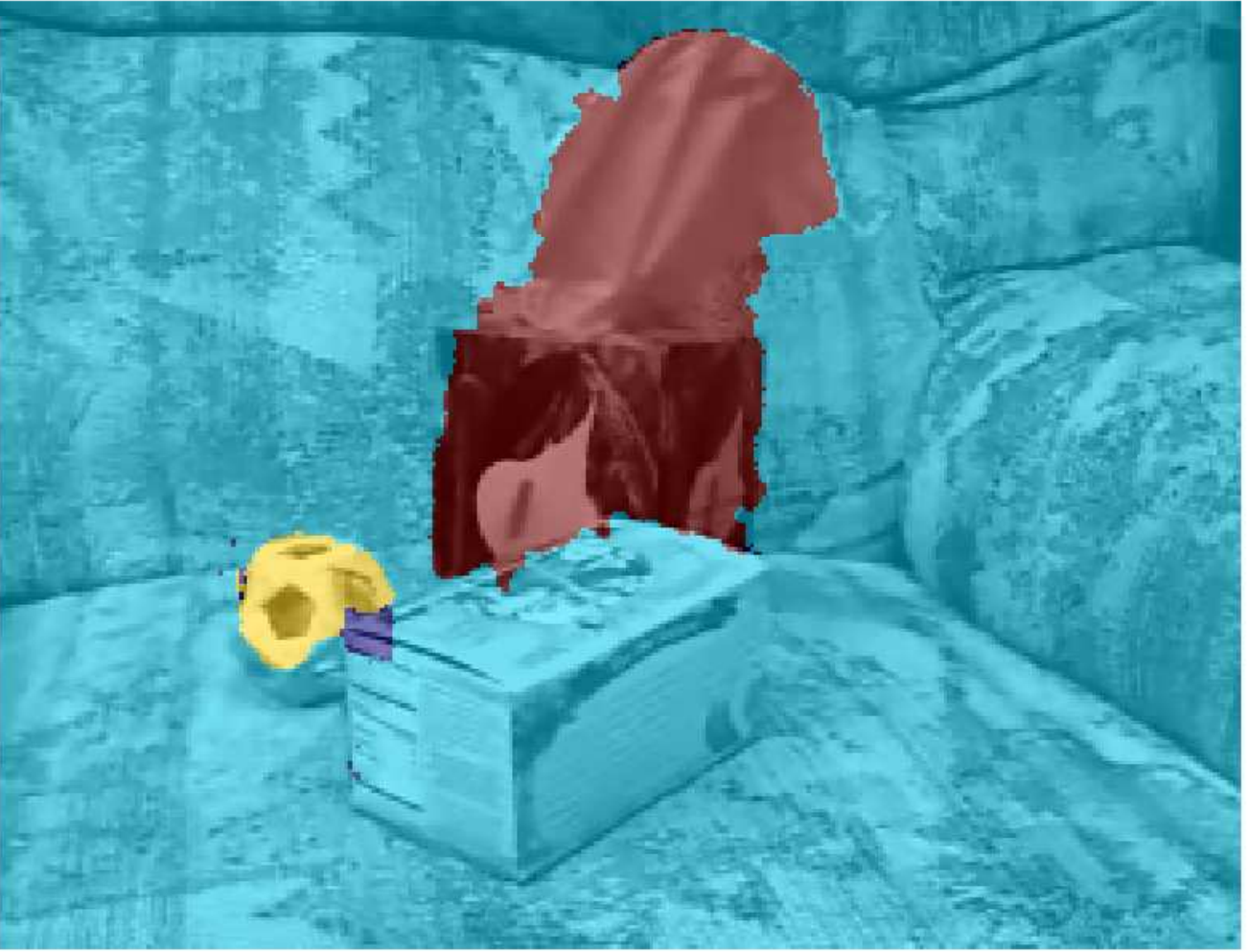}}

\end{center}

		\caption{\sl Top: Effects of increasing layer cost $\gamma$ on model selection. From left to right, the number of regions $\hat{\sigma}$ is estimated as $4$ (background, hand-arm, body and pivot leg, swinging leg), $3$ (background, body, and swinging arm) and $2$ (whole body). Note that the pivot foot is attached to the ground, and is therefore classified as such. Middle: Effects of model selection on representative samples from the CMU dataset. Ground truth is on the left, segmentation with manual setting of $L=4$ is in the middle, and segmentation with automatic model selection is on the right. Allowing automatic model selection enables the detection of the horse handles as separate detachable objects. On the bottom row, the algorithm fails to detect the closer cleenex box as a detachable object, because of insufficient parallax, so the box is either lumped with the horizontal cushion, or with the entire couch.}
\label{figure-label-cost}
\end{figure*}

Some of the failure modes are alleviated by automatic model selection (Fig. \ref{figure-label-cost}). 

\subsection{Quantitative assessment}

Our quantitative evaluation follows the lines of \cite{arbelaezMFM09}. The covering score of a set of ground truth segments $S^{\prime}$ by a set of segments $S$ can be defined as
\begin{equation}
	Score(S^{\prime},S) = \dfrac{1}{\sum_{s^{\prime} \in S^{\prime}} |s^{\prime}|} \sum_{s^{\prime} \in S^{\prime}} \max_{s \in S} \dfrac{|s \cap s^{\prime}|}{|s| + |s^{\prime}|}
\label{eq-score}
\end{equation}
 {Note that comparing our approach to \cite{steinSH08} is not straightforward, and possibly unfair, since the latter is an over-segmentation method where the number of segments are predetermined; our algorithm, on the other hand, performs automatic model selection. To be fair to \cite{steinSH08}, we have selected the cases where their algorithm yields a single segment, discarding all others that would negatively bias their outcome. \cite{steinSH08} reports  segmentation covering scores of 0.72 for the pedestrian, 0.84 for the tissue box and 0.71 for the squirrel which are depicted in red in Fig \ref{fig-compare}. By comparison, our algorithm achieves scores of 0.90, 0.95 and 0.90 respectively.}  

{We have also compared our method to normalized cut \cite{shiM02}, as the superpixel graphs depicted at Fig. \ref{fig-painted-child} can be partitioned using this technique. However, normalized cut  also requires the number of segments to be known \emph{a priori}, therefore, in our experiments, we have used self-tuning spectral clustering proposed by \cite{zelnikP04} which addresses this limitation. Our performance on the whole dataset considering all the ground truth objects is shown in Table \ref{table-perf}, which shows that our algorithm outperforms \cite{zelnikP04} in most of the sequences.}

\begin{table*}[hbt]
\begin{center}
  \resizebox{1.0\textwidth}{!}{
  \begin{tabular}{ |l | c | c | c | c | c | c | c | c|}
    \hline
                         & Bench & Car2 & Chair1 & Coffee Stuff &  Couch Color &  Couch Corner &  Fencepost & Hand3\\
    \hline
        Score with model selection                             & 0.89  & 0.52 & 0.78 & 0.43 & 0.63 & 0.95 & 0.42 & 0.74 \\ \hline
        Score \cite{zelnikP04}                   & 0.67  & 0.52 & 0.66 & 0.40 & 0.40 & 0.93 & 0.42 & 0.65 \\ \hline
        Score with true L   & 0.89  & 0.53 & 0.78 & 0.40 & 0.72 & 0.96 & 0.41 & 0.73 \\ \hline
        Score when forcing  L = 2        & 0.89  & 0.52 & 0.67 & 0.41 & 0.32 & 0.76 & 0.38 & 0.73 \\ \hline
        \hline
                             & Intrepid & Post & Rocking Horse & Squirrel4 &  Trash Can &  Tree & Walking Legs & Zoe1\\
        \hline
        Score with model selection                             & 0.85  & 0.98 & 0.78 & 0.90 & 0.75 & 0.69 & 0.92 & 0.72 \\ \hline
        Score \cite{zelnikP04}                   & 0.55  & 0.98 & 0.70 & 0.75 & 0.73 & 0.89 & 0.64 & 0.71 \\ \hline
        Score with  true L  & 0.66  & 0.98 & 0.77 & 0.91 & 0.75 & 0.74 & 0.91 & 0.72 \\ \hline
        Score when forcing L = 2        & 0.66  & 0.98 & 0.77 & 0.91 & 0.75 & 0.74 & 0.80 & 0.72 \\ \hline
  \end{tabular}
  }
\end{center}
\caption{\sl Performance of our approach on the CMU dataset computed based on the covering score (\ref{eq-score}) and compared to \cite{zelnikP04}, \cite{ayvaciS11} in case the correct number of layers is provided and \cite{ayvaciS11} when $L$ is set to $2$.}
\label{table-perf}
\end{table*}

As seen in Table \ref{table-perf}, testing with and without automatic model selection yields comparable results when the {\em correct} number of layers $L$ is given. However, automatic model selection significantly improves performance when the assumed number of layers is incorrect. 

In terms of running time, once occluded regions are detected, it takes $6.3$ seconds for CVX \cite{cvx} to solve the linear program (\ref{equation:relaxed-depth-ordering-soft}) with $310$ depth ordering constraints on a frame over-segmented to $4012$ superpixels. The run-time for occlusion detection in \cite{ayvaciRS11IJCV} can be reduced to a few seconds per frame using Split Bregman, depending on the size of the images; superpixelization is not a necessary step, but recent recent work has shown that it too can be performed at a rate of a few frames per second \cite{fulkersonS10}.

\section{Discussion}
\label{sect-discussion}

We have presented a method for detecting ``objects'' in a scene. While functionally important properties such as graspability cannot be ascertained from passive imaging data, we have defined properties that a {\em moving image} of an object must have in order to correlate to topological properties of the {\em scene}, such as being partially surrounded by the medium. We have defined these objects as {\em detachable}, conscious that this may be a misnomer in some cases, for instance houses and trees, despite Fig. \ref{fig-webster}.

Occlusions play a key role in the detection of detachable objects. Leveraging prior work, we show that once (binary) occlusion regions are available, integrating local ordering information into a coherent depth ordering map can be achieved by simple linear programming. The key to our approach is to convert a supervised segmentation problem into an unsupervised one using occlusions as the supervision mechanism. As a result, we have a fully unsupervised method for detecting and segmenting an unknown number of objects, and estimating their number in the meantime, all by solving a linear program. 

Our method is not panacea. Despite our efforts to manage errors in the occlusion detection stage, we still suffer from complete failures of the occlusion detection mechanism. In many cases, this is due to insufficient motion in the scene, and the results improve under extended temporal observation. 

Nevertheless, our results can still be useful as initialization of a more involved optimization over an extended temporal observation, that we and others have already developed. 

Our approach has a few tuning parameters, but fewer than most competing schemes since we perform model selection. Of course, even model selection requires tuning the tradeoff between complexity and fidelity, and there is no ``right'' choice of parameters. 

Our approach also shares the limitation of all schemes that break down the original problem (detached object detection, in our case) into a number of sequential steps, whereby failure of the early stages of processing cause failure of the entire pipeline. This predicament comes with the benefit of solving an otherwise very complex computational problem using efficient numerical schemes.

   \section*{Acknowledgment}
 This work was supported by ONR N000141110863. This technical report represents a revised version of a manuscript originally submitted for review on Feb. 22, 2011, that in turn was a revision of an earlier manuscript submitted on Nov. 19, 2010, and registered on that date as Technical Report UCLA-CSD-100036.

\bibliographystyle{plain}
\bibliography{ayvaciS11mdl}

\end{document}